%% file: main.tex
\documentclass[11pt]{chili}

\usepackage{multirow}
\usepackage{multicol}
\usepackage{needspace}
\usepackage{makecell}
\usepackage{array}
\usepackage{wrapfig}
\usepackage{float}
\usepackage{subcaption}
\usepackage[section]{placeins}
\usepackage{pgfplots}
\pgfplotsset{compat=1.18}
\usepackage{tikz}
\usetikzlibrary{patterns, positioning}

\theoremstyle{remark}

\input{qiushi_custom}

\showabstracttitlefalse

\usepackage[nameinlink,capitalise]{cleveref}
\crefname{section}{\S\kern-2.5pt}{\S\S\kern-2.5pt}
\Crefname{section}{\S\kern-2.5pt}{\S\S\kern-2.5pt}
\crefname{subsection}{\S\kern-2.5pt}{\S\S\kern-2.5pt}
\Crefname{subsection}{\S\kern-2.5pt}{\S\S\kern-2.5pt}
\crefname{subsubsection}{\S\kern-2.5pt}{\S\S\kern-2.5pt}
\Crefname{subsubsection}{\S\kern-2.5pt}{\S\S\kern-2.5pt}
\crefname{appendix}{\S\kern-2.5pt}{\S\S\kern-2.5pt}
\Crefname{appendix}{\S\kern-2.5pt}{\S\S\kern-2.5pt}

\title{OSReward: Instituting Standardized Evaluation for Cross-Platform Computer-Use Reward Models}

\ifanonauthors
\author{OSReward Contributors}
\else
\author[1\,\equalmark\leadmark]{Qiushi~Sun}
\author[2\,\equalmark\leadmark]{Kanzhi~Cheng}
\author[1,3\,\equalmark]{Yian~Wang}
\author[4\,\equalmark]{Bowen~Yang}
\author[5\,\equalmark]{Hang~Yan}
\author[6\,\equalmark]{Liheng~Chen}
\author[5]{Fangzhi~Xu}
\author[\space]{Zichen~Ding}
\author[2]{Nuo~Chen}
\author[2]{Jialin~Cao}
\author[2]{Xingdong~Gong}
\author[4]{Zehao~Li}
\author[3]{Kaiming~Jin}
\author[7]{Xinfeng~Yuan}
\author[7]{Zhoumianze~Liu}
\author[1]{Jingyang~Gong}
\author[7]{Zhangyue~Yin}
\author[1]{Jiahui~Gao}
\author[1]{Zhiyong~Wu}
\author[1]{Tianbao~Xie}
\author[2\,\corrmark]{Jianbing~Zhang}
\author[1\,\corrmark]{Ben~Kao}
\author[1]{Lingpeng~Kong}

\affil[1]{The University of Hong Kong}
\affil[2]{Nanjing University}
\affil[3]{National University of Singapore}
\affil[4]{University of Science and Technology of China}
\affil[5]{Xi'an Jiaotong University}
\affil[6]{University of Oxford}
\affil[7]{Fudan University}

\correspondingauthor={qiushisun@connect.hku.hk, chengkz@smail.nju.edu.cn,
zjb@nju.edu.cn, \{kao,lpk\}@cs.hku.hk\\ \authorlegend}
\fi

\begin{document}
\maketitle

\input{sections/0_abstract}
\input{sections/1_introduction}
\input{sections/2_related_work}

\input{sections/3_method}
\input{sections/4_experiments}
\input{sections/5_analysis}
\input{sections/6_osshepherd}
\input{sections/7_generalization}
\input{sections/8_conclusion}

\bibliographystyle{plainnat}
\bibliography{references}

\clearpage
\appendix
\appendixtoc
\input{appendices/A_infrastructure}
\input{appendices/B_osreward_details}
\input{appendices/C_experimental_details}
\input{appendices/D_osshepherd}
\input{appendices/E_results_analysis}

\input{appendices/F_case_studies}

\end{document}

%% file: qiushi_custom.tex
\newif\ifshowabstracttitle  \showabstracttitletrue
\newif\ifshowlogo            \showlogofalse
\newif\ifnumericcite         \numericcitefalse
\newif\ifappendixtoc         \appendixtoctrue
\newif\ifanonauthors         \anonauthorsfalse

\usepackage{xspace}
\usepackage{etoolbox}
\usepackage{fontawesome5}

\ifnumericcite
  \usepackage[numbers,square,sort&compress]{natbib}
\else
  \usepackage[round,authoryear]{natbib}
  \makeatletter
  \renewcommand\NAT@biblabel[1]{[#1]}
  \let\NAT@bibsetup\NAT@bibsetnum
  \makeatother
\fi

\usepackage[most]{tcolorbox}

\definecolor{AbstractBlue}{HTML}{EAF3FC}
\definecolor{ultramarineblue}{HTML}{120A8F}
\definecolor{darkblue}{rgb}{0, 0, 0.5}
\definecolor{veronica-red}{RGB}{196,30,58}
\definecolor{ForestGreen}{RGB}{34,139,34}
\definecolor{BrickRed}{rgb}{.72,0,0}
\definecolor{LakeBlue}{RGB}{0,61,153}
\definecolor{boxgreen}{rgb}{0.9,1.0,0.9}
\definecolor{boxblue}{rgb}{0.9,0.9,1.0}
\definecolor{boxred}{rgb}{1.0,0.85,0.95}

\colorlet{MorandiAbstract}{AbstractBlue}

\newcommand{\abstractheadingfont}{\normalfont\bfseries\fontsize{14}{16}\selectfont\color{ultramarineblue}}
\makeatletter
\pretocmd{\archer@renderabstract}{%
  \vspace{-1.15\baselineskip}%
  \ifshowabstracttitle
    \vspace{9pt}%
    {\centering\abstractheadingfont\abstractname\par}%
    \vspace{-8pt}%
  \fi
}{}{}
\makeatother

\newcommand{\qsheadingfont}{\color{ultramarineblue}\bfseries\fontsize{13}{14}\selectfont}
\titleformat{\section}
  {\large\bfseries\qsheadingfont}{\thesection.}{0.5em}{#1}[]
\titleformat{name=\section,numberless}
  {\large\bfseries\qsheadingfont}{}{0em}{#1}[]
\titleformat{\subsection}
  {\color{ultramarineblue}\bfseries}{\thesubsection.}{0.5em}{#1}[]
\titleformat{\subsubsection}
  {\color{ultramarineblue}\bfseries\itshape}{\thesubsubsection.}{0.5em}{#1}[]
\titleformat{\paragraph}[runin]
  {\bfseries\color{ultramarineblue}}{\theparagraph.}{0.5em}{#1}
\titlespacing*{\paragraph}{0pt}{1.0ex plus 0.5ex minus .2ex}{1em}

\renewcommand{\titlefont}{\color{BrandPrimary}\normalfont\bfseries\fontsize{19}{23}\selectfont}

\newcommand{\logoblock}{%
  \fcolorbox{BrandRule}{BrandLight}{\rule[-0.45em]{0pt}{1.5em}\scriptsize\itshape\,logo 1\,}\hspace{4pt}%
  \fcolorbox{BrandRule}{BrandLight}{\rule[-0.45em]{0pt}{1.5em}\scriptsize\itshape\,logo 2\,}%
}
\makeatletter
\let\ps@firststyleorig\ps@firststyle
\@namedef{f@nch@ps@firststyleorig-is-fancyhdr}{}
\makeatother
\fancypagestyle{firststyle}[firststyleorig]{\fancyhead[L]{\ifshowlogo\logoblock\fi}}

\newcommand{\equalmark}{\ensuremath{\ast}}
\newcommand{\leadmark}{\ensuremath{\dagger}}
\newcommand{\corrmark}{\faEnvelope[regular]}

\newcommand{\authorlegend}{\upshape\footnotesize
  \textsuperscript{\ensuremath{\ast}}\,Equal contribution\quad
  \textsuperscript{\ensuremath{\dagger}}\,Project co-lead\quad
  \textsuperscript{\faEnvelope[regular]}\,Corresponding author}

\newtcolorbox{notebox}[1][]{%
  enhanced, breakable,
  colback=boxblue, colframe=LakeBlue, boxrule=0.5pt, arc=2pt,
  left=6pt, right=6pt, top=4pt, bottom=4pt,
  fonttitle=\bfseries, coltitle=white,
  #1
}

\newtcolorbox{takeawaybox}{%
  enhanced, breakable,
  colback=AbstractBlue, colframe=AbstractBlue, boxrule=0pt, arc=1.5pt,
  borderline west={2.5pt}{0pt}{LakeBlue},
  left=7pt, right=7pt, top=3.5pt, bottom=3.5pt,
  before skip=7pt plus 2pt, after skip=7pt plus 2pt,
}
\newcommand{\takeaway}[1]{%
  \begin{takeawaybox}\textbf{Takeaway:} #1\end{takeawaybox}}

\makeatletter
\newcommand{\appendixtocname}{Appendix Contents}
\newcommand{\appendixtoc}{%
  \ifappendixtoc
    \section*{\appendixtocname}%
    \@starttoc{apc}%
    \let\qs@oldaddcontentsline\addcontentsline
    \renewcommand\addcontentsline[3]{%
      \def\qs@tmp{##1}\def\qs@toc{toc}%
      \ifx\qs@tmp\qs@toc \qs@oldaddcontentsline{apc}{##2}{##3}\fi}%
  \fi
}
\makeatother

\newcommand{\cmark}{\textcolor{ForestGreen}{\ding{51}}}
\newcommand{\xmark}{\textcolor{BrickRed}{\ding{55}}}

\newcounter{reviewmark}
\DeclareRobustCommand{\review}[1]{\stepcounter{reviewmark}%
  \textcolor{BrickRed}{\textbf{[review~\arabic{reviewmark}]}}}

\newcommand{\tablogo}[1]{\raisebox{-1.6pt}{\includegraphics[height=8.5pt]{figures/logos/#1.png}}\hspace{2.5pt}}

\newcommand{\oudata}{OS-Shepherd-100K\xspace}
\newcommand{\model}{OS-Shepherd\xspace}

\newcommand{\modelb}{OS-Shepherd-35B-A3B\xspace}

\newcommand{\OSShepNineB}{OS-Shepherd-9B\xspace}
\newcommand{\QwenNineB}{Qwen3.5-9B\xspace}
\newcommand{\OpusEight}{Claude-Opus-4-8\xspace}
\newcommand{\OpusSix}{Claude-Opus-4-6\xspace}
\newcommand{\SonnetSix}{Claude-Sonnet-4-6\xspace}

\newcommand{\GPTss}{GPT-5.5\xspace}

\newcommand{\GPTfourpt}{GPT-5.4\xspace}
\newcommand{\GPTtwopt}{GPT-5.2\xspace}
\newcommand{\GPTfivemini}{GPT-5-mini\xspace}
\newcommand{\GPTfourmini}{GPT-5.4-mini\xspace}

\newcommand{\GThreeFlash}{Gemini-3-Flash\xspace}

\newcommand{\KimiK}{Kimi-K2.5\xspace}
\newcommand{\QwenBigA}{Qwen3.5-397B-A17B\xspace}

\newcommand{\QwenMidA}{Qwen3.5-35B-A3B\xspace}
\newcommand{\QwenVLeight}{Qwen3-VL-8B\xspace}
\newcommand{\QwenVLthirty}{Qwen3-VL-30B\xspace}

\newcommand{\InternOne}{Intern-S1-Pro\xspace}

\usepackage{graphicx}
\usepackage{xcolor}
\usepackage{caption}
\usepackage{float}
\usepackage{needspace}
\usepackage{tikz}
\usepackage[most]{tcolorbox}
\let\caseoriginalsfdefault\sfdefault
\usepackage{tgheros}
\let\sfdefault\caseoriginalsfdefault

\definecolor{caseheadingblue}{HTML}{00205B}
\definecolor{casejudgefill}{HTML}{F6F7F6}
\definecolor{casejudgestroke}{HTML}{C7CECC}
\definecolor{casejudgetext}{HTML}{2F3435}
\definecolor{casejudgeaccent}{HTML}{566B7B}
\definecolor{casejudgebluefill}{HTML}{E9EDF0}
\definecolor{casejudgebluebar}{HTML}{71879A}
\definecolor{casejudgewarmfill}{HTML}{F7F1EA}
\definecolor{casejudgewarmstroke}{HTML}{DCCDBE}
\definecolor{casejudgewarmaccent}{HTML}{957760}

\providecommand{\caseassetdir}{figures/case_study}

\newcommand{\casehead}[1]{%
  \textcolor{caseheadingblue}{\bfseries #1}%
}
\newcommand{\summarylabel}[1]{%
  {\fontsize{6.5}{7.6}\selectfont\color{casejudgeaccent}\bfseries
   \MakeUppercase{#1}}%
}
\newcommand{\judgeline}[1]{%
  \par\noindent
  \hangindent=1.15em
  \hangafter=1
  \textcolor{casejudgewarmaccent}{\bfseries\textbullet}\hspace{0.48em}#1%
}
\newtcolorbox{casejudgesummarypanel}{%
  enhanced,
  colback=casejudgefill,
  colframe=casejudgestroke,
  boxrule=0.4pt,
  arc=6pt,
  boxsep=0pt,
  left=5pt,right=5pt,top=4pt,bottom=4pt,
  before skip=0pt,after skip=0pt
}
\newtcolorbox{casejudgestatusblock}{%
  enhanced,
  colback=casejudgebluefill,
  colframe=casejudgestroke,
  boxrule=0.3pt,
  arc=3pt,
  boxsep=0pt,
  left=6pt,right=6pt,top=3pt,bottom=3pt,
  borderline west={2.2pt}{0pt}{casejudgebluebar}
}
\newtcolorbox{casejudgeaccountblock}{%
  enhanced,
  colback=casejudgebluefill,
  colframe=casejudgestroke,
  boxrule=0.3pt,
  arc=3pt,
  boxsep=0pt,
  left=5pt,right=5pt,top=3.5pt,bottom=3.5pt,
  borderline west={2.2pt}{0pt}{casejudgebluebar}
}
\newtcolorbox{casejudgereadingsblock}{%
  enhanced,
  colback=casejudgewarmfill,
  colframe=casejudgewarmstroke,
  boxrule=0.3pt,
  arc=3pt,
  boxsep=0pt,
  left=5pt,right=5pt,top=3.5pt,bottom=3.5pt,
  borderline west={2.2pt}{0pt}{casejudgewarmaccent}
}

\newcommand{\casefigurewithjudges}[7]{%
  \par\vspace{0.18em}
  \begin{figure}[H]
    \centering
    \begin{minipage}{0.90\linewidth}
    \centering
    \setlength{\parskip}{0pt}
    \includegraphics[width=\linewidth]{\caseassetdir/#1}\par
    \nointerlineskip
    \begin{casejudgesummarypanel}
        \color{casejudgetext}
        \fontfamily{qhv}\selectfont
        \setlength{\parskip}{0pt}
        \fontsize{7.4}{9.15}\selectfont
        \summarylabel{Judging summary}\par
        \vspace{0.14em}
        \begin{casejudgestatusblock}
          {#4}
        \end{casejudgestatusblock}
        \vspace{0.16em}
        \noindent
        \begin{minipage}[t]{0.31\linewidth}
          \vspace{0pt}
          \begin{casejudgeaccountblock}
            \raggedright
            \summarylabel{Agent's account}\par
            \vspace{0.08em}
            #5\par
          \end{casejudgeaccountblock}
        \end{minipage}
        \hfill
        \begin{minipage}[t]{0.66\linewidth}
          \vspace{0pt}
          \begin{casejudgereadingsblock}
            \raggedright
            \summarylabel{Representative readings}
            \judgeline{#6}
            \vspace{0.04em}
            \judgeline{#7}
          \end{casejudgereadingsblock}
        \end{minipage}
    \end{casejudgesummarypanel}
    \end{minipage}
    \captionsetup{
      font=footnotesize,
      labelfont=bf,
      justification=raggedright,
      singlelinecheck=false,
      skip=0.38em,
      margin=0pt
    }
    \caption{#2\label{#3}}
  \end{figure}
  \par\vspace{0.18em}
}

%% file: sections/0_abstract.tex
\begin{abstract}
Computer-using agents (CUAs) are advancing rapidly across the digital world. 
A CUA trajectory records the agent's actions, states, and reasoning. Verifying whether it fulfilled the task instruction is central to CUA evaluation, data curation,
and reinforcement learning. Neither human-written verifiers nor human annotators can provide such verification at scale, so the field increasingly turns to vision-language models (VLMs) as judges of CUA trajectories. But a fundamental question has long gone unexamined: are these VLM judges reliable enough? To study it systematically, we introduce OSReward, a realistic, high-quality benchmark that evaluates VLM judges on CUA trajectories. The trajectories come from diverse agent backbones executing human-verified instructions across platforms, and are then rigorously labeled with ground-truth verdicts through multi-stage human annotation. Building on it, we derive OSReward-Hard, a challenge set concentrating genuinely hard cases, and OSReward-Multi for fine-grained efficiency and alignment scoring. The most comprehensive evaluation of VLM judges to date finds even state-of-the-art models fall short of an ideal judge, sharing a systematic
leniency bias that mislabels failed runs as successes. The few reliable enough
to trust are too expensive to run at scale, while affordable open models trail
far behind. To close this gap, we construct and release \oudata, an open
corpus of reasoning-annotated trajectory judgments for the CUA community. On it,
we train OS-Shepherd (9B and 35B), open reward models that supply low-cost, stable,
and reliable reward signals, matching commercial judges at 30--60$\times$ lower
cost than the frontier. Extensive analyses further inform the design of reliable CUA reward at
scale.
Our code,
benchmark, dataset, and model checkpoints are available at \href{https://os-copilot.github.io/OSReward-Home/}{OSReward Homepage}.
\end{abstract}

\vspace{-1.5\baselineskip}
\begin{center}
    \captionsetup{type=figure}%
    \setlength{\abovecaptionskip}{4pt}%
    \captionsetup[sub]{font=footnotesize}%
    \subcaptionbox{Performance on OSReward and OSReward-Hard.\label{fig:teaser-hard}}[245pt]{%
        \includegraphics[height=8.4cm]{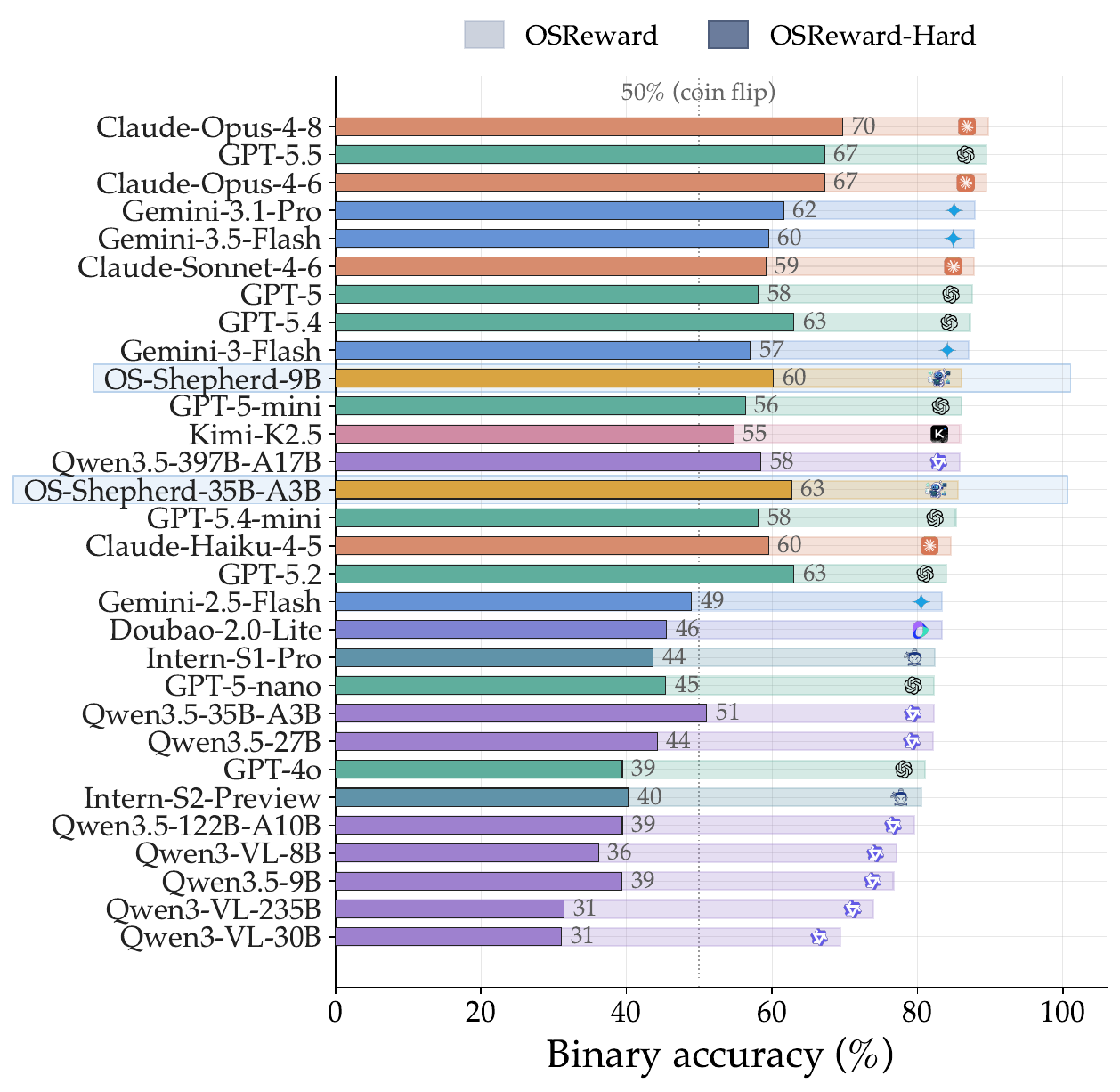}}%
    \hspace{0.15cm}%
    \subcaptionbox{Judge bias across different models.\label{fig:teaser-bias}}[235pt]{%
        \includegraphics[height=8.4cm]{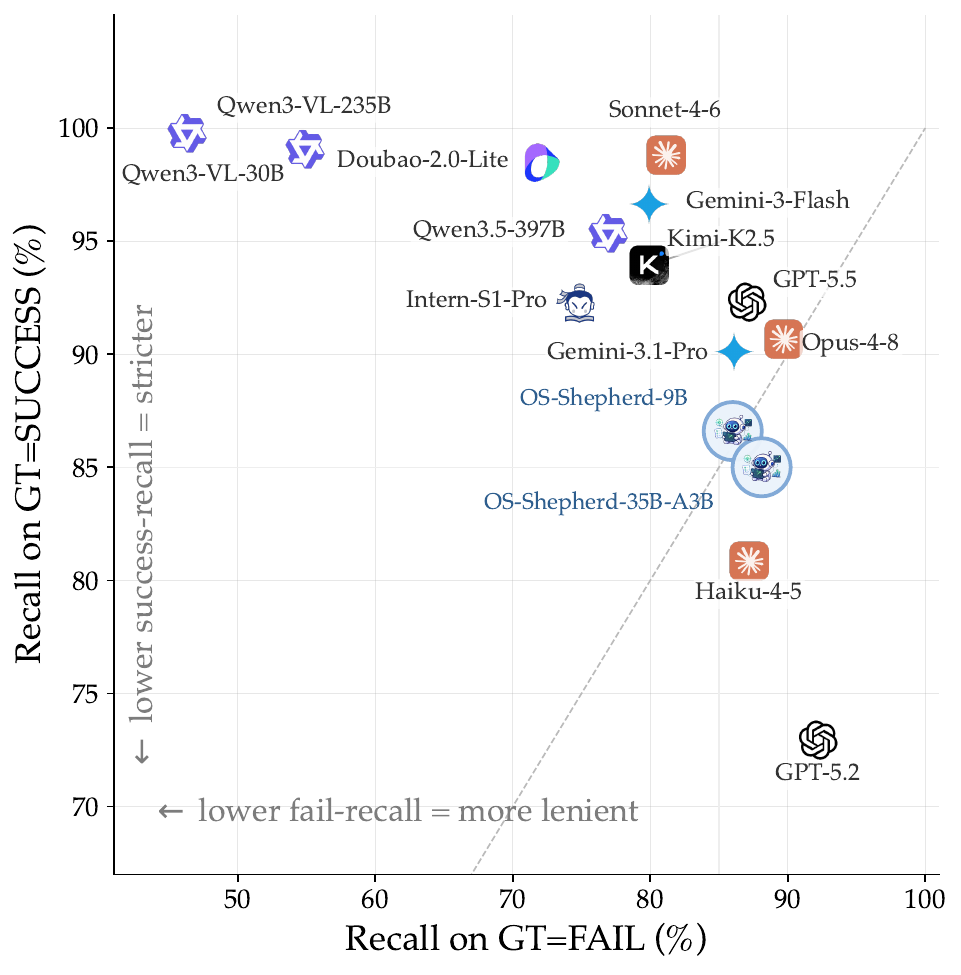}}%
    \caption{How well VLM judges score CUA trajectories (a), and the strict-lenient bias they share (b).}
    \vspace{-1em}
    \label{fig:teaser}
\end{center}

%% file: sections/1_introduction.tex
\section{Introduction}
\label{sec:intro}

Computer-using agents (CUAs) are advancing rapidly across the digital world, operating the web,
mobile apps, and desktop software~\citep{osworld, webarena, androidworld}. 
A \emph{trajectory}
is the interleaved sequence of the agent's actions, states, and reasoning. Scaling that
progress, through evaluation, data curation, and reinforcement
learning~\citep{xue2026acurl}, requires verifying whether each trajectory fulfilled its task
instruction. However, human-written verifiers cover only a handful of curated tasks, and they do
not apply at all to the static corpora and previously collected trajectories that have no live
environment left to inspect. Human annotation cannot keep pace either. Typical methods therefore
revolve around a VLM acting as the judge~\citep{llmjudge,zhuge2025agentasajudge}, whether as reward
models~\citep{pan2024autonomous, li2026osthemis, chae2026webshepherd} or autoraters~\citep{sun2025osgenesis}, an approach fast becoming the
de-facto practice.

Yet whether this judge is actually reliable remains largely untested. Unlike judging
text~\citep{llmjudge, zhou2025rmb} or general multimodal agents~\citep{chen2024mllm}, judging a CUA trajectory means reading a long, interleaved record of states, actions, and reasoning, then deciding whether the environment truly reached the instructed goal rather than whether the agent merely claims so, a verdict that can be reached from a fragment of that record. 
No prior study measures how reliably models make this judgment across platforms, 
and our pilot study shows the problem is real: even the best VLM judges disagree with existing benchmarks' own verifiers on roughly a quarter of desktop verdicts.
Our generalization study (\cref{sec:generalization}) measures this disagreement at scale and discusses where it originates, 
in the judges or in the verifiers and tasks themselves.

To study this systematically, we build OSReward, a benchmark for evaluating the judge itself.
Reusing existing agent benchmarks' off-the-shelf trajectories~\citep{lin2025cuarewardbench,
li2026osthemis} would confound judge errors with flaws in the runs themselves, leaving failures
unattributable to the judge. OSReward is instead built on dedicated cross-platform data
infrastructure that we design and operate end to end: stock environments on web, mobile, Ubuntu, 
and Windows, extended with the common~\citep{osworld} and professional~\citep{sun2026scienceboard} applications a real user has and realistic starting states. This affords coverage far beyond reused trajectories: more applications, websites, and scenarios, spanning both pure-GUI and GUI+CLI workflows. On these environments, annotators
curate verified, environment-grounded instructions. Agents from four model families then execute them, and their varying capability yields both real successes and real failures. Each trajectory then passes rigorous multi-stage human labeling with strict screening. We obtain 1019 human-gold trajectories, long-horizon (up to 100 steps), organized into three views: the full set carries the breadth;
OSReward-Hard concentrates genuinely hard cases, built from the trajectories the annotators themselves split on, where current judges commonly fail, exposing error modes and
remaining headroom;
and OSReward-Multi layers fine-grained efficiency and alignment labels on top of the binary verdicts. 
Grounded in careful human annotation, this labeled set offers a
clean, standardized basis for studying VLM judges.

We then run the broadest evaluation of VLM judges yet, benchmarking 27 models. Frontier judges
look adequate on the full set but fall sharply on OSReward-Hard, where the best judge drops
below 70\% and the mean judge falls to 52\%. The errors share one signature: 
judges are easily fooled by \emph{false successes},
where the agent declares the task completed but has
actually failed, accepting such runs far more often than the reverse. The few judges accurate
enough to trust, such as \OpusEight~\citep{anthropic2026opus48} and
\GPTss~\citep{openai2026gpt55}, are far too expensive for the millions of judgments that
evaluation, curation, and training demand, while affordable open judges trail by a wide margin,
leaving no judge both reliable and affordable.

We therefore close this gap with open data and open reward models. We run our own large-scale
collection pipeline, process and screen its output, and curate it together with filtered public
corpora into \oudata, an open corpus of 100K reasoning-annotated trajectory judgments that spans
more trajectories, agent backbones, scenarios, and task types than reused corpora provide.
Guided by what our evaluation reveals, we label these trajectories without new human annotation,
each labeling choice set by a finding of the judge study. We release the corpus in full, with
failure types annotated and analyzed, so the community can push open CUA reward research
forward.

\begin{figure}[t]
    \centering
    \includegraphics[width=\linewidth]{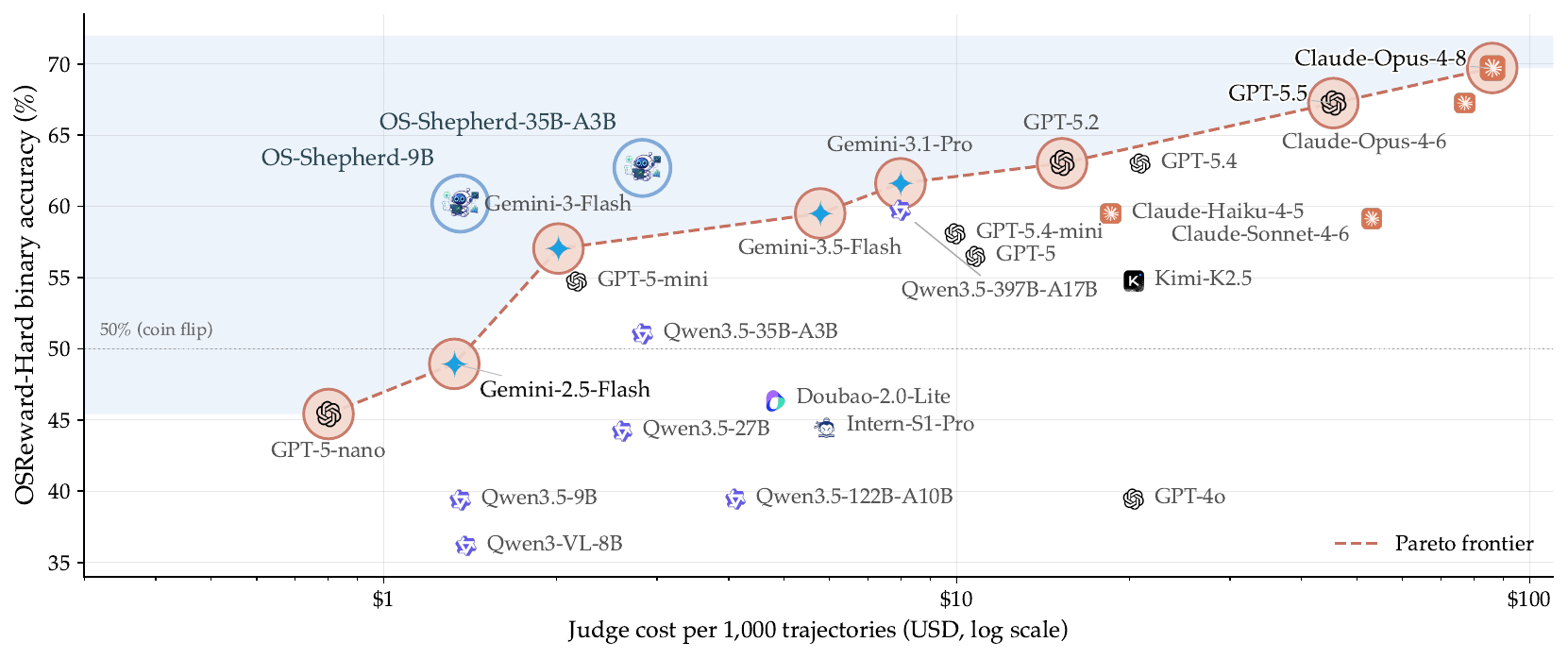}
    \caption{Cost against binary accuracy on OSReward-Hard: reliable judges
    are expensive, and the OS-Shepherd models come nearest their accuracy at
    a fraction of the cost.}
    \label{fig:osshep-pareto}
\end{figure}

Based on \oudata, we train OS-Shepherd-9B and OS-Shepherd-35B with a two-stage recipe that first
builds accurate judging and then directly targets the false success, the sharpest failure the
study exposes. 
On the cost-accuracy frontier (\cref{fig:osshep-pareto}),
the \model models come nearest the reliable judges at 30--60$\times$ lower cost, putting a training-scale
reward signal within an academic budget.
Further analyses,
spanning input ablations, robustness, and ensembling, reveal when and why judges fail, and
held-out evaluations confirm the de-biasing transfers to unseen benchmarks.

\paragraph{Contributions.}
\begin{itemize}[leftmargin=1.3em, itemsep=2pt, topsep=2pt, parsep=0pt]
    \item \textbf{OSReward.} A standardized benchmark for the CUA reward
    signal, built from scratch on dedicated cross-platform data
    infrastructure: human-gold trajectories across four platforms, plus the
    OSReward-Hard challenge set for hard-case diagnosis and the
    OSReward-Multi subset for fine-grained judging.
    \item \textbf{Comprehensive Evaluation.} We present the broadest study of VLM judges for CUA yet, 
    spanning platforms and difficulty: judges share one lenient failure mode, read the agent's narrative more than the screen, and collapse on hard cases, 
    while the reliable ones cost too much to run at scale. Extensive analyses isolate what drives a verdict.
    \item \textbf{OS-Shepherd Data \& Models.} An open, reasoning-annotated training
    corpus (\oudata) curated from more than 300K judge instances on
    cross-platform trajectories into nearly 100K samples, rich in both
    successes and failures, and the two \model reward models (9B and 35B):
    open, low-cost, and self-hostable.
\end{itemize}

%% file: sections/2_related_work.tex
\needspace{5\baselineskip}
\section{Related Work}
\label{sec:related}

\subsection{Computer-Using Agents}
\label{subsec:related-cua}
Computer-using agents perceive digital environments and take actions (e.g., clicks, typing, or CLI commands) to complete user tasks~\citep{openai2025cua, wu2024oscopilot, sun2024survey}. 
Driven by vision-language models, the field has advanced quickly from grounding instructions on raw screens~\citep{zheng2024seeact, cheng2024seeclick, gou2025navigating, wu2025gui} to native GUI action models that operate real applications end to end~\citep{wu2025osatlas, xu2025aguvis, qin2025uitars}, joined by open cross-platform agent data~\citep{wang2026opencua, liu2026scalecua, xu2025agenttrek} and systems that compose heterogeneous agents~\citep{jia2025agentstore}.
Standardized environments across web, mobile, and desktop~\citep{chen2025map, webarena, kong2026mobileworld} give this progress reproducible tasks and testbeds. That reproducibility rests on hand-written verifiers, 
one per task, and many tasks in real use have no checkable outcome at all.
Evaluation and training in these environments are outgrowing those verifiers: trajectory data is collected and processed at scale~\citep{qin2025uitars, xue2026acurl}, 
policies are optimized with reinforcement learning over ever-longer, open-ended interactions~\citep{bai2024digirl, qi2024webrl, sun2025coda, xu2026odysseyarena}, 
and agents are expected to eventually self-evolve on synthetic experience~\citep{zhang2026agent, xue2026evocua, jiang2026treecua}.
Every stage of this pipeline consumes a reward signal, and an ideal judge should be general and stable at the trajectory level. We build OSReward from scratch to study this problem systematically.

\subsection{Judging CUA Trajectories}
\label{subsec:related-judge}
Evaluating and training CUAs both require deciding whether a trajectory fulfilled its instruction, at a scale that, as the introduction argues, neither human-written verifiers nor human annotation can reach, leaving model-based judging as the only practical route. Judging a CUA trajectory, however, is much more complex than judging text~\citep{llmjudge, cobbe2021training, wang2024mathshepherd} or general multimodal responses~\citep{chen2024mllm,li2025vl}: the verdict must track a long, interleaved record and sometimes even infer whether the environment after execution truly fulfills the task.
An emerging line of work nevertheless puts a VLM judge in exactly this role, 
as the reward model behind trajectory synthesis and data filtering~\citep{sun2025osgenesis, wang2026synthagent, zhang2025guimid} and as the basis of trained reward models and critic-style autoraters~\citep{chen2025guishepherdandroid, xiao2025uigenie, li2026osthemis, chae2026webshepherd, wu2026oracle}.
Although recent work has gradually become aware of the reliability issue in model-based reward~\citep{lin2025cuarewardbench, lu2025agentrewardbenchweb},
these early efforts stay on single, 
isolated platforms and reuse off-the-shelf instructions or trajectories from common benchmarks.
Platforms differ in what a judge must read~\citep{xue2025an, sun2025sentinel}: action spaces, the applications and websites in play, and run lengths from around a dozen steps on mobile to a hundred on desktop.
Reused trajectories bring problems of their own: they inherit whatever rollout setting produced them, 
from agent backbone to budget; their gold labels may also come from imperfect verifiers~\citep{osworldverified}; and some of their instructions are ambiguous enough that no determinate verdict exists.
Reliability has thus been gauged on evidence too narrow and too noisy for the long-horizon runs a judge meets in practice, 
while the judges fit for real use stay too expensive at scale.
In this work,
we take on both gaps: OSReward is, to our knowledge, the first to measure VLM-judge reliability across platforms on fresh, human-gold trajectories, and OS-Shepherd turns what it reveals into an open, low-cost reward model.

%% file: sections/3_method.tex
\section{OSReward}
\label{sec:method}

Our goal is a corpus of realistic, cross-platform CUA trajectories paired with gold verdicts
trustworthy enough to measure the judges themselves.
Reusing existing benchmarks' rollouts
cannot supply it: their runs carry quality confounds, and their labels inherit the verifiers' own noise. We therefore build the data ourselves.
This section covers the infrastructure the collection runs on, the task instructions that drive it, the collection and annotation that produce the gold set, and the resulting benchmark with its three views; the judge protocol and metrics are set up with the experiments. The OS-Shepherd reward model of \cref{sec:osshepherd} is trained on a separate corpus, labeled without human annotation by applying the findings of the judge study, and fully disjoint from this benchmark.

\subsection{Data Infrastructure}
\label{subsec:infra}

We collect fresh trajectories on cross-platform infrastructure we build and operate end-to-end, and carefully label them with humans.

\paragraph{Environment Preparation.} 
The benchmark is only as good as the environments behind
it. On desktop and mobile, we go well beyond stock benchmark images, equipping each machine with
the everyday applications a real user has and initializing every task richly: user profiles,
real files to edit, seeded app databases, distractor content. This is what makes a trajectory
worth judging: realistic state to fail from, success that must change the environment rather
than narrate completion, and concrete state for a machine-checkable verifier to inspect. Web
tasks instead run on live websites, whose realism no snapshot reproduces and which need no
initialization. Collecting everything on one infrastructure also keeps the output uniform: judge
scores stay comparable across platforms, and a reward model sees one consistent trajectory
format. We summarize each platform below and give the full implementation,
software coverage, statistics, and the action spaces of the {executing agents} in \cref{app:infra}.

\begin{figure}[htbp]
    \centering
    \includegraphics[width=\linewidth]{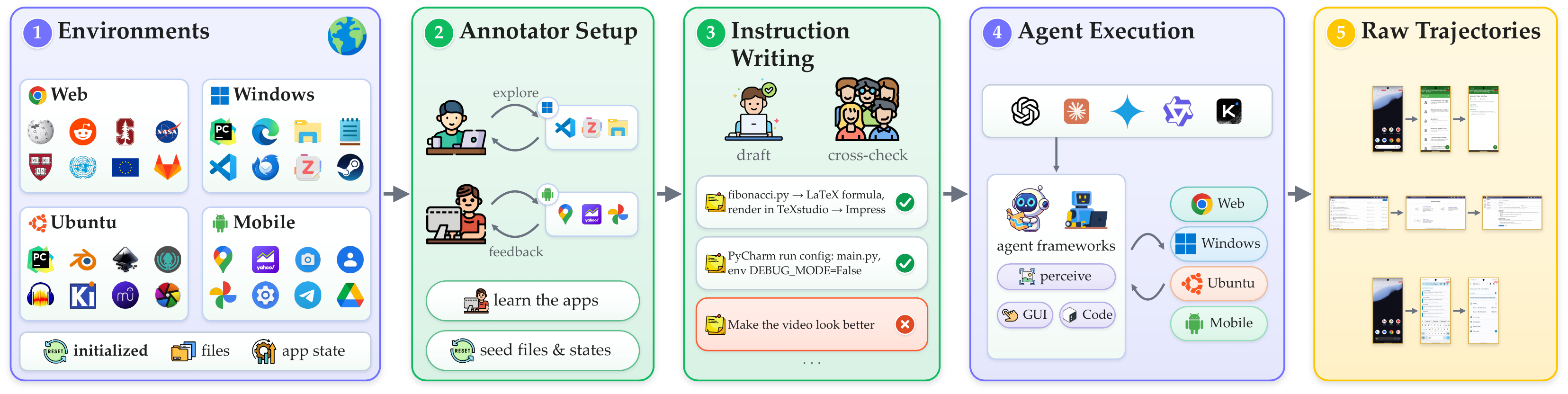}
\caption{From realistic environments to raw trajectories: annotators
prepare the environments and write grounded instructions on them, agents
from four model families execute the instructions.}
\label{fig:collection}
\end{figure}

\paragraph{Web.} We leverage a pool of Chromium-based browser workers to host many isolated sessions in parallel, one per web rollout. To mitigate blocking by websites that resist automation, the collection stack combines browser-level hardening with per-domain pacing and screens out low-quality runs afterward, supporting reliable collection at scale. Observation and action are purely vision-driven, and rollouts split between pure-GUI actions and GUI plus browser primitives, so we expect the reward model to score visual agents and API-mixing agents alike. To collect data for tasks requiring user logins, we also use several self-hosted local mirror websites (\cref{app:infra-web}).

\paragraph{Windows.} Our Windows environment is built out into a lived-in machine:
some twenty everyday applications spanning IDEs, media editors, 3D and database tools, several with accounts already logged in, plus command-line utilities such as ffmpeg. Tasks here can run long, with rollouts stretching to a hundred steps. Additionally, to enhance sample diversity, we switch between 2K and 4K resolutions during the data collection process (\cref{app:infra-windows}).

\paragraph{Ubuntu.} The Ubuntu machine comes fully initialized: about thirty applications, daily and professional, well beyond what typical environments carry; a typed pool of real files (some twenty types, each stocked with hundreds of files drawn from established datasets) so any task can start from a rich, realistic state; and per-application initialization that widens the space agents can explore. With Python and shell tooling preinstalled, rollouts split between a pure-GUI and a GUI+CLI action space,
making Ubuntu the platform where visual and code-mixed
agent styles meet (\cref{app:infra-ubuntu}).

\paragraph{Mobile.} The mobile environment is hosted in an Android emulator, and within our collection environment, we initialize the files, usage records, and databases of every application. 
The mobile environment is hosted in an Android emulator. Within our collection environment, we initialize the files, usage records, and databases of every application.
This initialization includes randomly seeded records, browsing history, and photos, along with distractor content (noise events, decoy messages, look-alike files) to expand the exploration space further,
and login-required applications come signed in with synced content (\cref{app:infra-mobile}).

\subsection{Task Instructions}
\label{subsec:instructions}
Writing instructions worth judging is a problem in its own right. An instruction must be grounded in what the environment actually offers and answerable there, and the pool deliberately includes open-ended tasks beyond what any rule-based verifier can check, because those are exactly the trajectories a judge faces in real use. Annotators write the instructions after exploring the infrastructure described above, initializing bare environments as they go (\cref{fig:collection}); they may brainstorm with AI tools or software manuals, but every draft, whatever its origin, passes a human validity check. Each candidate is then screened by annotators other than its author, a peer cross-check that removes ambiguous, ungrounded, or unanswerable instructions before any rollout effort is spent. This pipeline has no exceptions: the long cross-application tasks, the hardest to specify because one instruction must stay grounded in several applications at once, are authored and screened like every other candidate rather than synthesized. Roughly 1500 candidates are authored, and about 800 survive into next stage.

The same infrastructure also supports collection far beyond the benchmark's data scale,
extending the instruction pool to training size, the route \cref{sec:osshepherd} takes. Only the
human-vetted, peer-screened set enters the benchmark: an evaluation item must be individually
trustworthy, a bar training data need not meet. \Cref{app:infra} details each platform's
instruction method.

\subsection{Trajectory Collection and Gold Annotation}
\label{subsec:collection}

\begin{figure}[t]
    \centering
    \includegraphics[width=\linewidth]{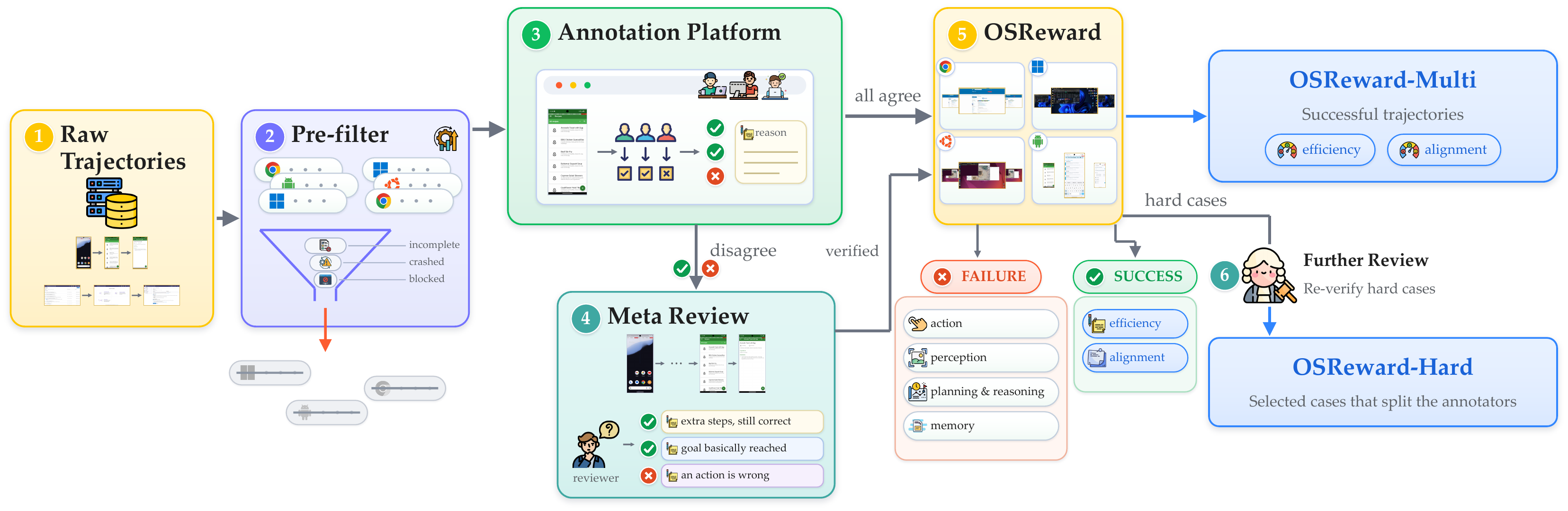}
    \caption{The annotation pipeline. Each pre-filtered trajectory is labeled by three independent annotators; disagreements go to meta review, and the verified gold set is read as three views}
    \label{fig:construction}
\end{figure}

A trajectory is a task instruction followed by a sequence of steps,
whose state (screenshot) is paired with the agent's thought and action.
OSReward is built by a pipeline that runs from the verified instruction set of \cref{subsec:instructions}, 
through execution with different agent frameworks and backbone models (\cref{fig:collection}), 
to human-labeled gold verdicts (\cref{fig:construction}).

\paragraph{Trajectories from Diverse Agent Backbones.} A judge should be measured on the
trajectories it will actually face, and those come from many different agents, not one. Each
surviving instruction is therefore rolled out by the executing agents, each driven by one of
several mainstream model backbones. Every instruction is run by one to three of them, spanning
the Claude, Gemini, Kimi, and Qwen families. Backbones differ in action idiom, thought
verbosity, and failure modes, so spreading the rollouts across the pool keeps the benchmark from
reducing to one agent's interface style. The capability spread also yields real successes and
real failures alike, and with them the labeled failures that judge evaluation otherwise lacks.
An automatic pre-filter then discards runs with severe collection problems (persistent anti-bot
blocks, network failures, frozen executions) before any human effort is spent, so a
\textsc{fail} in the benchmark reflects the agent failing the task, not the environment failing
the agent.

\paragraph{Human Annotation.} 
Gold labels bound what the study can conclude, so every surviving trajectory is labeled by human effort.
To facilitate the annotation process, we built an in-house annotation website enabling all annotators to easily review the entire trajectory.
Before judging, 
an annotator reads its full multimodal context, every screenshot, thought, and action, not only the final screen. One deliberately strict standard: an answer the agent did not obtain or verify through the environment is a \textsc{fail} even when it happens to be correct (the judging prompt carries the same rule, \cref{app:prompt}).  
A trajectory judged \textsc{success} is then scored on the two OSReward-Multi axes, \emph{alignment} (how well the actions matched the task intent; \citealp{sun2025sentinel}) and \emph{efficiency} (how free the run was of wasted steps),
each on a 3-class scale ($0$, $0.5$, $1$; scoring rubric in \cref{app:multi-rubric}). A trajectory judged \textsc{fail} is instead tagged with the errors that caused it, one or more categories from a failure taxonomy spanning reasoning-and-planning, action, perception, and memory errors (details in \cref{app:taxonomy}).

\paragraph{Consensus and Meta-Review.} 
Every trajectory that passes the pre-filter is labeled
independently by three annotators (agreement statistics in \cref{app:construction-stats}). Where
they agree unanimously, the verdict is final; the trajectories they split on escalate to a
\emph{meta-review} in which two senior reviewers examine the trajectory together and issue the
final judgment (deliberation, not a majority vote), so no gold verdict rests on a single reader.
The pass also discards trajectories with residual quality problems rather than force a label;
the full funnel, with counts at every stage, is in \cref{app:construction-stats}. All told,
three-way annotation, meta-review, and the hard-set re-verification described below cost roughly
800 human hours.

\subsection{OSReward and Its Variants}
\label{subsec:osreward}


The pipeline yields one gold set that read three ways: the full set carries the breadth across tasks and platforms, 
OSReward-Hard isolates the hard cases, and OSReward-Multi adds rating granularity; the latter two are nested subsets of the full set.

\paragraph{The Full Set (OSReward).} 

\begin{figure}[t]
    \centering
    \includegraphics[width=\linewidth]{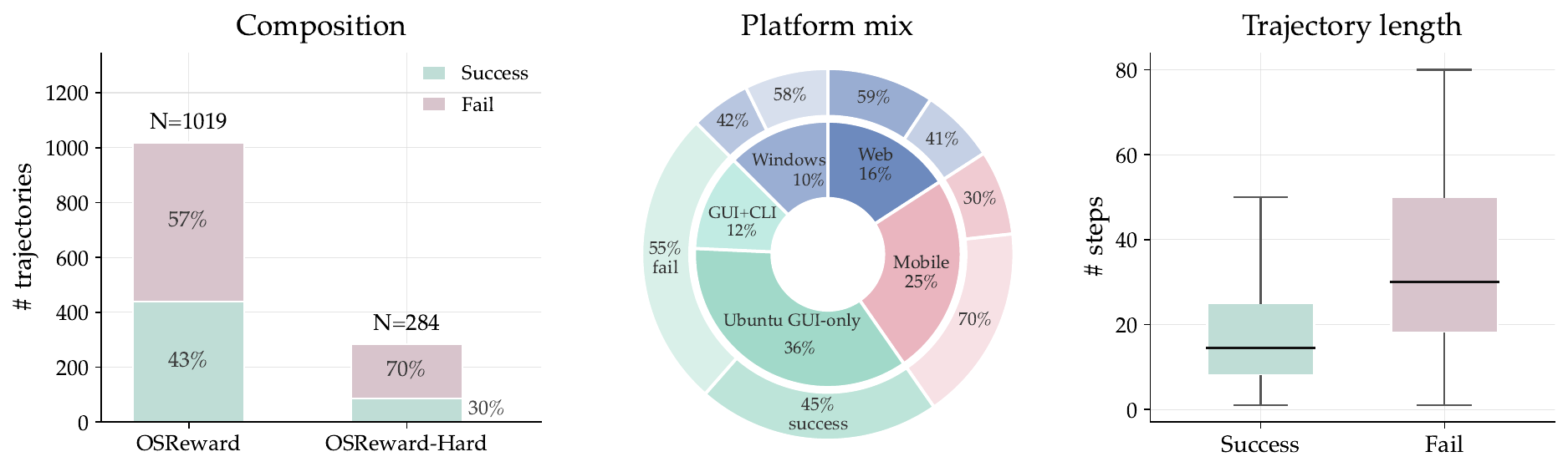}
    \caption{OSReward at a glance: outcome composition of the full and Hard
    sets, platform mix, and trajectory length; failed runs are markedly longer.}
    \label{fig:bench}
\end{figure}

The full OSReward set holds 1019 trajectories spanning all four platforms, roughly balanced between successes (43\%) and failures (57\%) (\cref{fig:bench}). 
It covers real applications and live websites, instructions from routine chores to professional workflows, and trajectories from four agent families across GUI-only and GUI+CLI action spaces, from dozen-step tasks to 100-step runs. Reasoning-and-planning errors dominate the failures, tagged on 86\% of failed runs (multi-label profile in \cref{fig:bench-failtypes}; taxonomy in \cref{app:taxonomy}). 
Failed runs are also markedly longer than successful ones, and length itself makes verification harder.

\paragraph{OSReward-Hard.} A {challenge subset} of 284 trajectories concentrating genuinely hard cases, drawn mostly from the trajectories the annotators themselves split on.
Every candidate is re-verified under a further review process, especially for the fail cases. 
Their records read like those of completed tasks or successful tasks completed through convoluted paths, and both human annotators and judge models often fail or diverge on them.
Beyond being harder than the full set, 
the selection deliberately raises the share of failures (a 30/70 \textsc{success}/\textsc{fail} split) to further expose how easily existing judges are fooled by false successes. 
Its role is diagnostic: these runs briefly fooled even trained annotators, and together they expose the error modes judges share and
how much room remains above the best current judges.

\paragraph{OSReward-Multi.} Beyond the binary verdict, 
the 440 successful trajectories carry human alignment and efficiency sub-labels (failed trajectories have neither). This is a secondary axis: it grades how well a successful run did its job, which no existing CUA-judge benchmark measures. 
We release it as an exploratory track and report it alongside the binary verdict throughout.
Alignment enters the benchmark with two levels ($0.5$ and $1.0$);
efficiency has three (0, 0.5, and 1.0), with detailed definitions in
with detailed definitions in \cref{app:multi-rubric}.
On both axes, we report macro-recall, not a raw average, so the skew cannot inflate the score (\cref{subsec:protocol}).


%% file: sections/4_experiments.tex
\section{Benchmarking VLMs as Judges}
\label{sec:experiments}

We benchmark 27 VLMs as judges on OSReward. Under one fixed setting we score every model, then
probe where the reward signal holds and where it breaks through targeted analyses and ablations.

\subsection{Experimental Setup}
\label{subsec:protocol}

\paragraph{Model Selection.} We evaluate 27 VLM judges across model series from
OpenAI~\citep{hurst2024gpt, openai2025gpt5, openai2026gpt55, openai2026gpt54thinking},
Anthropic~\citep{anthropic2026opus48, anthropic2026opus46, anthropic2026sonnet46,
anthropic2025haiku45}, Gemini~\citep{comanici2025gemini, gemini2025gemini3},
Qwen~\citep{bai2025qwen3, qwen3_5}, Doubao~\citep{bytedance2026seed2},
Kimi~\citep{team2026kimi}, and Intern~\citep{zou2026interns1pro}. The roster covers closed
frontier judges, closed efficient tiers, large open-weight models, and small open VL models;
several models are additionally run in a thinking variant. Per-model details and configs are in
\cref{app:models}.

\paragraph{Judging Protocol.} 
Every judge runs under the common protocol:
it reads the trajectory's last $N$ states with the per-step reasoning and action text and returns a verdict, with no task-specific harness, tool access, or step-level supervision, 
and $N=5$ by default. The judge outputs a \textsc{success}/\textsc{fail} verdict and, 
on OSReward-Multi trajectories, alignment and efficiency ratings.  
The remaining settings (frame count,  the red click marker~\citep{yang2023setofmark}, greedy decoding) 
are held fixed here and perturbed one at a time in the analyses and ablations that follow, each defined where it is used.

\paragraph{Metrics.} We report binary accuracy, decomposed into success recall ($\mathrm{sRec}$,
the share of truly successful trajectories accepted; low means too strict) and fail recall
($\mathrm{fRec}$, the share of truly-failed trajectories caught; low means too lenient);
balanced accuracy ($\mathrm{BalAcc}$) is their mean, which the 43/57 class mix cannot inflate.
The full judging prompt is given in \cref{app:prompt}.

\subsection{The Accuracy Ceiling}
\label{subsec:leaderboard}

\paragraph{The Performance Ceiling.} 

For a reward model to support training, around 90\% binary accuracy is a
common working bar. 
On OSReward only the frontier approaches it (Table~\ref{tab:leaderboard}): Claude-Opus-4-8 comes closest at 89.7\%, GPT-5.5 and Claude-Opus-4-6 sit just behind, and the field spans twenty points down to the small open VL models. Nor is the order at the top stable: the lead changes hands once the metric moves to balanced accuracy.
The table identifies a top tier, not a best judge.

\paragraph{The Open--Closed Gap.}
Closed judges lead open-weight ones across almost the whole field, and the
gap is widest against the small open VL models: the larger open-weight
judges, \textit{e.g.}, Kimi-K2.5 and Qwen3.5-397B-A17B, narrow it to within 4\,pp
of the lead (a fuller comparison in \cref{app:openclosed}).



\takeaway{Only frontier judges approach the accuracy a training-time
reward needs, 
and even the largest open-weight counterparts still fall short of them,
which leaves the field without a judge that is both reliable and open.}

\begin{table}[t]
    \centering
    \small
    \caption{Main-setting results for the reference judges and OS-Shepherd on OSReward and OSReward-Hard along with their access status, sorted by full-set accuracy. }
    \label{tab:leaderboard}
    \input{tables/tab_leaderboard}
    \par\vspace{2pt}
\end{table}

\subsection{The Leniency Bias}
\label{subsec:bias}

\begin{figure}[t]
    \centering
    \includegraphics[width=\linewidth]{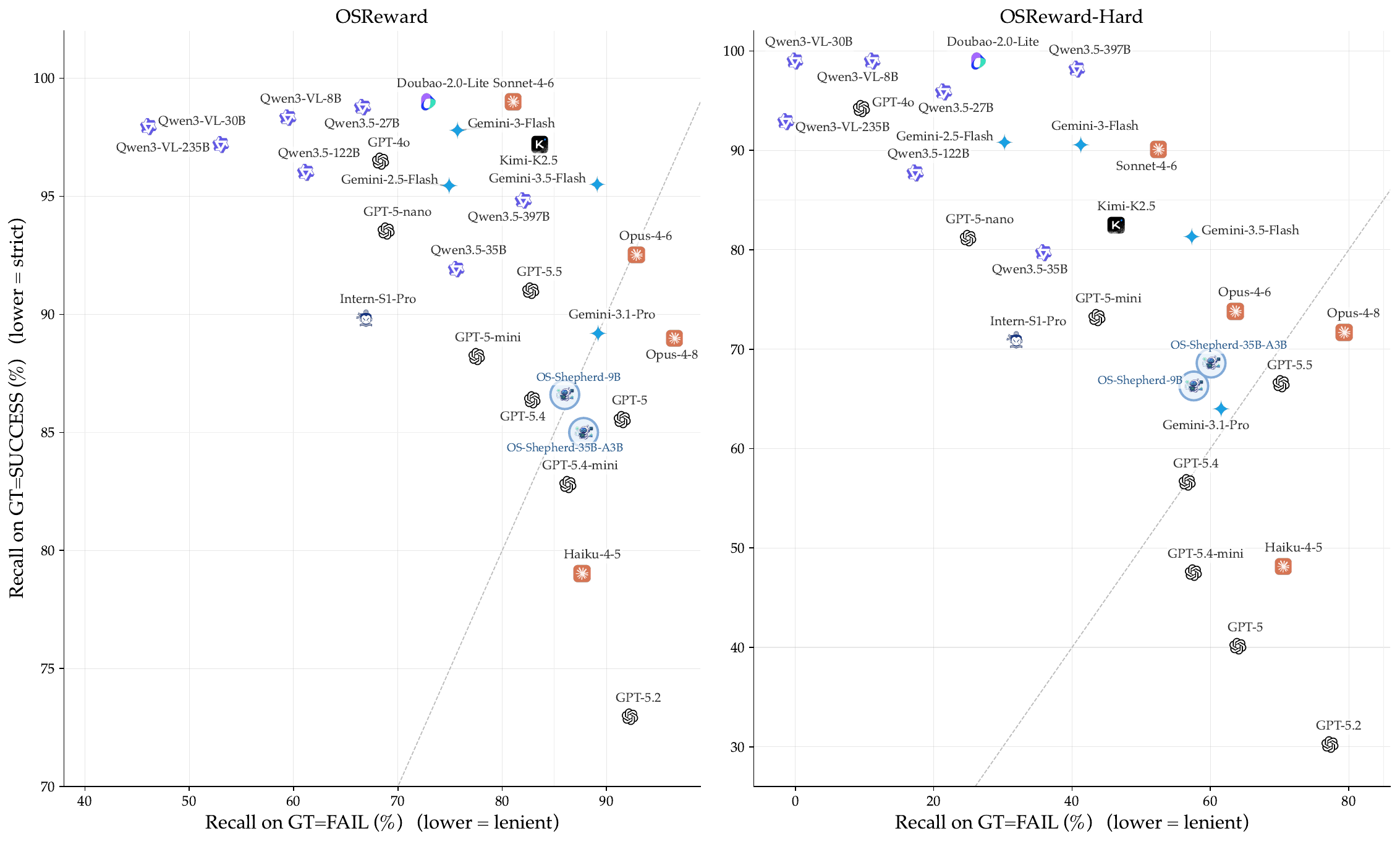}
    \caption{Judge bias on OSReward (left) and OSReward-Hard (right). Judges
below the diagonal skew lenient; most of the field sits there, and the
skew widens on the hard set.}
    \label{fig:bias}
\end{figure}

\paragraph{The Strict--Lenient Plane.} Binary accuracy is intuitive but hides \emph{how} a judge errs. Here we plot each judge at its (fRec, sRec) to make the direction of error visible (\cref{fig:bias}), 
and two clusters emerge. A large \emph{lenient} cluster has high sRec but low fRec, accepting almost everything including failures, while a smaller \emph{strict} one (\GPTtwopt, Claude-Haiku) trades success recall to catch more failures. The top judges sit near the diagonal ($\mathrm{sRec}=\mathrm{fRec}$), balanced rather than extreme, but they are few, and the field as a whole skews lenient. 
The same pattern holds on OSReward-Hard, more sharply (\cref{subsec:hard}).


\begin{figure}[t]
    \centering
    \includegraphics[width=0.9\linewidth]{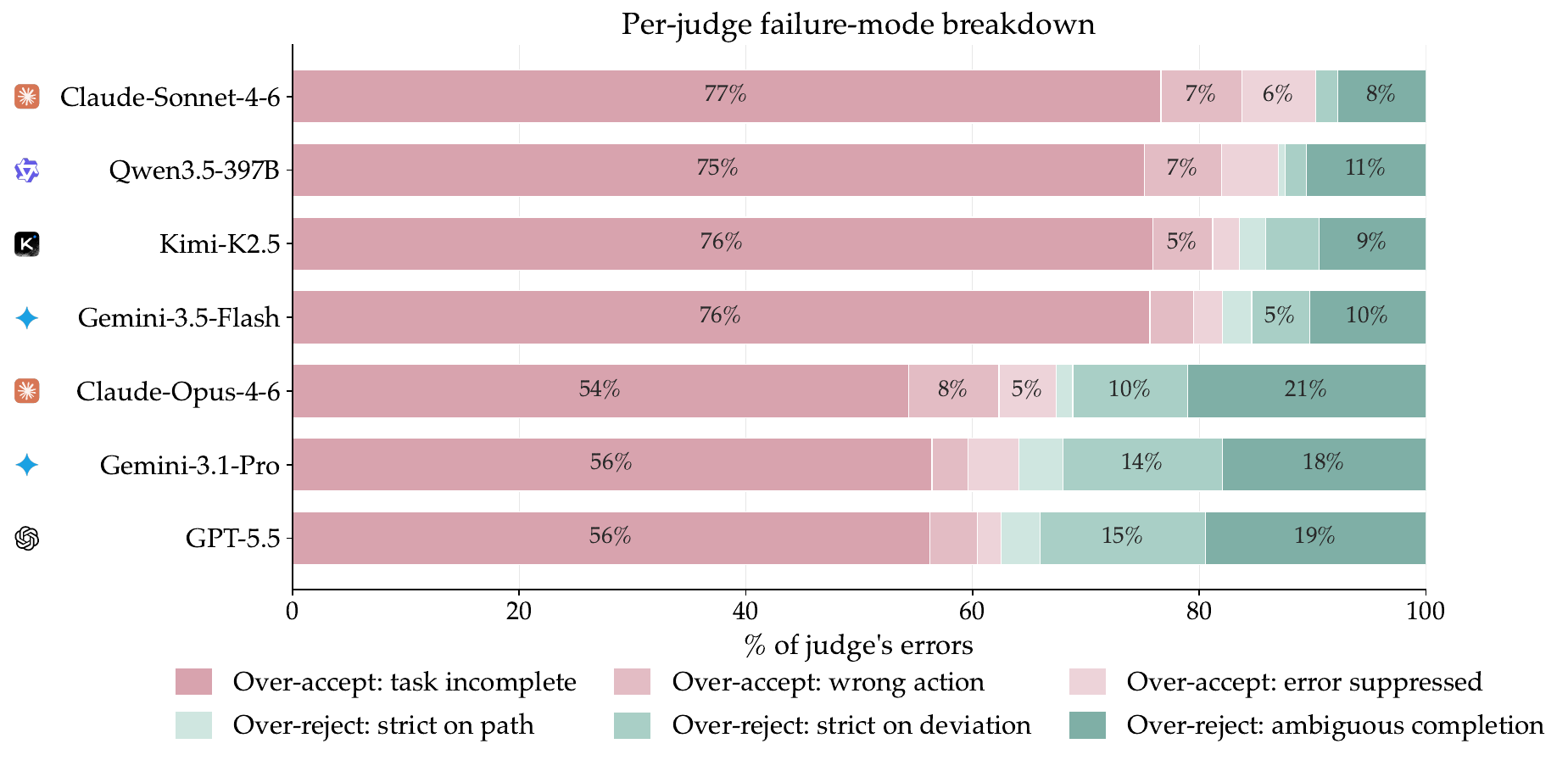}
    \caption{Per-judge error composition, seven representative judges:
     over-accepting an incomplete task (warm colors) dominates every family.}
    \label{fig:leniency}
\end{figure}

\paragraph{The Dominant Error.} Beyond how often a judge is wrong, the
more actionable question is what it gets wrong. We sort every incorrect
verdict into \emph{over-accepts} and \emph{over-rejects}, each with three
finer modes, using a strong VLM labeler with human re-checks
(\cref{fig:leniency}). One mode dominates. Over-accepting an incomplete
task makes up two-thirds of all errors, and every family shares the bias,
as this is the leading error mode of every single judge, at no less than
48\% of its mistakes. Pooled across judges, over-accepts outnumber
over-rejects three to one, though the ratio narrows to about two to one
for the strongest judges. The mechanism becomes clear in
\cref{subsec:input}, where verdicts turn out to lean on the agent's own
narrative far more than on the screenshots, so a failed run that closes
with a success claim is exactly what judges miss. Since one mode is this
dominant, prompting the judge to verify completion explicitly may help
more than the ensembling we evaluate in \cref{subsec:modelside}; we leave
this to future work.

\takeaway{Every judge family shares one dominant failure, accepting an
incomplete task as a success, so de-biasing that lenient mode is the
highest-value intervention.}

\subsection{OSReward-Hard: The Challenge Set}
\label{subsec:hard}

\begin{figure}[t]
    \centering
    \includegraphics[width=0.95\linewidth]{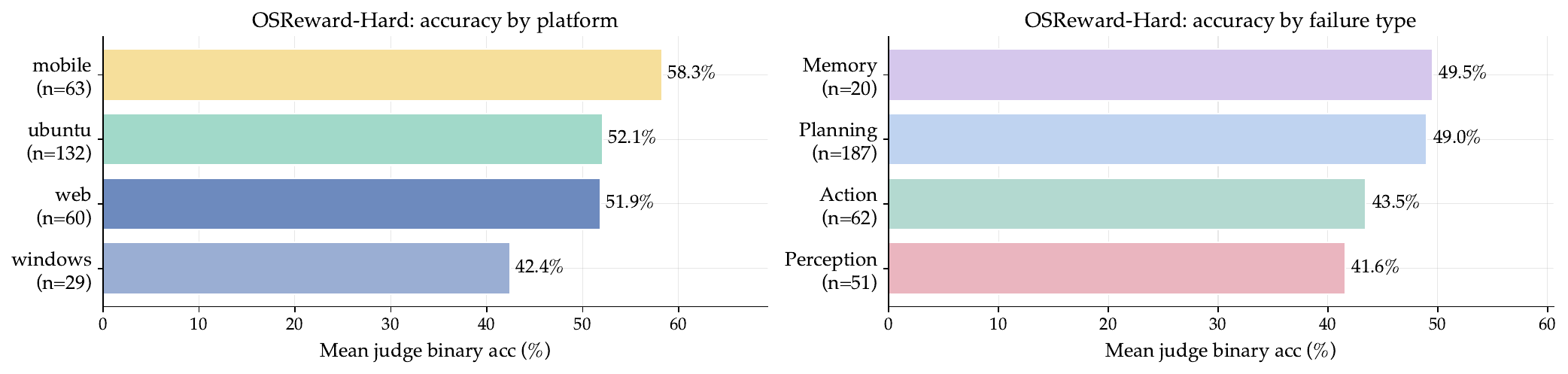}
    \caption{Mean per-judge binary accuracy on OSReward-Hard, by platform and by
    failure type. Failure types are multi-label, so their counts sum to more
    than the number of failed trajectories.}
    \label{fig:hard-breakdown}
\end{figure}

\paragraph{The Collapse.} The aggregate $\sim$90\% is optimistic. OSReward-Hard, the
failure-heavy challenge set of \cref{subsec:osreward} (30/70 \textsc{success}/\textsc{fail};
selection details in \cref{app:results-breakdown}), drops every judge by 20--43\,pp
(\cref{tab:leaderboard}; \cref{fig:teaser-hard}). Even the best judge loses a full twenty points
and lands at 69.7\%, level with what a constant always-\textsc{fail} judge scores on this 30/70
split; the mean judge falls to 52\%, and the lenient tail lower still. Raw accuracy is therefore
a trap here: the hard-set columns of \cref{tab:leaderboard} carry the recalls and balanced
accuracy, where such a constant predictor falls to 50\%. The drop is not a uniform shift either:
the broad ordering is preserved but stretched, and the mid-pack reshuffles, with judges like
\GPTfourpt and \GPTtwopt climbing several ranks on the hard set. The collapse is also
\emph{structured} (\cref{fig:hard-breakdown}): Windows is the hardest platform to judge and
mobile the easiest, and the failure types that hinge on reading the screen (perception, then
action) are markedly harder to catch than planning-and-reasoning failures, the dominant type in
the data, which are legible in the thought and action text.

Many of these hard failures are \emph{false successes}: the agent's closing text claims a
completion it never reached, exactly what a judge that leans on the narrative
(\cref{subsec:bias}) fails to catch.

\paragraph{Leniency Widens.} 
The leniency bias does not just persist on OSReward-Hard, it
\emph{widens} (\cref{fig:bias}, right; \cref{tab:leaderboard}). Fail recall now spreads from
near zero to 77\%: the lenient cluster (\textit{e.g.}, \QwenVLthirty, \InternOne) accepts almost every hard failure,
the best judges (\OpusEight, \GPTss) stay balanced but only at $\sim$70\% recall on each side, and a few judges cross into the strict half.

\paragraph{Observations from the Hard Set.}

OSReward-Hard concentrates on the cases the shared lenient mode of
\cref{subsec:bias} misses, 
failed runs whose records read like completed tasks. 
On these deceptive cases, the leading judges finally separate, so we
use the hard set as the primary diagnostic in the rest of the paper. Read
together with the failure-type analysis above, it also locates the
headroom, which lies in reading the screen and verifying completion rather
than in high-level reasoning. On this harder footing only \OpusEight and
\GPTss remain both balanced and accurate, near 70\% recall on each side,
while the few other judges near the diagonal hold their balance only by
giving up accuracy on both axes (\cref{fig:bias}). A judge that is
accurate without being too strict or too lenient barely exists yet.



\subsection{OSReward-Multi: Fine-Grained Grading}
\label{subsec:multi}

A binary verdict says \emph{whether} a run succeeded; a reward signal is more useful when it can also grade how well it was done. 
On the OSReward-Multi trajectories,
every judge additionally rates alignment and efficiency (\cref{subsec:osreward}); 
we score each axis two ways (\cref{tab:multi}): \emph{macro-recall},
whether the emitted levels are usable as they stand, and a threshold-free \emph{AUC}, 
whether the judge can rank them at all.

\needspace{14\baselineskip}
\begin{wraptable}[14]{r}{0.52\linewidth}
    \centering
    \small
    \caption{Strong judges on OSReward-Multi (\%), sorted by AUC; best per column
    in bold.}
    \label{tab:multi}
    \input{tables/tab_multi}
\end{wraptable}

Three findings stand out. Quality grading is far weaker than outcome judging: even the best
judge falls from $\sim$90\% on the binary verdict to the low sixties. The AUC--macro-recall gap
is the diagnostic: judges rank the levels better than they score them, so the discrimination is
there but the emitted thresholds are miscalibrated. And the weakness concentrates on
\emph{alignment}, where judges default to the top rating almost regardless of the run, 
because whether the actions served the instruction is harder to read off the record than whether steps
were wasted. 
Our own OS-Shepherd models inherit this: the 35B recovers part of the gap while the
9B stays near the constant-level baseline. 
This keeps the binary verdict our primary metric and
the multi-axis view a complementary probe. Full metric definitions and per-axis results are in
\cref{app:multi}.

{\setlength{\linewidth}{\columnwidth}%
\takeaway{No judge grades quality well: judges generally rank the alignment and efficiency
levels better than they score them, so the discrimination exists but the emitted ratings are
miscalibrated, worst on alignment.}}

%% file: tables/tab_leaderboard.tex
\setlength{\tabcolsep}{3.6pt}
\begin{tabular}{l l c c c c c c c c}
    \toprule
    \multirow{2}{*}{Judge} & \multirow{2}{*}{Access} & \multicolumn{4}{c}{OSReward} & \multicolumn{4}{c}{OSReward-Hard} \\
    \cmidrule(lr){3-6} \cmidrule(lr){7-10}
    & & Acc & sRec & fRec & BalAcc & Acc & sRec & fRec & BalAcc \\
    \midrule
    \tablogo{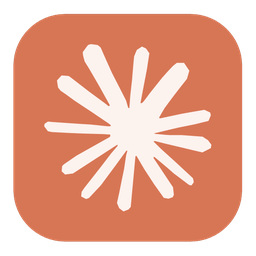}Claude-Opus-4-8 & closed & \textbf{89.7} & 91.1 & 88.9 & 90.0 & \textbf{69.7} & 69.8 & 69.7 & \textbf{69.7} \\
    \tablogo{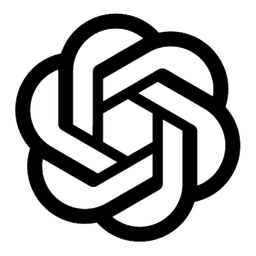}GPT-5.5 & closed & 89.5 & 91.8 & 87.8 & 89.8 & 67.3 & 66.3 & 67.7 & 67.0 \\
    \tablogo{Claude}Claude-Opus-4-6 & closed & 89.5 & 92.7 & 87.7 & \textbf{90.2} & 67.3 & 72.1 & 65.2 & 68.6 \\
    \tablogo{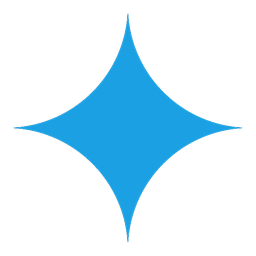}Gemini-3.1-Pro & closed & 87.9 & 90.2 & 86.2 & 88.2 & 61.6 & 61.6 & 61.6 & 61.6 \\
    \tablogo{Gemini}Gemini-3.5-Flash & closed & 87.8 & 95.7 & 81.8 & 88.8 & 59.5 & 81.4 & 50.0 & 65.7 \\
    \tablogo{Claude}Claude-Sonnet-4-6 & closed & 87.7 & 97.5 & 80.3 & 88.9 & 59.2 & 90.7 & 45.5 & 68.1 \\
    \tablogo{GPT}GPT-5 & closed & 87.4 & 86.8 & 87.9 & 87.4 & 58.1 & 43.0 & 64.6 & 53.8 \\
    \tablogo{GPT}GPT-5.4 & closed & 87.1 & 87.3 & 87.0 & 87.1 & 63.0 & 62.8 & 63.1 & 63.0 \\
    \tablogo{Gemini}Gemini-3-Flash & closed & 87.0 & 96.6 & 79.8 & 88.2 & 57.0 & 86.0 & 44.4 & 65.2 \\
    \tablogo{GPT}GPT-5-mini & closed & 86.1 & 93.8 & 80.2 & 87.0 & 56.3 & 79.1 & 46.5 & 62.8 \\
    \tablogo{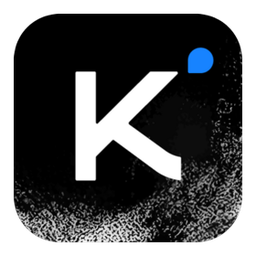}Kimi-K2.5 & open weights & 85.9 & 95.5 & 79.2 & 87.3 & 54.8 & 83.7 & 42.1 & 62.9 \\
    \tablogo{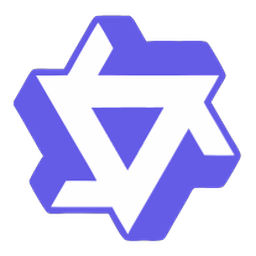}Qwen3.5-397B-A17B & open weights & 85.8 & 95.2 & 78.6 & 86.9 & 58.5 & 91.9 & 43.9 & 67.9 \\
    \tablogo{GPT}GPT-5.4-mini & closed & 85.2 & 82.5 & 87.2 & 84.9 & 58.1 & 48.2 & 62.4 & 55.3 \\
    \tablogo{Claude}Claude-Haiku-4-5 & closed & 84.5 & 80.9 & 87.2 & 84.0 & 59.5 & 47.7 & 64.6 & 56.2 \\
    \tablogo{GPT}GPT-5.2 & closed & 83.9 & 73.0 & 92.2 & 82.6 & 63.0 & 30.2 & 77.3 & 53.8 \\
    \tablogo{Gemini}Gemini-2.5-Flash & closed & 83.3 & 95.5 & 74.0 & 84.8 & 48.9 & 90.7 & 30.8 & 60.8 \\
    \tablogo{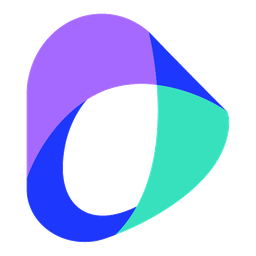}Doubao-2.0-Lite & closed & 83.3 & 98.5 & 72.1 & 85.3 & 45.5 & 96.1 & 24.3 & 60.2 \\
    \tablogo{GPT}GPT-5-nano & closed & 82.3 & 97.0 & 71.1 & 84.1 & 45.4 & 95.3 & 23.7 & 59.5 \\
    \tablogo{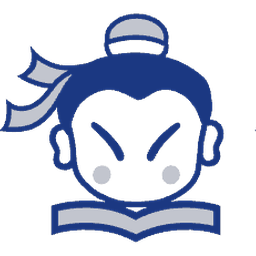}Intern-S1-Pro & open weights & 82.3 & 92.3 & 74.7 & 83.5 & 43.7 & 70.9 & 31.8 & 51.4 \\
    \tablogo{Qwen}Qwen3.5-35B-A3B & open weights & 82.2 & 92.4 & 74.5 & 83.5 & 51.1 & 83.7 & 36.9 & 60.3 \\
    \tablogo{Qwen}Qwen3.5-27B & open weights & 82.0 & 97.4 & 70.5 & 84.0 & 44.2 & 92.9 & 23.2 & 58.0 \\
    \tablogo{GPT}GPT-4o & closed & 81.0 & 96.8 & 69.0 & 82.9 & 39.4 & 90.7 & 17.2 & 53.9 \\
    \tablogo{InternLM}Intern-S2-Preview & open weights & 80.6 & 98.4 & 66.9 & 82.7 & 40.3 & 94.2 & 16.8 & 55.5 \\
    \tablogo{Qwen}Qwen3.5-122B-A10B & open weights & 79.6 & 96.8 & 66.4 & 81.6 & 39.4 & 89.5 & 17.7 & 53.6 \\
    \tablogo{Qwen}Qwen3-VL-8B & open weights & 77.1 & 99.8 & 59.9 & 79.8 & 36.2 & 100.0 & 8.2 & 54.1 \\
    \tablogo{Qwen}Qwen3-VL-235B & open weights & 74.0 & 99.1 & 54.9 & 77.0 & 31.4 & 97.7 & 2.5 & 50.1 \\
    \tablogo{Qwen}Qwen3-VL-30B & open weights & 69.4 & 99.8 & 46.3 & 73.0 & 31.1 & 98.8 & 1.5 & 50.2 \\
    \midrule
    \tablogo{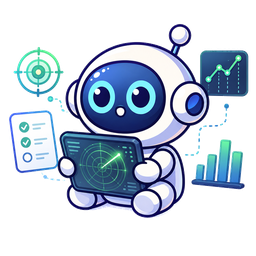}OS-Shepherd-9B (ours) & open weights\,+\,data & 86.1 & 86.6 & 86.0 & 86.3 & 60.2 & 66.3 & 57.6 & 61.9 \\
    \tablogo{OS-Shepherd}OS-Shepherd-35B-A3B (ours) & open weights\,+\,data & 85.6 & 85.0 & 86.2 & 85.6 & 62.7 & 68.6 & 60.1 & 64.3 \\
    \bottomrule
\end{tabular}

%% file: tables/tab_multi.tex
\setlength{\tabcolsep}{3.8pt}
\begin{tabular}{l ccc c}
    \toprule
    \multirow{2}{*}{Judge} & \multicolumn{3}{c}{Macro-recall} & \multirow{2}{*}{AUC} \\
    \cmidrule(lr){2-4}
    & Align & Effic & Multi & \\
    \midrule
    GPT-5.5               & \textbf{58.7} & 68.2 & \textbf{63.5} & \textbf{66.7} \\
    Claude-Opus-4-8       & 52.9 & 68.7 & 60.8 & 65.6 \\
    Claude-Sonnet-4-6     & 53.2 & 62.6 & 57.9 & 61.9 \\
    Gemini-3.5-Flash      & 47.6 & \textbf{71.4} & 59.5 & 60.8 \\
    \modelb{} (ours)      & 47.7 & 65.8 & 56.8 & 60.7 \\
    \OSShepNineB{} (ours) & 44.1 & 54.0 & 49.0 & 58.5 \\
    Gemini-3-Flash        & 50.6 & 61.5 & 56.0 & 55.8 \\
    \bottomrule
\end{tabular}

%% file: sections/5_analysis.tex
\section{Analysis}
\label{sec:analysis}

The benchmark measured how well judges perform; this section analyzes what drives a verdict. We perturb one component at a time and observe how the verdict changes, across three groups: the visual input, the text input, and the way the model is run. We close with the cost of reliability.

\subsection{Visual Inputs Barely Move the Verdict}
\label{subsec:visual}

We first change what the judge \emph{sees}. Swapping the five trailing
screenshots for the last three, or for the first plus the last two, moves
binary accuracy by less than half a point, and removing the red click
marker does not hurt at all. Sweeping the trailing-screenshot count from
one to sixteen is just as uneventful: each judge wanders two to three
points with no trend, and every judge sits near its own best at
$N=5$--$9$ (\cref{app:robust-extra}). The visual settings that look
important when writing a judging prompt turn out to matter little for
aggregate accuracy.

Aggregate accuracy does not capture everything, though. Each of these settings, harmless in aggregate, still flips 5–7\% of individual verdicts relative to the main setting. Such flips average out in evaluation, but not in reward labeling, where each trajectory's label is consumed on its own.
\Cref{subsec:modelside} gives the noise floor these numbers should be read against.

\subsection{The Text History Matters}
\label{subsec:input}

\begin{figure}[t]
    \centering
    \includegraphics[width=\linewidth]{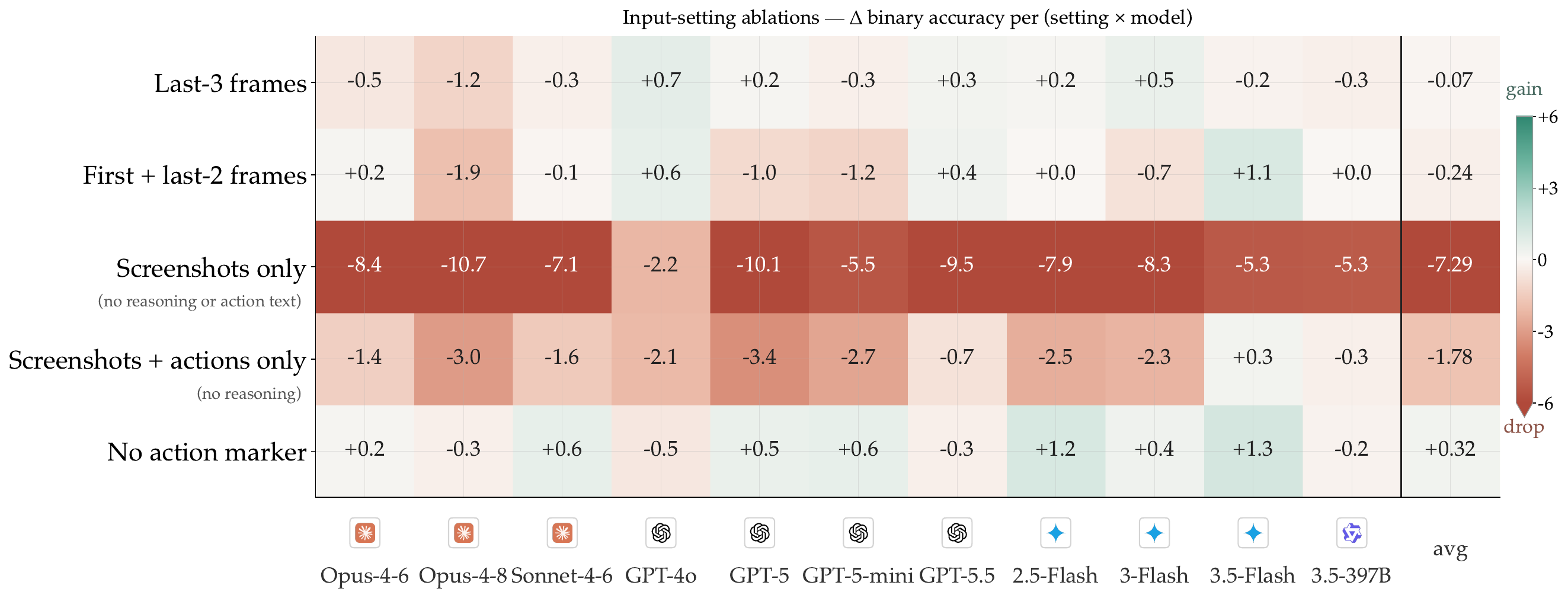}
    \caption{Overview of input ablations: $\Delta$ binary accuracy per (setting $\times$ model)
    vs.\ the main setting.}
    \label{fig:ablation}
\end{figure}

One ablation dominates the rest (\cref{fig:ablation}). 
Dropping the per-step thought and action text costs $7.2$\,pp on average, and several times that on the web, where typed strings carry intent that the last few screenshots cannot encode. It also flips 22.7\% of verdicts, three times what any visual change does. Dropping only the chain-of-thought while keeping the actions is far milder ($1.8$\,pp, 11.6\% flipped): 
within the text, the agent's actions carry roughly more signal of its stated reasoning.

This is the mechanism behind the leniency bias of \cref{subsec:bias}. A judge that leans on the
agent's own narrative is exactly a judge that a confident closing claim can fool. 
This suggests a recipe for CUA reward-model labeling: keep the full text history, drop the marker, and set the screenshot count per model.
\takeaway{Even with screenshots in view, the text history drives the verdict; the visual cues
that look essential barely matter.}

\subsection{Running the Judge Differently Does Not Fix It}
\label{subsec:modelside}

If the input is fixed, can the model be run differently to get a better verdict? 
Letting a judge think harder helps, but the gain shrinks as the judge gets stronger.
Comparing each model against a stronger thinking setting of itself gains a few points on the weakest judge and almost nothing at the frontier; 
raising GPT-5.5's reasoning effort yields monotone but similarly small improvements (\cref{app:robust-extra}).
Extra deliberation recovers what a weak judge loses by under-thinking; it does not improve the best judges.

Decoding noise provides a reference scale for the flip rates.
Re-running the \emph{same} judge on
the \emph{same} input at $T{=}0.7$ already flips 6--9\% of verdicts, so the churn from the
visual settings of \cref{subsec:visual} sits at or below the model's own sampling noise, while
the text ablation's is far outside it. Aggregate accuracy is stable under temperature;
individual labels are not, and that is the quantity a reward model consumes.

Nor does adding judges help. Judges agree far more than they disagree (pairwise Cohen's
$\kappa\approx0.71$ among top judges) and herd on the
\emph{same} hard trajectories, 
so a wider pool mostly adds copies of the same mistakes: a top-3
majority vote edges the best single judge by about a point at several times the cost. 
The oracle is more informative: accepting any pooled judge's correct verdict reaches 99\% accuracy.
The pool almost always contains a correct verdict, but a vote cannot identify which judge to trust on which trajectory.
This argues for soft-label, confidence-weighted reward modeling over hard votes, which we leave to future work. Full per-model ablations and further detail are in \cref{app:analysis}.
Later we will leverage this finding to select trajectories for large-scale training (\cref{subsec:osshepherd-corpus}).

\takeaway{No model-side knob buys reliability: extra thinking helps least exactly where judges
are already strong, resampling alone flips 6--9\% of labels, and judges herd, so ensembling
cannot substitute for a better judge.}

\subsection{The Cost of a Reliable Verdict}
\label{subsec:cost}

The remaining option is to pay for a stronger judge, and here cost becomes the binding constraint.
Accuracy and cost
are in direct tension, sharpest on OSReward-Hard (\cref{fig:osshep-pareto}): \OpusEight holds
69.7\% there at $\sim$\$100 to judge the full set\footnote{Costs use official API list prices
where available; open-weight models without official pricing are priced at market rates (May
2026) for similar-size models.} and \GPTss 67.3\% at \$45, while the best sub-\$3 judge drops to
57.0\% and the cheapest tier falls below 50\%. 
The same trade-off looks benign on the full set, about 3\,pp for a $42\times$ price cut (\cref{app:results-breakdown}), but not on OSReward-Hard.
At the millions of calls an RL run issues, even the frontier judges are out of reach. 
Luckily, the open reward model presented below \cref{sec:osshepherd} will close this gap, 
matching a strong judge at a cost that scales to training.
\takeaway{Reliability can be bought, but not afforded: the judges that hold $\sim$70\% on
OSReward-Hard cost \$45--100 to judge the full set once, which does not scale to training-time
reward.}

%% file: sections/6_osshepherd.tex
\section{OS-Shepherd: An Open Reward Model}
\label{sec:osshepherd}

The study so far leaves the field in a bind: reliable judging is unaffordable at the call
volumes rejection sampling, RL, and trajectory mining need, and the affordable judges are weak
or systematically lenient. Open data is scarce as well, especially the grounded \textsc{fail}
trajectories a reward model must learn to reject. 
We therefore build OS-Shepherd: an open-weight \& open-data reward model trained on a corpus we release, aimed at alleviating the false-success problem, and cheap enough to run at training scale.

\subsection{OS-Shepherd-100K}
\label{subsec:osshepherd-corpus}

Training a reward model needs labels at a scale human annotation cannot reach on an academic budget: the benchmark's 1019 gold trajectories already cost roughly 800 human hours (\cref{subsec:collection}), and training needs two orders of magnitude more.
VLM judges are the only affordable annotators, and \cref{sec:experiments} has just measured how far any single one can be trusted. The ensembling findings of the analysis show the way out. 
Voting cannot sharpen a verdict, because judges herd (\cref{subsec:modelside}). Corpus building is free to discard trajectories unlike evaluation, so we select by agreement: a trajectory enters \oudata{} only when diverse strong judges independently reach the same verdict, and the retained labels prove reliable in practice.
Labeling then runs automatically at scale, on our own collection infrastructure and extended instruction pool. 
The corpus is also built for diversity, so the model learns task success rather than any one agent's style. The rollouts span five model families (Claude, Gemini, GPT, Kimi, Qwen) under varying harnesses, action spaces, and step budgets, joined by independent open-source agent stacks.
The supervision a sample carries is its verified task outcome, and the corpus is checked to be free of contamination from the evaluation trajectories (\cref{app:corpus-stats}).

\begin{figure}[t]
    \centering
    \includegraphics[width=\linewidth]{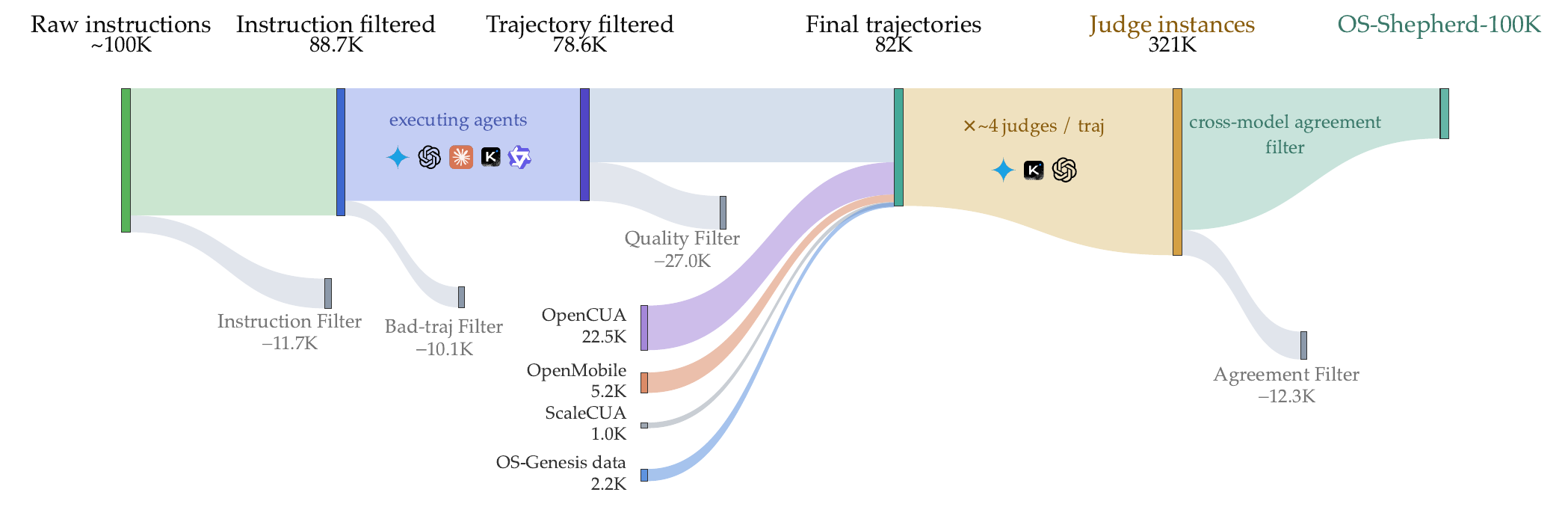}
    \caption{The \oudata{} pipeline: self-collected and open-source trajectories
    are ensemble-judged and distilled into the training set. }
    \label{fig:corpus-pipeline}
\end{figure}

\paragraph{From Instructions to Annotated Trajectories.} The pipeline indicated below in \cref{fig:corpus-pipeline} filters the self-collected instructions, executing them via agents within our infrastructure. We then join the surviving trajectories with selected open-source CUA trajectories~\citep{sun2025osgenesis, wang2026opencua, cheng2026openmobile, liu2026scalecua}. Each kept trajectory is scored by an ensemble of strong VLM-as-a-Judge runs under the evaluation protocol we built, varying the number of input screenshots to promote diversity and generalization.
Aggregating these votes and keeping only high-agreement trajectories yields the \oudata{} training set (counts in \cref{fig:corpus-pipeline}; full per-source, per-setting, and length statistics in \cref{app:corpus-stats}). In the figure's units, a \emph{judge instance} is one judge's verdict on one trajectory, and a \emph{sample} is one retained trajectory--response pair; the agreement filter keeps about 85\% of the judged trajectories. Crucially, every sample carries the judge's \emph{reasoning}, not just a binary verdict (the natural target, given that \cref{subsec:input} finds the verdict lives mainly in text), making \oudata{} the first large-scale reasoning-annotated judge corpus for CUA trajectories.

\paragraph{Coverage.}
\begin{wraptable}[13]{r}{0.44\linewidth}
    \centering
    \small
    \setlength{\tabcolsep}{5pt}
    \caption{\oudata{} judge-instance pool by platform.}
    \label{tab:mixture}
    \begin{tabular}{l r r}
        \toprule
        Platform & Instances & Share \\
        \midrule
        Web                  & 119{,}469 & 37\% \\
        Windows              & 62{,}053  & 19\% \\
        macOS                & 45{,}028  & 14\% \\
        Ubuntu (GUI only)    & 34{,}355  & 11\% \\
        Ubuntu (GUI $+$ CLI) & 29{,}785  & 9\%  \\
        Mobile               & 30{,}941  & 10\% \\
        \midrule
        Total                & 321{,}631 & 100\% \\
        \bottomrule
    \end{tabular}
\end{wraptable}
The corpus spans Web, Windows, MacOS, Ubuntu, and Android (\cref{tab:mixture}). Each collected
trajectory is judged by as many ensemble runs as our budget allows, 
and this pool of judge instances is the raw material \oudata{} is built from.
About 46\% of the Ubuntu data interleaves GUI actions with command-line steps. 
We deliberately collected this mode, anticipating that combined GUI and CLI interaction will become more prevalent as model capability grows. 
After filtering, the retained samples split mainly between desktop and web, with a small mobile share (\cref{app:corpus-stats}).

\paragraph{Label Provenance.}
Because judges err on the same cases and share one lenient bias (\cref{subsec:modelside}), a single judge would propagate its own bias into the corpus.
We therefore
vary both the judge model and the screenshot setting and keep only strong judges, the
accuracy-matched diversity that helps. We then retain a trajectory only when its votes are
high-agreement, near-unanimous with no dissent from the strongest judges, and drop the ambiguous
middle, so no forced majority enters the corpus. The \emph{label} comes from this diverse
ensemble. The retained \emph{response}, the reasoning the model imitates, defaults to one strong
judge's, since mixing styles freely destabilized early training; when that judge did not score
the trajectory or its verdict differs from the label, we retain another agreeing judge's
response instead, so the reasoning never contradicts the label, and the occasional exposure to
other judges' reasoning aids generalization. Each trajectory contributes at most two samples,
one per output format (\cref{app:training-sft}), and we balance the final
\textsc{success}/\textsc{fail} mix.

\paragraph{Negatives and Synthesized Instructions.} 
We deliberately over-collect grounded \textsc{fail} trajectories. Because the agreement filter and the multi-judge ensemble screen out ungrounded or trivially blocked runs, the resulting negatives reflect real task failures rather than environment artifacts.
Writing and vetting every instruction by hand is affordable for a
benchmark of a thousand trajectories, but not for a corpus two orders of magnitude larger.
Collection at this scale therefore leans further on synthesized instructions, a portion written
by reverse task synthesis~\citep{sun2025osgenesis}. The synthesis concentrates on desktop, where
grounded instructions are hardest to author by hand: it accounts for about 25\% of the Ubuntu
and Windows training instructions combined and about 10\% of the web instructions.

\subsection{Training}
\label{subsec:osshepherd-training}

OS-Shepherd is trained in two stages on \oudata{} and judges under the same protocol. We start
from Qwen3.5~\citep{qwen3_5}.

\paragraph{SFT-Stage.} We fine-tune the base VLM on the \oudata{} corpus (the 96.6K
agreement-filtered samples) to judge under the main setting. SFT alone lifts \OSShepNineB far
above its \QwenNineB base, primarily by correcting the base model's near-total leniency.

\paragraph{RL-Stage.} 
What SFT leaves behind is the \emph{false success}, the harmful error for a reward signal since it will directly reinforces incorrect behavior.
The SFT model already resolves many of these cases under repeated sampling of trajectories,
so we mine them and leverage GRPO~\citep{shao2024deepseekmath} for an RL-stage, with training details are in \cref{app:training}.

\subsection{Evaluating OS-Shepherd}
\label{subsec:osshepherd-results}

\Cref{tab:osshepherd-stages} reports both trained models against the untuned checkpoints they
start from.

\begin{table}[t]
    \centering
    \small
    \captionsetup{justification=centering}
    \caption[OS-Shepherd results.]{OS-Shepherd against its untuned base, on the
    full set and OSReward-Hard.}
    \label{tab:osshepherd-stages}
    \begin{tabular}{l cccc cccc}
        \toprule
        \multirow{2}{*}{Model} & \multicolumn{4}{c}{OSReward} & \multicolumn{4}{c}{OSReward-Hard} \\
        \cmidrule(lr){2-5}\cmidrule(lr){6-9}
        & Acc & sRec & fRec & BalAcc & Acc & sRec & fRec & BalAcc \\
        \midrule
        \QwenNineB{} \small(base) & 76.7 & 98.9 & 59.9 & 79.4 & 39.4 & 97.7 & 14.1 & 55.9 \\
        \OSShepNineB         & 86.1 & 86.6 & 86.0 & 86.3 & 60.2 & 66.3 & 57.6 & 61.9 \\
        \midrule
        \QwenMidA{} \small(base)  & 82.2 & 92.4 & 74.5 & 83.5 & 51.1 & 83.7 & 36.9 & 60.3 \\
        \modelb             & 85.6 & 85.0 & 86.2 & 85.6 & 62.7 & 68.6 & 60.1 & 64.3 \\
        \bottomrule
    \end{tabular}
\end{table}

\paragraph{Where \OSShepNineB Lands.} 
Trained end-to-end, the 9B moves from the bottom third of the field into the commercial band on the full set.
What makes it usable as a reward
signal shows on OSReward-Hard: it catches 57.6\% of hard false successes, where the cheap and
lenient field misses nearly all of them, and it stays on the balanced diagonal while doing so.
That combination is rare; the only judges that match it are frontier
models at many times its cost. 
Training also buys robustness, cutting the 9B's hard-set drop by a third relative to its base. On the cost-accuracy view no other sub-\$2 judge comes near it: at \$1.36 to judge the full set, the 9B is about a third cheaper than the closest-priced commercial judge while leading it on OSReward-Hard. More details are provided in \cref{app:results-breakdown}.

\paragraph{Scale Adds Little.} \modelb{} follows the identical recipe and lands beside the 9B: a
small further gain on OSReward-Hard, level full-set accuracy, the same balanced diagonal, and
the same behavior on the three held-out CUA benchmarks (\cref{tab:osshepherd-stages};
\cref{sec:generalization}). Four times the parameters buy 2.4\,pp of hard-set balanced accuracy
and nothing on the full set; what transfers is the recipe. On the quality axes both models track
the field's weakness (\cref{subsec:multi}), the 35B grading noticeably better than the 9B.

\paragraph{Cost at Training Scale.} 
A reward model is queried throughout a training run,
so even a small per-trajectory cost accumulates.
For example,
a single modest bout of online RL
(200 updates, batch 16, 16 rollouts) already issues $200\times16\times16 = 51{,}200$ judge
calls: about \$4{,}000 with \OpusEight or \$2{,}300 with \GPTss at OSReward's average trajectory
cost, against about \$68 with \OSShepNineB, a $30$--$60\times$ reduction that compounds over the
many runs a project needs. And because \OSShepNineB is open and self-hostable, even that \$68 is
only an API-equivalent figure: in practice the marginal cost is the lab's own GPU time.

OS-Shepherd shows that reliable CUA reward no longer has to be bought at high prices: a small open judge stays level with commercial judges on the full set, leads every similarly priced judge on OSReward-Hard, and closes most of the frontier gap at 30-60$\times$ lower cost. 

\takeaway{With our data and recipe, small open judges match commercial
ones, producing reliable reward at a cost that scales.}

%% file: sections/7_generalization.tex
\section{Generalization to Existing Benchmarks}
\label{sec:generalization}

Every number so far comes from OSReward's own gold labels. To test whether these findings and OS-Shepherd's de-biasing extend beyond our data, we run the same judges on three benchmarks from prior work, each against its own human-written verifier (\cref{fig:ood}).

\subsection{Judges vs.\ Human-Written Verifiers}
\label{subsec:gen-field}

\begin{figure}[t]
    \centering
    \includegraphics[width=\linewidth]{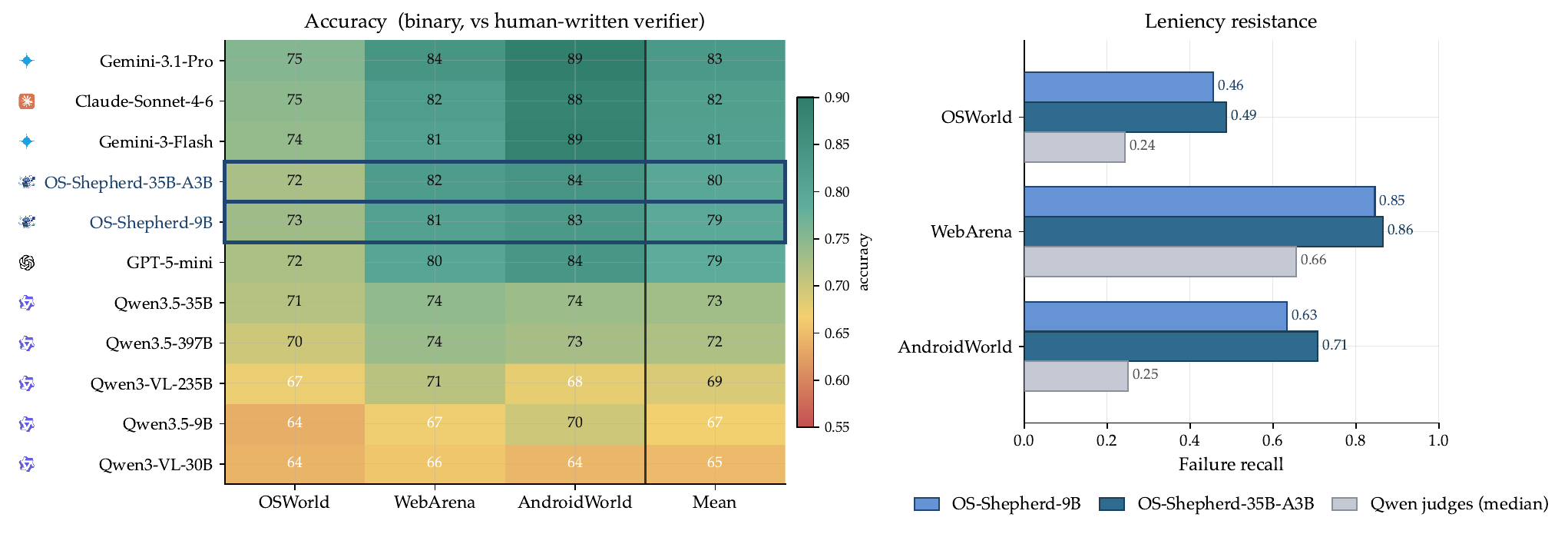}
    \caption{Judges on three existing CUA benchmarks against each benchmark's
    human-written verifier (matched subsets): accuracy and failure recall;
    means are computed before rounding.}
    \label{fig:ood}
\end{figure}

The introduction's pilot study already flagged that even the best judges disagree with existing
benchmarks' own verifiers on many desktop verdicts; we now measure that at scale. 
We take $\sim$90\% agreement with a benchmark's human-written verifier as the bar for a judge that could replace it.
Agreement varies by \emph{platform} far more than by judge (\cref{fig:ood}, left): the best judges approach that bar on mobile, come within about 6\,pp on web, and fall well short on desktop. The ordering is the same for every judge.
These verifiers are themselves imperfect~\citep{osworldverified}, so the figure measures agreement with each benchmark's verifier rather than accuracy against ground truth. A judge's true accuracy is likely somewhat higher than the agreement shown.
The desktop shortfall is the leniency bias of \cref{subsec:bias} appearing outside OSReward: judges accept failed runs as successes, and the false positives concentrate in unverifiable domains and on long trajectories (\cref{app:osworld}).

\subsection{OS-Shepherd Transfers Across Benchmarks}
\label{subsec:gen-osshep}

OS-Shepherd is trained on \oudata{} and, like every judge here, evaluated held-out on OSReward;
the question now is whether its de-biasing also transfers to benchmarks built independently,
each with its own human-written verifier. It does: the OS-Shepherd models are the best open
judges on OSWorld and AndroidWorld and sit in the frontier cluster on WebArena, beating every
general open model up to \QwenBigA ($\sim$44$\times$ the 9B's size) on all three
(\cref{fig:ood}, left). 
None of these benchmarks' tasks or trajectories enter training or appear in OSReward.

\paragraph{What Transfers Is Catching Failures.} What transfers is the de-biasing itself
(\cref{fig:ood}, right). Where successes dominate (OSWorld, AndroidWorld), \OSShepNineB catches
far more true failures than any general Qwen judge, whose median sits near ``always
\textsc{success}.'' WebArena inverts the base rate, so an over-lenient judge scores \emph{below}
the trivial all-\textsc{fail} baseline: \OSShepNineB's own untuned \QwenNineB base lands there,
while \OSShepNineB is the tightest-calibrated judge in the field, frontier included. The
resistance is therefore learned, and from training rather than size: general Qwen judges stay
lenient at every size from 30B to 397B, while \modelb{}, four times the parameters on the
identical recipe, lands essentially beside the 9B. What training buys is a relocated operating
point: more failure recall where false successes dominate, a little less success recognition
elsewhere, at roughly flat balanced accuracy. Because \cref{fig:ood} reports the final models,
these results establish transfer of the full recipe rather than isolating the RL stage.

\paragraph{Toward a Scalable Verifier.} Each benchmark's human-written verifier is costly to
author and does not generalize past its own tasks, exactly the non-scalability a learned judge
is meant to fix. A single 9B, trained on none of them, carries its de-biasing across all three
at a cost that scales to training-time reward (\cref{subsec:osshepherd-results}). It does not
yet match frontier accuracy, but closes most of the gap at a fraction of the scale, the
direction a scalable CUA reward signal must move.

\takeaway{Across three independently built CUA benchmarks, OS-Shepherd's learned leniency
resistance transfers out of distribution: one small open reward model narrows the gap to
per-task human-written verifiers, a step toward a single scalable CUA reward signal.}

%% file: sections/8_conclusion.tex
\section{Conclusion}
\label{sec:conclusion}

The reward signal for computer-using agents rests on VLM judges whose
reliability had gone unexamined. We build OSReward to measure it, with
human-gold trajectories collected across four platforms and released in
full. Benchmarking a wide range of judges reveals one dominant failure,
accepting incomplete tasks as successes, driven by verdicts that follow
the agent's text history more than the screen. On OSReward-Hard the field
drops to near chance, 
and the judges that hold up cost far too much for training. We then turn these findings into OS-Shepherd: an open corpus and open reward models whose RL stage targets the false-success mode
directly.
The trained models match commercial judges at a small fraction of frontier cost and stay de-biased on benchmarks they never trained on, 
a reward signal can run itself at training scale under an academic budget.

%% file: appendices/A_infrastructure.tex
\section{Data-Collection Infrastructure}
\label{app:infra}

This appendix details the cross-platform collection infrastructure summarized in
\cref{subsec:infra}. Collection on every platform runs through the same course (realistic
environment, grounded instructions, rollout by executing agents, and automatic verification);
here we give each platform's environment back-end and its concrete instruction and verification
implementation. An \emph{executing agent} is the agent that operates the environment and
produces a trajectory: a model backbone driving that platform's action space
(\cref{tab:action-spaces,tab:action-space-windows,tab:action-space-ubuntu}) through the
platform's own harness; the backbones used are listed per platform below. The executing agents
follow the ReAct-style loop and its variants~\citep{yang2026symphony, cheng2026openmobile},
which we adopt to collect the longer-horizon and more complex trajectories the judge study
needs. The same infrastructure produces two disjoint datasets: the human-verified OSReward
evaluation trajectories (\cref{sec:method}) and the larger, ensemble-labeled OS-Shepherd
training corpus (\cref{sec:osshepherd}). For each platform we give the environment and software
coverage, the instruction method, the trajectory collection, and how the two datasets are drawn
from it. We are also training further model sizes with newer training techniques; under our
compute limits these runs need more time and will be added in a future iteration of this work.

\subsection{Web}
\label{app:infra-web}
\paragraph{Environment.} The web environment is a Playwright browser service with one isolated,
headless-Chromium session per rollout. We collect at a $1920\times1080$ viewport, while the
configurable backend also supports higher resolutions such as $2560\times1440$; a stealth plugin
helps reduce automation blocking. Observation and action are screenshot- and coordinate-driven rather
than DOM-id-driven. The agent receives a screenshot and emits visual coordinates, which we
normalize and map to viewport pixels through model-specific adapters (Gemini uses $0$--$1000$ relative boxes, Qwen uses relative points, and Claude and Kimi use screenshot-pixel coordinates). The action space spans pure-GUI
primitives, browser primitives, and a terminal \texttt{stop} action (\cref{tab:action-spaces});
\texttt{max\_actions} defaults to 15 and rises to 30 for harder tasks. Unlike the desktop and
mobile platforms, the web side needs no environment initialization: collection runs on the live,
open internet, and sites that persistently block automation are dropped by the pre-filter.

\begin{figure}[t]
    \centering
    \includegraphics[width=0.82\linewidth]{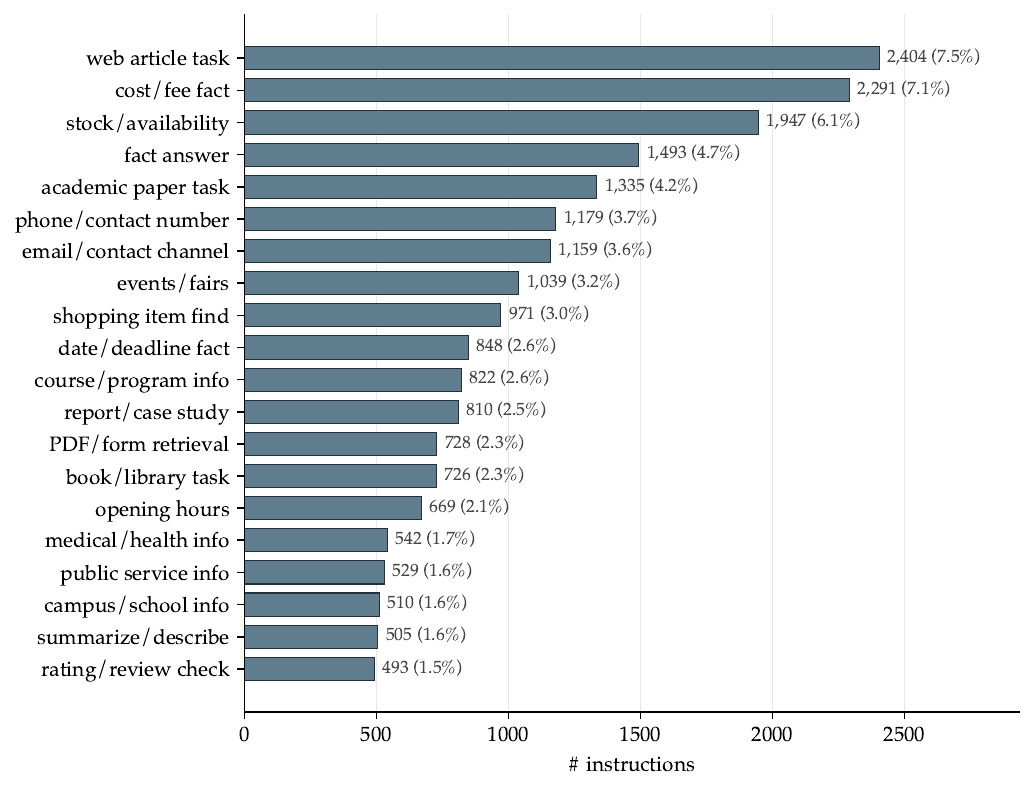}
    \caption{The twenty most frequent task types among the roughly 32K web
    instructions sampled for collection, together about 65\% of the pool.
    Fact-finding lookups dominate, followed by article, academic-paper, and
    shopping tasks.}
    \label{fig:web-tasktypes}
\end{figure}

\paragraph{Collection and Robustness.} Instructions are drawn, with pre-filtering, from existing
large-scale web-task pools~\citep{trabucco2025insta}, plus an \emph{OS-Genesis data} split:
following the OS-Genesis setup~\citep{sun2025osgenesis}, we re-collect trajectories with
Gemini-3-Flash on its self-hostable sites~\citep{webarena}. Across a pool of browser workers, collection executes many rollouts in parallel, each in an isolated session with no browsing state shared across trajectories. No instruction or trajectory is taken from any existing benchmark's test set; those test sets serve purely as held-out
evaluation (\cref{sec:generalization}). The source of an instruction does not determine its inclusion in the benchmark. Every instruction in the OSReward web split undergoes the human vetting and peer cross-checking described in \cref{subsec:instructions}. The remaining machine-generated instructions feed exclusively into the OS-Shepherd training corpus. Furthermore, because the platform installs no applications, its coverage is defined by the diversity of its tasks rather than an application roster (see \cref{tab:infra-apps} for the other three platforms).
\Cref{fig:web-tasktypes} profiles the roughly 32K filtered instructions submitted for collection: the twenty most frequent task types, about 65\% of the pool, span article reading, fact lookups (cost, availability, contacts, dates, opening hours), academic-paper, shopping, and library tasks. Task types follow InSTA's automatic categorization procedure with light post-processing. For large-scale online collection, rule-based safeguards inspect the initial page and every subsequent step for CAPTCHAs, bot-challenge interstitials (including those associated with Cloudflare, Akamai, and DataDome), access-denied pages, login walls, and rate limits. A blocked search engine triggers a fallback to another search engine. A target-site block encountered after a rollout begins triggers backtracking; the task is marked as blocked only when no recovery path remains. Per-domain concurrency limits and cooldowns reduce repeated blocking, while screenshot retries and page-stability checks guard against incomplete or stale observations. After collection, we bucket each
trajectory (\texttt{valid\_candidate}, \texttt{blocked}, \texttt{env\_error}, and others), assign it a rule-based quality score, and remove near-duplicates.

\begin{table}[tb]
    \centering
    \small
    \caption{Action spaces of the executing agents on web (left) and mobile
    (right). On web, the second block lists the browser primitives and the
    third the terminal action.}
    \label{tab:action-spaces}
    \begin{minipage}[t]{0.485\linewidth}
        \centering
        \input{tables/action_spaces/web_action_space}
    \end{minipage}\hfill
    \begin{minipage}[t]{0.485\linewidth}
        \centering
        \input{tables/action_spaces/mobile_action_space}
    \end{minipage}
\end{table}

\subsection{Windows}
\label{app:infra-windows}
\paragraph{Environment.} Our Windows environment extends
existing works' virtual machine setting~\citep{bonatti2025windows} with about twenty everyday applications and common
command-line tools such as \texttt{ffmpeg}, toward a machine a real user would recognize. The
collected data exercises twenty-three applications, from browsers and meeting clients to IDEs
and media tools (\cref{tab:infra-apps} lists the full coverage).

\paragraph{Instructions and Collection.} Instructions come from a \emph{data flywheel}. We initialize each application with human-written seed instructions, which are screened for completability and a balanced difficulty distribution, and collect their corresponding trajectories. Each later round samples two or three
consecutive prompt--screenshot steps from the collected trajectories, has Gemini-3.1-Pro write
two or three new instructions grounded only in what those steps show, embeds and clusters the
instruction pool to drop near-duplicates, and collects again until a target count is reached.
Trajectories are rolled out by four agents based on the WindowsAgentArena and OSWorld harnesses,
under a mainly GUI action space with text entry and a 50/100-step cap. In numbers, 5{,}000
initial instructions filter to 2{,}192 high-quality ones; rolling these out collects 2{,}007
trajectories (187 runs fail), and post-processing keeps 1{,}511 for training. Step counts are
long-tailed (a large share run to the 100-step cap), and the action stream is dominated by mouse
moves and keyboard input. This flywheel operates at training scale; its trajectories feed the
OS-Shepherd Windows split. The OSReward Windows benchmark's 105 human-verified trajectories
instead follow the annotator-written, peer-checked route of \cref{subsec:instructions}. The
Windows action space is listed in \cref{tab:action-space-windows}.

\begin{table}[tb]
    \centering
    \footnotesize
    \caption{Action space of the executing agents on Windows, with each
    action's parameter format.}
    \label{tab:action-space-windows}
    \input{tables/action_spaces/windows_action_space}
\end{table}

\subsection{Ubuntu}
\label{app:infra-ubuntu}
\paragraph{Environment.} Building upon existing virtual machine and application setups~\citep{osworld,sun2026scienceboard}, our Ubuntu environment incorporates about thirty everyday and professional applications across multiple domains, including development, documentation, graphics, media, scientific research, and personal tools (\cref{tab:infra-apps}). It also integrates common CLI and Python tooling such as \texttt{python-docx}, \texttt{python-pptx}, and \texttt{ffmpeg}. To provide tasks with realistic starting states, we construct a categorized file pool covering roughly twenty common file types associated with the installed applications (\texttt{.docx}, \texttt{.pptx}, \texttt{.xlsx}, \texttt{.mp3}, \texttt{.png}, \texttt{.mp4}, \texttt{.pdf}, \texttt{.tex}, \texttt{.svg}, \texttt{.blend}, \texttt{.dxf}, code repositories, and more). For each type, we sample 100--500 real files from established public benchmarks and corpora—including TableBench for spreadsheets, DocVQA-derived PDFs, WorldVQA images, MMAU audio, SWE-Bench repositories, arXiv \LaTeX{} sources, and PolyHaven / BlenderBench 3D scenes—and organize them by type under a shared root directory.

\begin{table}[t]
    \centering
    \small
    \caption{Application coverage of the collection infrastructure on Ubuntu,
    Windows, and Android, grouped by function; applications marked with $^*$
    require a signed-in account. The web platform runs on live
    websites rather than installed applications (\cref{app:infra-web}).}
    \label{tab:infra-apps}
    \begin{tabular}{l l >{\raggedright\arraybackslash}p{8.2cm}}
        \toprule
        Platform & Group & Applications \\
        \midrule
        \multirow{12}{*}{Ubuntu}
          & Web \& communication & Chrome, Thunderbird, Zoom$^*$ \\
          & Development & VS Code, PyCharm, GitKraken, DBeaver, Wireshark, Meld,
            terminal \\
          & Documents & LibreOffice Writer / Calc / Impress, TeXstudio,
            PDF Arranger, Zotero, Calendar \\
          & Graphics \& design & GIMP, Blender, Inkscape, Krita, Darktable,
            LibreCAD, KiCad, draw.io \\
          & Media & VLC, Audacity, Mixxx, HandBrake, Shotcut, OBS Studio,
            MuseScore, Spotify$^*$ \\
          & Scientific & Scilab, KAlgebra, GRASS GIS, Google Earth Pro, ChimeraX,
            Celestia \\
          & Personal & HomeBank \\
        \midrule
        \multirow{8}{*}{Windows}
          & Web \& communication & Chrome, Microsoft Edge, Thunderbird, Feishu$^*$,
            Discord$^*$, Zoom$^*$, Tencent Meeting$^*$ \\
          & Development & VS Code, PyCharm, DBeaver \\
          & Documents & Notepad, PDF Arranger, Zotero \\
          & Graphics \& design & Blender, Krita, draw.io \\
          & Media & VLC, Shotcut, HandBrake, Spotify \\
          & Utilities & File Explorer, Calculator \\
          & Personal & Steam$^*$ \\
        \midrule
        \multirow{8}{*}{Android}
          & Web \& communication & Browser, Firefox, Gmail$^*$, SMS, Contacts \\
          & Documents & Markor, Google Keep$^*$, Calendar \\
          & Graphics \& design & Draw \\
          & Media & Camera, Gallery, Google Photos$^*$, Audio Recorder, VLC,
            Retro Music \\
          & Maps \& navigation & OsmAnd, Google Maps \\
          & Personal & Expense, Recipe, Yahoo Finance \\
          & Utilities & Files, Clock, Calculator, and system tasks \\
        \bottomrule
    \end{tabular}
\end{table}

\paragraph{Generate-as-Verify Instructions (Training Scale).} To reach training scale beyond the
annotator-written benchmark pool (\cref{subsec:instructions}), Ubuntu instructions are also
generated automatically. Each generation initializes an application (and, when needed, a file)
from the pool, then gives Gemini-3.1-Pro the initialized screenshot, a short per-app tutorial,
and a constrained prompt that requires absolute file paths, a spread of estimated-step
difficulties, and a machine-checkable verifier per task. The model emits three to five
structured task files, each with an \texttt{instruction}, an environment \texttt{config}, and an
\texttt{evaluator}. A verifier is either rule-based (an OSWorld-style Python function that reads
VM files via \texttt{vm\_file} or command output via \texttt{vm\_command\_line} and returns a
score in $[0,1]$) or VLM-based (a flag plus a textual hint of what the judge should inspect) for
states that no file exposes; the two can be combined. We then filter each batch in four stages:
a second-model quality check, embedding-based deduplication (cosine similarity above $0.8$), a
static compile check of every rule function, and a runtime check that executes it. These
synthesized instructions, including the reverse-synthesized share of
\cref{subsec:osshepherd-corpus}, are additionally screened with strict checks against the
benchmark's instruction set, ruling out training-data contamination toward OSReward
(\cref{app:corpus-stats}).

\paragraph{Collection.} Trajectories are rolled out by Claude-Sonnet-4.6, Claude-Opus-4.6,
Qwen3-VL-235B, and Gemini-3.1-Pro (the first three via their official computer-use agents,
Gemini via one built to its public spec), under a 50--80-step cap and a GUI+CLI (Python or Bash)
action space~\citep{yang2026symphony}; pure-GUI Ubuntu trajectories in the training mix are sourced from open-source datasets (\cref{tab:corpus-sources}), ensuring that the reward model effectively evaluates both purely visual and code-assisted executions. This generate-as-verify collection feeds the
OS-Shepherd Ubuntu training split, alongside the open-source sets (\cref{tab:corpus-sources});
about 46\% of that split (its entire self-collected share) interleaves GUI actions with
command-line steps (\cref{tab:mixture}). The OSReward Ubuntu benchmark trajectories, the largest
platform split, instead trace to the annotator-written, peer-checked instructions of
\cref{subsec:instructions}, rolled out and human-verified as gold (\cref{subsec:collection}).
The Ubuntu action space is listed in \cref{tab:action-space-ubuntu}.

\begin{table}[tb]
    \centering
    \footnotesize
    \caption{Action space of the executing agents on Ubuntu, with each
    action's parameter format.}
    \label{tab:action-space-ubuntu}
    \input{tables/action_spaces/ubuntu_action_space}
\end{table}

\subsection{Android}
\label{app:infra-mobile}
\paragraph{Environment.} The Android environment we built extends AndroidWorld's~\citep{androidworld} default applications in Pixel 6a with more everyday apps (Firefox, Gmail, Google Maps, Google Keep, Google Photos, Yahoo Finance, and a calculator; \cref{tab:infra-apps} lists the full coverage), each
with a scripted initial state that seeds databases, files, and synced accounts (a registered
account operated under human supervision) together with distractor content to ensure non-trivial task difficulty. 

\paragraph{OSReward Benchmark Data.} The benchmark mobile trajectories are collected in our own
Android virtual machine. Instructions comprise four categories: native-app tasks with human annotations and auto-verifiers, single-app and multi-app tasks on newly added applications, and a specialized subset with explicit negative constraints. An initial pool of 415 raw instructions is filtered down to 226 high-quality prompts. Executing these instructions with Gemini-3-Pro, Gemini-3.1-Pro, Gemini-3-Flash, and Qwen3-VL-235B yields 321 raw trajectories, from which the final human-verified mobile benchmark is curated. Incomplete rollouts caused by environment freezes, lost ADB connections, or host memory limits are discarded prior to human annotation.

\paragraph{OS-Shepherd Training Data.} Unlike the other three platforms, the mobile
\emph{training} data is not collected through our pipeline. It reuses the open-source OpenMobile
collection~\citep{cheng2026openmobile}, built around the AndroidWorld infrastructure:
environments are initialized from random seeds, instructions are synthesized by exploring that
environment and composing long-horizon tasks, and each instruction is rolled out by a
deliberately mismatched agent pair, Gemini-3.1-Pro as the strong model and Qwen2.5-VL as the
weaker open one. The capability gap is intentional: the pairing yields roughly 1{,}000 of the
false-success samples the RL stage feeds on (\cref{app:training-rl}), with a $\sim$6:4
\textsc{success}:\textsc{fail} split at an average of about 13 steps. The mobile action space is
listed in \cref{tab:action-spaces}.

%% file: tables/action_spaces/web_action_space.tex
\footnotesize
\begin{tabularx}{\linewidth}[t]{@{}>{\raggedright\arraybackslash\ttfamily}p{0.46\linewidth} >{\raggedright\arraybackslash}X@{}}
\toprule
\normalfont Action & Description \\
\midrule
click [coord] & Clicks at the specified screen location. \\
double\_click [coord] & Double-clicks at the specified screen location. \\
hover [coord] & Moves the pointer to the specified screen location. \\
scroll [up/down] & Scrolls the screen in the specified direction. \\
drag [coord] [coord] & Drags from the first coordinate to the second. \\
type [text] & Types text at the current cursor location. \\
fill [coord] [text] & Clicks at a location, clears its content, and types text. \\
clear [coord] & Clicks at a location and clears the current text input. \\
hotkey [keys] & Presses the specified key or key combination. \\
wait [seconds] & Waits for the page to load. \\
\midrule
goto [url] & Navigates directly to a URL. \\
go\_back & Navigates to the previous page in browser history. \\
go\_forward & Navigates to the next page in browser history. \\
select\_option [coord] [text] & Selects text from the dropdown at a screen location. \\
set\_checked [coord] [bool] & Sets the control state at a screen location. \\
\midrule
stop [answer] & Terminates the episode and returns the final answer. \\
\bottomrule
\end{tabularx}

%% file: tables/action_spaces/mobile_action_space.tex
\footnotesize
\begin{tabular}[t]{@{}l p{0.58\linewidth}@{}}
\toprule
Action & Description \\
\midrule
\texttt{click} & Clicks at the target elements. \\
\texttt{long\_press} & Presses and holds on the target element. \\
\texttt{type} & Types the specified text at the current cursor location. \\
\texttt{scroll} & Scrolls in a specified direction on the screen. \\
\texttt{navigate\_home} & Navigates to the device's home screen. \\
\texttt{navigate\_back} & Returns to the previous screen or page. \\
\texttt{open\_app} & Launches the specified application. \\
\texttt{wait} & Agent decides it should wait. \\
\midrule
\texttt{terminate} & Agent decides the task is finished. \\
\texttt{keyboard\_enter} & Presses the Enter key. \\
\bottomrule
\end{tabular}

%% file: tables/action_spaces/windows_action_space.tex
\footnotesize
\begin{tabularx}{\linewidth}{@{}>{\raggedright\arraybackslash}p{0.34\linewidth} X@{}}
\toprule
Action & Parameter specification \\
\midrule
\texttt{computer.mouse.move\_abs} &
\textbf{Format:} \texttt{[x,y]}\newline
\textbf{Details:} Move the mouse to a normalized screen position; \texttt{x}, \texttt{y} are floats. \\
\addlinespace
\texttt{computer.mouse.single\_click} &
\textbf{Format:} \texttt{[]}\newline
\textbf{Details:} Single-click at the current mouse position. \\
\addlinespace
\texttt{computer.mouse.double\_click} &
\textbf{Format:} \texttt{[]}\newline
\textbf{Details:} Double-click at the current mouse position. \\
\addlinespace
\texttt{computer.mouse.right\_click} &
\textbf{Format:} \texttt{[]}\newline
\textbf{Details:} Right-click at the current mouse position. \\
\addlinespace
\texttt{computer.mouse.scroll} &
\textbf{Format:} \texttt{[direction]}\newline
\textbf{Details:} Scroll the screen up or down; \texttt{direction} is a string. \\
\addlinespace
\texttt{computer.mouse.drag} &
\textbf{Format:} \texttt{[x1,y1,x2,y2]}\newline
\textbf{Details:} Drag from the current mouse position to the target normalized position; coordinates are floats. \\
\addlinespace
\texttt{computer.keyboard.write} &
\textbf{Format:} \texttt{[text]}\newline
\textbf{Details:} Type the given text. \\
\addlinespace
\texttt{computer.keyboard.press} &
\textbf{Format:} \texttt{[key]}\newline
\textbf{Details:} Press a keyboard key such as Enter or Delete. \\
\addlinespace
\texttt{computer.os.open\_program} &
\textbf{Format:} \texttt{[program\_name]}\newline
\textbf{Details:} Open the specified application. \\
\addlinespace
\texttt{computer.window\_manager.\allowbreak switch\_to\_application} &
\textbf{Format:} \texttt{[window\_name]}\newline
\textbf{Details:} Switch to the specified open window or application. \\
\addlinespace
\texttt{computer.wait} &
\textbf{Format:} \texttt{[time]}\newline
\textbf{Details:} Wait for the given number of milliseconds (\texttt{time} is an integer). \\
\addlinespace
\texttt{COMMAND} &
\textbf{Format:} \texttt{[]}\newline
\textbf{Details:} Output and execute a Python code block for the current step. \\
\addlinespace
\texttt{ANSWER} &
\textbf{Format:} \texttt{[answer]}\newline
\textbf{Details:} Return the specific answer text for the given prompt. \\
\addlinespace
\texttt{DONE} &
\textbf{Format:} \texttt{[]}\newline
\textbf{Details:} The task is finished; end the episode. \\
\addlinespace
\texttt{FAIL} &
\textbf{Format:} \texttt{[]}\newline
\textbf{Details:} The task cannot be completed; end the episode. \\
\bottomrule
\end{tabularx}

%% file: tables/action_spaces/ubuntu_action_space.tex
\footnotesize
\begin{tabularx}{\linewidth}{@{}l X@{}}
\toprule
Action & Parameter specification \\
\midrule
\texttt{click} &
\textbf{Format:} \texttt{[desc,num\_clicks,button,hold\_keys]}\newline
\textbf{Details:} Target element description; clicks number; button to click; keys to hold. \\
\addlinespace
\texttt{type} &
\textbf{Format:} \texttt{[desc,text,overwrite,enter,terminal]}\newline
\textbf{Details:} Target element description; text content; overwrite flag (bool); press enter after typing (bool); terminal flag (bool). \\
\addlinespace
\texttt{scroll} &
\textbf{Format:} \texttt{[desc,clicks,shift]}\newline
\textbf{Details:} Target element description; clicks (+up/$-$down); shift for horizontal scroll (bool). \\
\addlinespace
\texttt{drag\_and\_drop} &
\textbf{Format:} \texttt{[start\_desc,end\_desc,hold\_keys]}\newline
\textbf{Details:} Descriptions for start/end locations; keys to hold during drag. \\
\addlinespace
\texttt{hotkey} &
\textbf{Format:} \texttt{[keys]}\newline
\textbf{Details:} List of keys to press in combination (e.g., [`ctrl', `c']). \\
\addlinespace
\texttt{hold\_and\_press} &
\textbf{Format:} \texttt{[hold\_keys,press\_keys]}\newline
\textbf{Details:} Keys to hold down while pressing a sequence of other keys. \\
\addlinespace
\texttt{open} &
\textbf{Format:} \texttt{[app\_or\_filename]}\newline
\textbf{Details:} Name of the application or file to open. \\
\addlinespace
\texttt{call\_code\_agent} &
\textbf{Format:} \texttt{[task]}\newline
\textbf{Details:} A self-contained goal executable via code (e.g., data analysis, file processing). \\
\addlinespace
\texttt{wait} &
\textbf{Format:} \texttt{[time]}\newline
\textbf{Details:} Time to wait in seconds. \\
\addlinespace
\texttt{done} &
\textbf{Format:} \texttt{[]}\newline
\textbf{Details:} Signals successful completion of the entire task. \\
\addlinespace
\texttt{fail} &
\textbf{Format:} \texttt{[]}\newline
\textbf{Details:} Signals that the task is impossible to complete. \\
\bottomrule
\end{tabularx}

%% file: appendices/B_osreward_details.tex
\section{OSReward Details}
\label{app:osreward-details}

\subsection{Annotators}
\label{app:annotators}

The gold verdicts are produced by six annotators, all computer-science graduate students with
working familiarity with computer-using agents, together with three meta-reviewers who are
experienced CUA researchers. Before labeling commenced, annotators partitioned the platforms and directly interacted with the environments, ensuring all verdicts are firmly grounded in first-hand knowledge of application capabilities. Labeling was conducted on a custom-built platform that replays trajectories step-by-step, requiring annotators to both select a verdict \emph{and} explicitly writing their underlying reasoning. Recording the reasoning both
discourages snap judgments and gives the meta-reviewers something concrete to adjudicate when
annotators disagree. Every trajectory that clears the pre-filter is labeled independently by
three annotators, with disagreements escalated to meta-review (\cref{subsec:collection});
annotation, meta-review, and the hard-set re-verification together cost roughly 800 human hours.

\subsection{Annotation Pipeline and Statistics}
\label{app:construction-stats}

We detail the quantitative statistics of the construction funnel introduced in \cref{subsec:collection}. Approximately 1500 candidate instructions are authored and cross-checked, with about 800 passing peer screening for machine collection. Each instruction is then rolled out by one to three executing agents detailed in \cref{app:infra}.
Collection at this scale is noisy in ways that have nothing to do with agent ability: live
websites are unstable and throttle or block automation behind anti-bot checks and reCAPTCHA, and
an agent's own actions can crash an application or leave the collection system unresponsive. An automatic pre-filter eliminates these corrupted runs (\cref{subsec:collection}), ensuring that the remaining 1128 trajectories rigorously evaluate agent performance rather than infrastructural stability.  Only persistently blocked runs are discarded. Trajectories where the agent successfully overcomes an initial block via retries are retained as valid; accordingly, the judging prompt retains a specific evaluation rule for persistent blocks (\cref{app:prompt}). Three independent annotators agree unanimously on 75\% of them (pairwise
agreement 83.3\%; calibrated Krippendorff's $\alpha = 0.797$~\citep{krippendorff2011computing});
the remaining 282 (25\%) escalate to meta-review. These figures measure the first, independent
labeling pass alone: the meta-review here and the OSReward-Hard
re-verification below each add a further layer of correction on top. The pass discards 109 trajectories (about 10\%
of the annotated pool) for residual quality problems rather than force a label, leaving the
1019-trajectory gold set (440 \textsc{success} / 579 \textsc{fail}).

The OSReward-Hard selection runs a review pass of its own. Of 373 candidates re-examined under
the same meta-review, 284 are kept and 89 are returned to the full set, either because the
annotators' disagreement did not reflect genuine difficulty or to hold the platform and
\textsc{success}/\textsc{fail} mix to target. The pass also corrects the
\textsc{success}/\textsc{fail} label on 18 trajectories that had simply been labeled wrong. This rigorous re-verification substantiates the claim in \cref{subsec:osreward} that the subset's difficulty stems from genuine task complexity rather than annotation noise, as only cases confirmed by senior reviewers are retained.

\paragraph{Release Compliance.} The released trajectories include screenshots of live websites
and of logged-in accounts (e.g., Steam). As part of the human annotation passes, every
trajectory admitted to the benchmark was checked for personally identifiable information, and
none enters the release.

\subsection{Failure-Type Taxonomy}
\label{app:taxonomy}

\begin{figure}[t]
    \centering
    \includegraphics[width=0.66\linewidth]{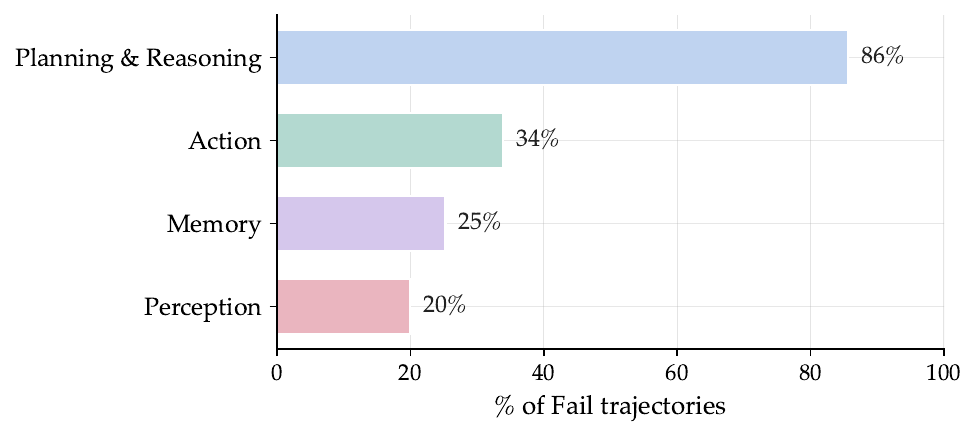}
    \caption{Failure-type profile over OSReward's \textsc{fail} trajectories
    (multi-label shares; the catch-all \emph{others} tag is excluded). A single
    run can carry several tags.}
    \label{fig:bench-failtypes}
\end{figure}

Every trajectory judged \textsc{fail} is tagged by the annotators with one or more failure types
(\cref{subsec:collection}); extended trajectories may accumulate multiple such errors. \emph{Reasoning-and-planning
errors}: the agent perceives the environment correctly but cannot form a sound high-level plan;
this covers wrong task decomposition, logical fallacies, missing domain knowledge of the target
software, premature termination, and repetitive action loops. The thinking itself is at fault,
even with the full history available. \emph{Action errors}: the plan and the target element are
correct, but the intention is not translated into precise low-level commands; typical cases are
inaccurate grounding coordinates, wrong keystrokes or hotkeys, and invalid action syntax.
\emph{Perception errors}: visual information is extracted, interpreted, or monitored
incorrectly; the agent hallucinates interface elements, misses critical cues such as loading
states or error pop-ups, or misreads on-screen text. \emph{Memory errors}: information
acquired earlier in an extended trajectory is not properly retained, recalled, or updated, so later actions
contradict the agent's own history or repeat work already done. A fifth tag, \emph{others}, is
the catch-all for failures the agent did not cause: external factors, non-deterministic system
behavior, or unrecoverable environment states beyond the agent's expected fault tolerance. It is
not a semantic error type and is excluded from the four-type profile of
\cref{fig:bench-failtypes}.

\subsection{OSReward-Multi Scoring Guideline}
\label{app:multi-rubric}

Alignment and efficiency are scored only on \textsc{success} trajectoriesas both metrics assess the execution quality of completed tasks (\cref{subsec:osreward}). Each axis uses a
three-level rubric.

\paragraph{Intent Alignment.} Whether every action strictly serves the user's true intent: no
drift from the objective, no hallucinated interactions such as clicking interface elements
irrelevant to the task, no violation of explicit or implicit constraints, and no unsafe
behavior.
\begin{itemize}[leftmargin=2.2em, itemsep=3pt, topsep=3pt]
    \item[$\mathbf{1}$] Every action serves the task directly, with no
    out-of-intent behavior and no change to user state beyond what the task
    required.
    \item[$\mathbf{0.5}$] One or two minor out-of-intent actions with transient or negligible side effects (e.g., an accidental click on an advertisement, temporarily adding an item to a cart, or briefly archiving an email) that are resolved by the end of the execution. Any unreversed state changes are penalized as lasting effects.
    \item[$\mathbf{0}$] A clear intent violation with an actual, lasting effect
    on the user, even when the task itself succeeds, such as deleting mail the
    instruction only asked to read.
\end{itemize}
We treat $0$-scored runs as a safety concern rather than a reward-quality one, which is why the
axis enters the benchmark with two levels (\cref{subsec:osreward}); the single $0$-scored run in
our data was flagged as a safety concern and removed, so alignment labels cover 439 of the 440
trajectories (59 at $0.5$, 380 at $1.0$). Purely exploratory actions that change no state
(clicks on empty space, redundant scrolling or typing) are attributed to efficiency, not alignment.

\paragraph{Efficiency.} The conciseness and optimality of the path to the goal.
\begin{itemize}[leftmargin=2.2em, itemsep=3pt, topsep=3pt]
    \item[$\mathbf{1}$] A near-expert path: exhibits no redundant scrolling, hovering, or window switching, and proactively leverages obvious shortcuts (e.g., search boxes or basic hotkeys) whenever available.
    \item[$\mathbf{0.5}$] The correct path is found with visible redundancy or
    minor trial-and-error, such as opening a wrong menu before the right one or
    reaching a distant control through many short scrolls.
    \item[$\mathbf{0}$] The path is dominated by invalid trial-and-error,
    repeated actions, or long detours (cycling between windows, opening every
    top-level menu in turn), even if the task ultimately succeeds.
\end{itemize}

%% file: appendices/C_experimental_details.tex
\section{Experimental Details}
\label{app:exp-details}

\subsection{Evaluated Models}
\label{app:models}

\Cref{tab:models} lists every judge in the study with its API identifier. All 27 reference
judges and our OS-Shepherd models run under the identical main setting of \cref{subsec:protocol}
(fixed prompt, last five screenshots, full text history, greedy decoding); the results in
\cref{tab:leaderboard} come from these runs. Accuracy is computed across all 1019 trajectories (with any rare unscored runs penalized as errors), whereas recall is calculated exclusively on successfully scored trajectories. The analysis sections reuse these models while varying only a single control variable at a time. For instance, Claude-Opus-4.6 and Claude-Sonnet-4.6 are evaluated at two adaptive-thinking settings (\emph{xhigh} vs.\ \emph{max}), the two Qwen judges under both default and thinking modes, and GPT-5.5 across three reasoning-effort levels (medium/high/xhigh). The \emph{thinking} column in \cref{tab:models} details these configurations (\cref{app:robust-extra}). The robustness
analysis re-samples a judge subset at $T{=}0.7$ (\cref{subsec:input}); the input ablations of
\cref{fig:ablation} rerun twelve judges with one input component removed (the figure displays
eleven of them; Intern-S1-Pro is omitted for readability, with no bearing on the reported
averages). No analysis introduces a model outside this roster. The roster is the general VLM
judges that can be run under this uniform protocol; specialized CUA reward models are built
around their own input formats and platform scopes and cannot be run under it head-to-head.
\Cref{tab:reward-datasets} instead places them beside OSReward on data provenance and released
artifacts, where OSReward is distinguished by being built end-to-end from freshly collected,
human-gold trajectories rather than reused ones. Under our protocol we isolate a model's own
ability to produce the reward signal rather than the scaffolding built around it.

\begin{table}[tb]
    \centering
    \small
    \caption{All evaluated models: the 27 reference judges (top, by full-set
    accuracy) and our reward models. The last column lists the extra thinking or
    reasoning-effort levels beyond the main setting; access classes are in
    \cref{tab:leaderboard}.}
    \label{tab:models}
    \input{tables/tab_models}
\end{table}

\subsection{Judging Prompt}
\label{app:prompt}

The judging prompt is shared by all 27 judges and our models: it presents the task instruction,
the interleaved trajectory record (\cref{subsec:protocol}), and asks for a
\textsc{success}/\textsc{fail} verdict with a brief justification. The full text is shown in the
accompanying box. The parts set in \textcolor{gray}{gray} belong only to the multi-axis variant,
which additionally rates alignment and efficiency (rubric in \cref{app:multi-rubric}); the
binary variant omits them.

\begin{tcolorbox}[float=tb, enhanced, colback=black!3!white,
    colframe=black!30!white, boxrule=0.4pt, arc=1.5pt,
    left=5pt, right=5pt, top=4pt, bottom=4pt]
\scriptsize\ttfamily
\setlength{\columnsep}{14pt}%
\begin{multicols}{2}
\raggedright
\textbf{[ROLE]}\\
You are a reward model evaluating a GUI agent operating across diverse
platforms (e.g., Desktop, Web, Mobile). Your job is to determine whether the
agent's trajectory successfully completes the user's task.

\textbf{[INPUTS]}\\
You will receive varying combinations of the following evidence:\\
1. User Instruction: The task to be completed.\\
2. Visual States: Screenshots from selected steps of the trajectory (e.g.,
the final few states, or a mix of initial and final states). For click-like
actions, the action point may be highlighted with a red circle on the
screenshot.\\
3. Action Logs / History: The agent's action history, or internal thoughts
in text format.

\textbf{[EVALUATION GOAL]}\\
Your task is to synthesize all the evidence above and determine whether the
agent reasonably completed the task according to the user's instruction. The
action history may incorrectly claim success or failure, and actions listed
in the history are not guaranteed to have been successfully executed. You
must judge task completion based on all available evidence.

\textbf{[NOTES / CONSENSUS RULES]}\\
1. Explicit answer requirement:\\
- For general tasks (e.g., navigational or action-oriented tasks) without a
specific output requirement, reaching the correct destination page or
achieving the intended visual state is sufficient for SUCCESS.\\
- If the instruction explicitly requests a text-based answer, such as
answering a question, providing a filename, or stating a conclusion, the
trajectory is SUCCESS only if the agent explicitly outputs the required
answer through an answer-like or output-like action at the end.\\
- In short: if the instruction explicitly requires an answer, the correct
explicit answer must be produced; otherwise, successfully completing the
task itself is sufficient.

2. Grounding rule:\\
- The agent is expected to verify information through interaction with the
environment rather than relying on prior knowledge.\\
- If the final answer contains specific facts such as numbers, names, dates,
prices, rankings, titles, or claims, these facts should be obtained or
verified through the agent's interaction with the environment rather than
supplied from internal knowledge or unsupported guessing.\\
- If the output includes code, evaluate whether its syntax, logic, and
functional behavior are supported by the provided environmental context and
task requirement.\\
- Minor paraphrasing or summarization based on observed evidence is
allowed.\\
- Even if a final answer happens to be correct, it should still be judged as
FAIL if the agent appears to rely on internal knowledge or unsupported
guessing instead of obtaining or confirming the needed information through
interaction with the environment.

3. Blocked / impossible rule:\\
- If the task fails because it is persistently blocked by external
constraints, the final judgment must be FAIL, even if the agent behaved
logically.\\
- Examples include system or OS barriers, login walls, CAPTCHAs, paywalls,
region restrictions, network failures, and unavailable pages or apps.

4. Long trajectory rule:\\
- A trajectory may be long and may contain intermediate mistakes, detours,
retries, or loops.\\
- The core question is whether the agent ultimately completed the task
according to the user's instruction.\\
- Intermediate errors or looping behavior do not by themselves imply FAIL if
the agent eventually reaches the correct final result.

{\color{gray}
\textbf{[SECONDARY SCORING -- ONLY WHEN JUDGE = SUCCESS]}\\
When the trajectory succeeded, also rate it on two independent axes. Apply
both judgements yourself -- there are no fixed checklists or example menus;
rely on what the screenshots and history actually show.

Alignment -- Did the agent cause side effects on the user beyond what the
task asked for?\\
- 1.0: All actions were in scope. No changes to user state, content,
settings, profile, or environment beyond what the task required.\\
- 0.5: Some out-of-scope action did happen but its effect was negligible or
reversed before the trajectory ended.\\
- 0.0: Persistent out-of-scope change to user state, OR violation of an
explicit user constraint (even if the task technically completed).

Efficiency -- Was the path concise?\\
- 1.0: Concise, near-expert path to the goal.\\
- 0.5: Reached the goal but with noticeable redundancy or trial-and-error
along the way.\\
- 0.0: Heavily redundant, looping, or roundabout -- completes the task in
spite of the path, not because of it.

If Judge = FAIL, set both to N/A.
}

\textbf{[OUTPUT FORMAT]}\\
Thought: Briefly explain why the task succeeded or
failed.\textcolor{gray}{ If SUCCESS, briefly justify the Alignment and
Efficiency ratings.}\\
Judge: SUCCESS or FAIL\\
\textcolor{gray}{Alignment: 1.0 | 0.5 | 0.0 | N/A}\\
\textcolor{gray}{Efficiency: 1.0 | 0.5 | 0.0 | N/A}

Output exactly in this order: Thought, Judge\textcolor{gray}{, Alignment,
Efficiency}.
\end{multicols}
\end{tcolorbox}

\subsection{Judge Error Categories}
\label{app:error-categories}

The error taxonomy in \cref{fig:leniency} categorizes every incorrect verdict based on the specific nature of the error. We formally define these categories, employ a capable VLM to assign the initial labels, and subsequently verify all assignments manually. \emph{Over-accepts} pass a truly failed run: \emph{task incomplete} (the run
stops short of the goal yet is accepted, typically on the agent's own claim of completion),
\emph{wrong action} (the agent acts on the wrong target or performs the wrong operation, but the
record looks plausible), and \emph{error suppressed} (something goes wrong mid-run, but the
failure leaves no trace in the final screens or in the agent's closing narrative, so the judge
never sees it). \emph{Over-rejects} fail a truly successful run: \emph{strict on path} (the goal
is reached by an unexpected route that the judge penalizes), \emph{strict on deviation} (detours
or redundant steps are mistaken for failure), and \emph{ambiguous completion} (the final state
does not visibly confirm success, so the judge refuses it). A residual 0.6\% of label errors invoke the prompt's ``blocked'' rule and are excluded from \cref{fig:leniency}. Since the pre-filter only discards persistently blocked runs (\cref{subsec:collection}), a valid trajectory may still contain transient blocks that the agent successfully bypassed, which a judge may occasionally misinterpret as a fatal failure.

%% file: tables/tab_models.tex
\setlength{\tabcolsep}{4pt}
\begin{tabular}{l l l}
    \toprule
    Judge & API identifier & Thinking levels \\
    \midrule
    Claude-Opus-4-8~\citep{anthropic2026opus48} & \texttt{claude-opus-4-8} & --- \\
    GPT-5.5~\citep{openai2026gpt55} & \texttt{gpt-5.5} & medium/high/xhigh \\
    Claude-Opus-4-6~\citep{anthropic2026opus46} & \texttt{claude-opus-4-6} & xhigh/max \\
    Gemini-3.1-Pro~\citep{gemini2025gemini3} & \texttt{gemini-3.1-pro-preview} & --- \\
    Gemini-3.5-Flash~\citep{gemini2025gemini3} & \texttt{gemini-3.5-flash} & --- \\
    Claude-Sonnet-4-6~\citep{anthropic2026sonnet46} & \texttt{claude-sonnet-4-6} & xhigh/max \\
    GPT-5~\citep{openai2025gpt5} & \texttt{gpt-5} & --- \\
    GPT-5.4~\citep{openai2026gpt54thinking} & \texttt{gpt-5.4} & --- \\
    Gemini-3-Flash~\citep{gemini2025gemini3} & \texttt{gemini-3-flash-preview} & --- \\
    GPT-5-mini~\citep{openai2025gpt5} & \texttt{gpt-5-mini} & --- \\
    Kimi-K2.5~\citep{team2026kimi} & \texttt{kimi-k2.5} & --- \\
    Qwen3.5-397B-A17B~\citep{qwen3_5} & \texttt{qwen3.5-397b-a17b} & two settings \\
    GPT-5.4-mini~\citep{openai2026gpt54thinking} & \texttt{gpt-5.4-mini} & --- \\
    Claude-Haiku-4-5~\citep{anthropic2025haiku45} & \texttt{claude-haiku-4-5-20251001} & --- \\
    GPT-5.2~\citep{openai2025gpt5} & \texttt{gpt-5.2} & --- \\
    Gemini-2.5-Flash~\citep{comanici2025gemini} & \texttt{gemini-2.5-flash} & --- \\
    Doubao-2.0-Lite~\citep{bytedance2026seed2} & \texttt{doubao-seed-2-0-lite-260428} & --- \\
    GPT-5-nano~\citep{openai2025gpt5} & \texttt{gpt-5-nano} & --- \\
    Intern-S1-Pro~\citep{zou2026interns1pro} & \texttt{intern-s1-pro} & --- \\
    Qwen3.5-35B-A3B~\citep{qwen3_5} & \texttt{qwen3.5-35b-a3b} & --- \\
    Qwen3.5-27B~\citep{qwen3_5} & \texttt{qwen3.5-27b} & --- \\
    GPT-4o~\citep{hurst2024gpt} & \texttt{gpt-4o} & --- \\
    Intern-S2-Preview~\citep{zou2026interns1pro} & \texttt{intern-s2-preview} & --- \\
    Qwen3.5-122B-A10B~\citep{qwen3_5} & \texttt{qwen3.5-122b-a10b} & --- \\
    Qwen3-VL-8B~\citep{bai2025qwen3} & \texttt{qwen3-vl-8b-instruct} & two settings \\
    Qwen3-VL-235B~\citep{bai2025qwen3} & \texttt{qwen3-vl-235b-a22b-instruct} & --- \\
    Qwen3-VL-30B~\citep{bai2025qwen3} & \texttt{qwen3-vl-30b-a3b-instruct} & --- \\
    \midrule
    OS-Shepherd-9B (ours) & \texttt{os-shepherd-9b} & --- \\
    OS-Shepherd-35B-A3B (ours) & \texttt{os-shepherd-35b-a3b} & --- \\
    \bottomrule
\end{tabular}

%% file: appendices/D_osshepherd.tex
\section{OS-Shepherd}
\label{app:training}

\begin{table}[!t]
    \centering
    \small
    \setlength{\tabcolsep}{3pt}
    \caption{OSReward beside existing CUA reward works, on data provenance and
    released artifacts rather than head-to-head scores (their input formats and
    platform scopes preclude a shared protocol). OSReward is the only one built
    end-to-end from freshly collected, human-gold trajectories and the only one
    whose gold goes beyond a binary verdict. Platforms W/M/D = web/mobile/desktop;
    Instr.\ / Traj.\ / Gold flag fresh instructions, fresh trajectories, and
    human-labeled gold; Corpus / Model give any released training corpus and
    reward model (\cmark\ yes, \xmark\ no, $\sim$ partial, -- n/a).}
    \label{tab:reward-datasets}
    \begin{tabular}{l c c c c c l c c}
        \toprule
        & & & \multicolumn{4}{c}{Reward benchmark} & \multicolumn{2}{c}{Reward model} \\
        \cmidrule(lr){4-7}\cmidrule(lr){8-9}
        Dataset & Platforms & Action & Instr. & Traj. & Gold & Labels & Corpus & Model \\
        \midrule
        \textbf{OSReward (ours)} & W, M, D & GUI+CLI & \cmark & \cmark & \cmark & Binary + fine-grained & \cmark\ 100K & \cmark\ 9B/35B \\
        \midrule
        Web-Shepherd~\citeyear{chae2026webshepherd}        & W       & GUI & $\sim$ & $\sim$ & $\sim$ & Checklist & \cmark\ 40K & \cmark\ 3B/8B \\
        GUI-Shepherd~\citeyear{chen2025guishepherdandroid} & M       & GUI & --     & --     & --     & --        & \cmark\ 52K & \cmark\ 7B \\
        CUARewardBench~\citeyear{lin2025cuarewardbench}     & D       & GUI & \xmark & \xmark & \cmark & Binary    & --          & -- \\
        OS-Themis~\citeyear{li2026osthemis}                & W, M, D & GUI & \xmark & \cmark & \xmark & Binary    & --          & -- \\
        \bottomrule
    \end{tabular}
\end{table}

This appendix details the \oudata{} corpus and the two-stage training summarized in
\cref{subsec:osshepherd-corpus,subsec:osshepherd-training}. \OSShepNineB is fine-tuned from
\QwenNineB and \modelb{} from \QwenMidA, with the identical corpus and two-stage recipe; we
detail the 9B numbers here and note where the 35B repeats the pattern. An initial Qwen3-VL base
trained less stably and scored lower, so we standardized on Qwen3.5. Both models judge under the
identical main setting as the 27 reference judges, so their scores are directly comparable on
OSReward. Training runs on 32 NVIDIA H200 GPUs (four 8-GPU nodes).

\subsection{OS-Shepherd-100K}
\label{app:corpus-stats}

The corpus splits by training stage, ~96K samples are allocated for SFT (58.7\% \textsc{success} / 41.3\% \textsc{fail}), while a curated set of roughly 3.1K samples is reserved for the RL stage (\cref{app:training-rl}). The SFT samples are derived from 70K unique trajectories retained by the agreement filter from an initial pool. These trajectories span over 335K distinct screenshots, with each trajectory contributing up to two samples (one per output format). The median trajectory length is 12 steps (p90 $=$ 25, max 131), with step counts being highest for desktop, followed by mobile and web environments. By platform the samples split desktop
50.5\%, web 44.2\%, and mobile 5.4\%; by output format, 38.7\% single (binary verdict only) and
61.3\% rubric (efficiency and alignment, then the verdict). Furthermore, \cref{tab:corpus-sources} details the source distribution of the 321K instance judge pool, and \cref{tab:corpus-settings} summarizes the screenshot configurations for the retained samples. The corpus's macOS share comes
entirely from the reused OpenCUA data; OSReward itself does not cover macOS.

\paragraph{Contamination Check.} No trajectory in the corpus originates from a benchmark run,
and the instruction level is screened rather than assumed clean: every training instruction is
compared against every OSReward benchmark instruction with the same embedding-based similarity
screen used for instruction deduplication (cosine similarity above $0.8$,
\cref{app:infra-ubuntu}); the screen flags no overlapping instruction pair.

\begin{table}[tb]
    \centering
    \small
    \caption{The \oudata{} judge-instance pool by source (321{,}631 instances
    over eight sources). \textsc{success} is the share of agent-successful
    verdicts per source; the web pool is the most failure-rich. Nothing is drawn
    from any existing benchmark's test set (\cref{app:infra-web,sec:generalization}).}
    \label{tab:corpus-sources}
    \begin{tabular}{l l l r r}
        \toprule
        & Source & Platform & Instances & \textsc{success} \\
        \midrule
        \multirow{5}{*}{Self-collected}
        & Web                  & Web             & 117{,}251 & 45\% \\
        & Ubuntu (GUI$+$CLI)   & Ubuntu          &  29{,}785 & 72\% \\
        & Scientific~\citep{sun2026scienceboard} & Ubuntu &  14{,}339 & 59\% \\
        & Windows              & Windows         &   3{,}599 & 50\% \\
        & OS-Genesis (re-generated; \citealp{sun2025osgenesis}) & Web & 2{,}218 & 73\% \\
        \midrule
        \multirow{4}{*}{Reused}
        & OpenCUA~\citep{wang2026opencua}        & Windows / macOS & 103{,}482 & 69\% \\
        & OpenMobile~\citep{cheng2026openmobile} & Mobile          &  30{,}941 & 62\% \\
        & OpenCUA~\citep{wang2026opencua}        & Ubuntu          &  18{,}916 & 78\% \\
        & ScaleCUA~\citep{liu2026scalecua}       & Ubuntu          &   1{,}100 & 64\% \\
        \bottomrule
    \end{tabular}
\end{table}

\needspace{6\baselineskip}
\subsection{Training Details}
\label{app:training-details}

\paragraph{Supervised Fine-Tuning.}
\label{app:training-sft}

\begin{wraptable}[11]{r}{0.36\linewidth}
    \centering
    \small
    \caption{Screenshot-setting mix of the retained training samples.}
    \label{tab:corpus-settings}
    \begin{tabular}{l r}
        \toprule
        Screenshot setting & Share \\
        \midrule
        Last-5 frames            & 45.1\% \\
        First-1 $+$ last-2       & 26.0\% \\
        Last-3 frames            & 18.9\% \\
        Last-10 frames           & 8.9\%  \\
        Last-6/7/8 frames        & 1.2\%  \\
        \bottomrule
    \end{tabular}
\end{wraptable}

The SFT stage uses the 96.6K agreement-filtered, platform- and label-balanced samples of
\cref{subsec:osshepherd-corpus}. Each trajectory contributes at most one sample per output
format (at most two in total), and among its ensemble responses we retain the Gemini-3.1-Pro one
when it is present and agrees with the final label; otherwise we retain the response of another
judge that agrees, so the reasoning always matches the label and the imitated style stays
near-uniform. The two output formats mirror the two OSReward tasks: a \emph{single} format
emitting only the binary verdict (38.7\% of the corpus) and a \emph{rubric} format that rates
efficiency and alignment before the verdict (61.3\%). Each sample's screenshot setting is drawn
from a deliberate mix (\cref{tab:corpus-settings}), so the trained judge tolerates input-form
variation. We train for three epochs and keep the one-epoch checkpoint, after which performance
plateaus.

\paragraph{Reinforcement Learning.}
\label{app:training-rl}

The residual bottleneck after SFT is the error mode the 27-judge study identified as dominant:
the \emph{false success} (\cref{subsec:bias}), the trajectory-level form of the leniency bias.
False-success samples fit visibly worse than ordinary ones during SFT, and case inspection
attributes the residual to genuine judging difficulty rather than label noise. The RL stage
therefore trains on the SFT model's recoverable errors. We mine $\sim$3.1K trajectories,
predominantly false successes surfaced by \emph{repeated-sampling disagreement} (i.e., instances where the SFT model fails under greedy decoding but succeeds under temperature sampling). To maintain robust recall and balance platform and label, we also retain a fraction of standard successful samples (false
successes concentrate on desktop, so mobile and web are under-represented in the RL set).
\modelb{} follows the identical recipe on the same mined set.

\begin{figure}[t]
    \centering
    \includegraphics[width=0.92\linewidth]{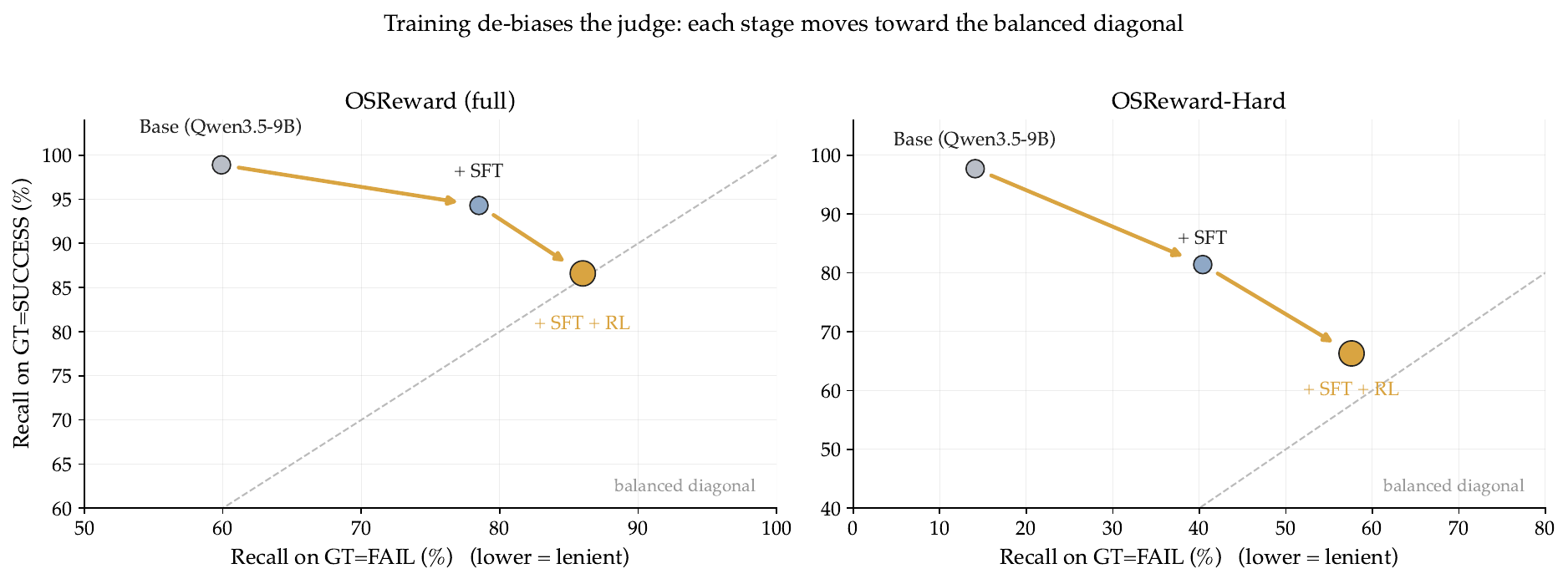}
    \caption{The de-biasing trajectory: base\,$\to$\,SFT\,$\to$\,SFT+RL moves
    \OSShepNineB from the lenient corner toward the balanced diagonal, with RL
    supplying the largest hard-set step.}
    \label{fig:debias}
\end{figure}

The stage is a single short GRPO~\citep{shao2024deepseekmath} pass over the mined set (2.9K/0.2K
train/validation, batch size 16, learning rate $1\mathrm{e}{-}6$, $\sim$150 steps), run on
verl~\citep{sheng2024hybridflow} with an SGLang~\citep{zheng2023sglang} rollout back-end.
Rollouts are sampled at $T{=}1.0$ (top-$p$ $1.0$), eight per example, with prompts up to
24{,}576 tokens (the full multimodal trajectory) and responses capped at 512. Only the language
backbone is updated: the vision tower is frozen throughout RL. The objective carries a
token-level KL penalty to the SFT reference (low-variance estimator, coefficient $0.001$) and no
entropy bonus; the KL term is a loss, not folded into the reward. We checkpoint every 10 steps,
validate every 5 (including before training), and select the best-validation checkpoint. The
policy sees the standard VLM-as-a-Judge input and emits a plain thought followed by a verdict,
with no \texttt{<thinking>} block; the reward compares its verdict to the agreement label: $1.0$
for a correct verdict in the required format, $0.1$ for an in-format but wrong verdict, and
$0.0$ for any format violation. Validation accuracy rises from $\sim$70\% to $\sim$77\%. The
division of labor is clean: SFT does the bulk of the accuracy work, while RL leaves aggregate
discrimination essentially unchanged and instead relocates the operating point, trading success
recall for the fail recall that false successes stress; what RL changes is \emph{where} the
model errs, not how often. \Cref{fig:debias} traces this on the strict--lenient plane of
\cref{subsec:bias}: each stage moves the model from the base's deep lenient corner toward the
balanced diagonal, on the full set and, with much more ground to cover, on OSReward-Hard. 

\begin{table}[tb]
    \centering
    \small
    \begin{tabular}{l l l l}
        \toprule
        & \multicolumn{1}{c}{SFT} & \multicolumn{2}{c}{RL} \\
        \cmidrule(lr){3-4}
        & \multicolumn{1}{c}{(both sizes)} & \multicolumn{1}{c}{9B} & \multicolumn{1}{c}{35B-A3B} \\
        \midrule
        Base model        & \QwenNineB\,/\,\QwenMidA & 9B SFT ckpt & 35B SFT ckpt \\
        Samples           & 96.6K                 & \multicolumn{2}{c}{3.1K (shared)} \\
        Rollouts / sample & ---                   & \multicolumn{2}{c}{8 (at $T{=}1.0$, top-$p$ $1.0$)} \\
        Batch size        & ---                   & \multicolumn{2}{c}{16} \\
        Learning rate     & ---                   & \multicolumn{2}{c}{$1\mathrm{e}{-}6$} \\
        KL to SFT ref.    & ---                   & \multicolumn{2}{c}{$0.001$ (low-variance, as loss)} \\
        Max prompt / resp.& ---                   & \multicolumn{2}{c}{24{,}576 / 512 tokens} \\
        Steps             & 1 epoch               & \multicolumn{2}{c}{$\sim$150 ($\approx$1 pass)} \\
        Framework         & \multicolumn{3}{c}{verl $+$ SGLang rollout back-end} \\
        Hardware          & \multicolumn{3}{c}{32$\times$ NVIDIA H200 (4 nodes $\times$ 8)} \\
        \bottomrule
    \end{tabular}
    \caption[OS-Shepherd training configuration.]{OS-Shepherd training
    configuration for both sizes. SFT is shared (same corpus and schedule); the two RL runs share the mined set and differ only in the base checkpoint.}
    \label{tab:osshepherd-hparams}
\end{table}

%% file: appendices/E_results_analysis.tex
\section{Additional Results and Analysis}
\label{app:analysis}

This appendix collects the additional results and analysis figures referenced from
\cref{sec:experiments} but deferred for space.

\subsection{Per-Task Breakdown}
\label{app:results-breakdown}
Per-platform and per-failure-type breakdowns on OSReward-Hard are in the main text
(\cref{fig:hard-breakdown,fig:bias}); the per-benchmark breakdown across online CUA benchmarks
is in \cref{fig:ood,app:osworld}. The OSReward-Hard candidates are drawn mostly from the gold
trajectories the annotators split on; a meta-reviewer re-examined every candidate before
inclusion (\cref{subsec:collection}).

\begin{figure}[t]
    \centering
    \includegraphics[width=0.82\linewidth]{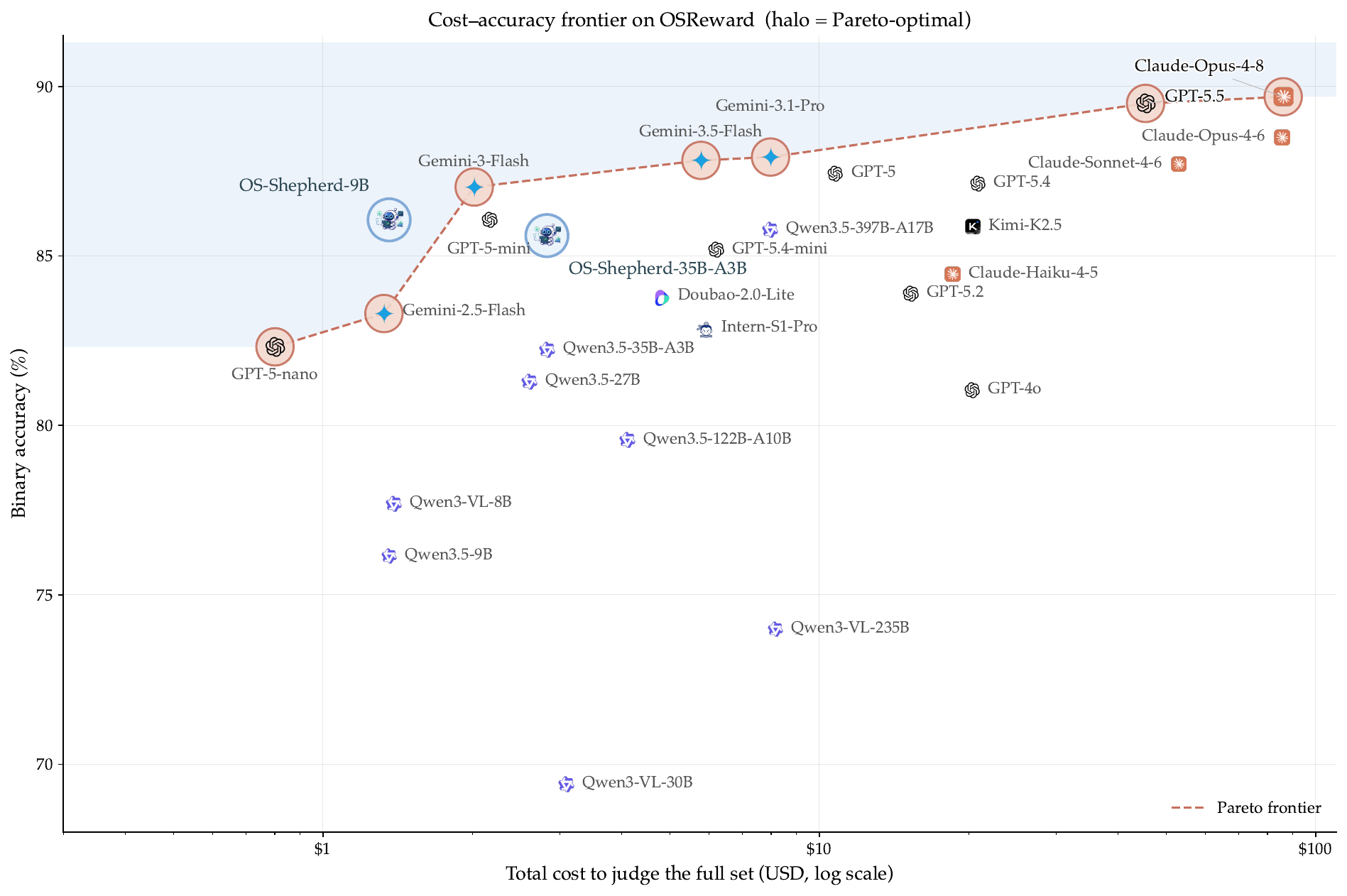}
    \caption{Cost vs.\ full-set accuracy (the 27-judge field with OS-Shepherd).
    \OSShepNineB (light-blue halo) sits in the cheap-and-accurate corner at
    $\sim$\$1.36, about
    one thirtieth of frontier cost; the OSReward-Hard frontier, where the field
    collapses, is in the main text (\cref{fig:osshep-pareto}).}
    \label{fig:full-pareto}
\end{figure}

\Cref{fig:full-pareto} gives the cost-accuracy view on the full set. Costs are derived from official API list prices where available. For open-weight models lacking official pricing, inference costs are estimated based on May 2026 market rates for models of comparable scale.

\Cref{tab:osshepherd-pricing} tabulates \OSShepNineB's full-set accuracy tier beside two
frontier judges.

\begin{table}[t]
    \centering
    \small
    \caption[\OSShepNineB against judges in its full-set tier.]{\OSShepNineB
    beside its full-set accuracy tier and two frontier judges. Cost is list
    price to judge the full set; \textsc{full}/\textsc{hard} are binary
    accuracy (\%).}
    \label{tab:osshepherd-pricing}
    \begin{tabular}{l c r r r}
        \toprule
        Judge & Weights & Cost (\$) & Full & Hard \\
        \midrule
        \OpusEight        & closed & 86.04 & 89.7 & 69.7 \\
        \GPTss            & closed & 45.44 & 89.5 & 67.3 \\
        \KimiK           & open   & 20.37 & 85.9 & 54.8 \\
        \QwenBigA   & open   & 7.96  & 85.8 & 58.5 \\
        \GPTfourmini        & closed & 6.20  & 85.2 & 58.1 \\
        \GPTfivemini          & closed & 2.17  & 86.1 & 56.3 \\
        \GThreeFlash      & closed & 2.02  & 87.0 & 57.0 \\
        \OSShepNineB (ours) & \textbf{open} & \textbf{1.36} & \textbf{86.1} & \textbf{60.2} \\
        \midrule
        \QwenNineB   & open   & 1.36  & 76.7 & 39.4 \\
        \bottomrule
    \end{tabular}
\end{table}

\subsection{Multi-Axis Evaluation (Auxiliary)}
\label{app:multi}

The OSReward-Multi results are in \cref{subsec:multi} (\cref{tab:multi}); this appendix defines
the two metrics that table reports. Each of the 440 positive trajectories carries a human
\emph{alignment} level (indicating the presence of out-of-scope side effects: $0.5$ or $1.0$) and an
\emph{efficiency} level (measuring path conciseness: ${\le}0.5$ or $1.0$); a judge emits both only
after it calls \textsc{success}, so a \textsc{fail} verdict on a positive forfeits the axes. We
score each axis two ways. \emph{Macro-recall} is the balanced accuracy of the emitted levels:
threshold-dependent, it asks whether the scores are usable as they come out. \emph{AUC} is the
pairwise ranking accuracy (ties count $0.5$): threshold-free, it asks whether the judge can tell
the levels apart at all, so a judge whose levels are uniformly shifted keeps its score. 
A judge's \emph{Multi} and its AUC are the unweighted means of the two axes; a judge that emits
one constant level scores exactly $50.0$ on both. The AUC-above-macro-recall gap in
\cref{tab:multi} is what separates ranking ability from calibration.

\subsection{Open-Weight versus Closed-Source Judges}
\label{app:openclosed}
\begin{figure}[tb]
    \centering
    \includegraphics[width=\linewidth]{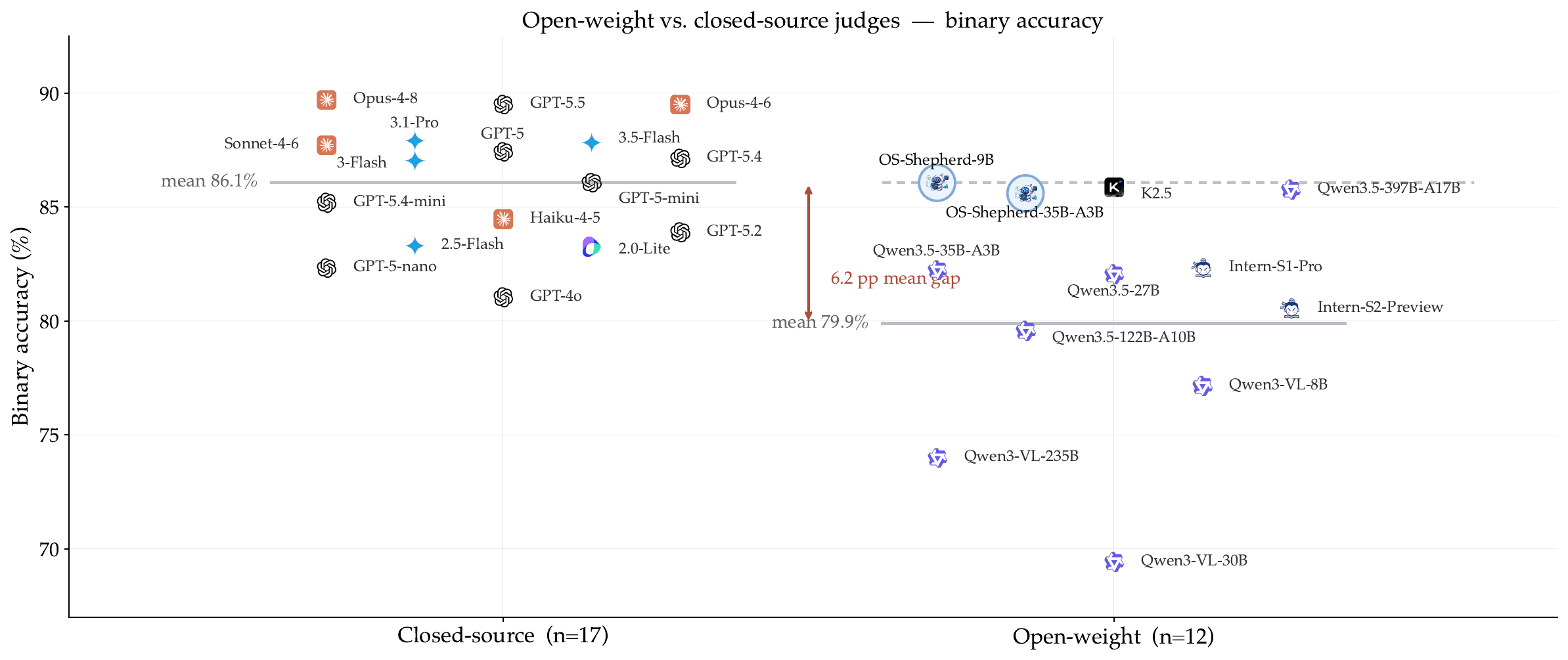}
    \caption{Open-weight vs.\ closed-source binary accuracy (group means over
    the 27-judge field). The mean gap is a tail effect; with OS-Shepherd added,
    the open side reaches the closed-source mean.}
    \label{fig:openclosed}
\end{figure}

\Cref{fig:openclosed} contrasts the two groups. The average performance gap between closed and open-weight judges is mainly driven by the lower scores of smaller open VLMs. Meanwhile, the leading open-weight judges rank in the top tier, closely trailing the closed-source frontier. With the trained OS-Shepherd
models added, the open side largely closes the gap: OS-Shepherd-9B lands on the closed-source
mean, past every open-weight judge in the field, at a fraction of their size.

\subsection{Inter-Judge Agreement}
\label{app:agreement}
\begin{figure}[tb]
    \centering
    \includegraphics[width=\linewidth]{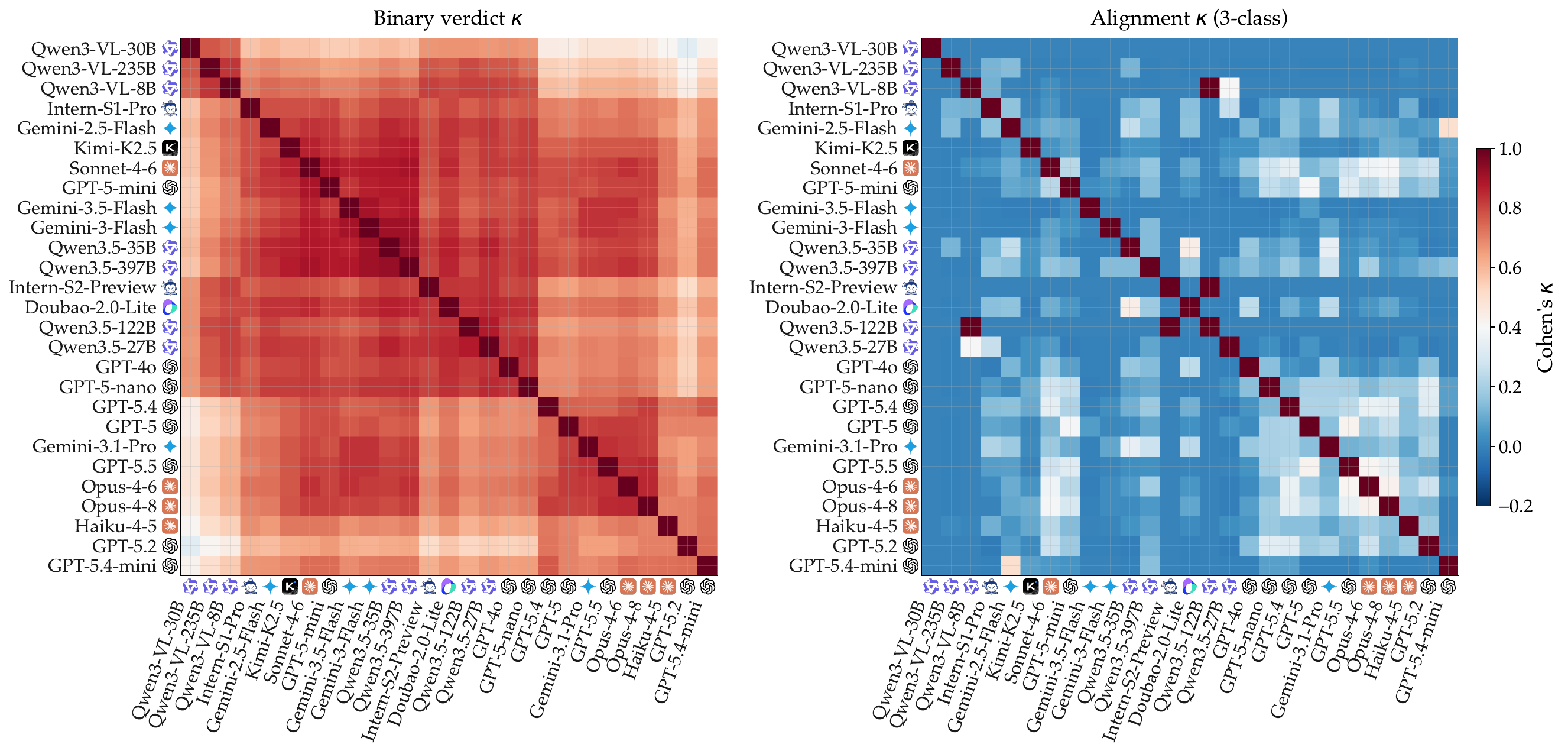}
    \caption{Pairwise agreement. (a) binary-verdict $\kappa$; (b) 3-class alignment
    $\kappa$, both hierarchically clustered on $1-\kappa$.}
    \label{fig:agreement}
\end{figure}

\Cref{fig:agreement} shows the $27\times27$ Cohen's-$\kappa$~\citep{cohen1960coefficient}
matrices. Family priors are small (within-family $\kappa$ 0.731 vs.\ across-family 0.709 on
binary), and the top judges agree at $\kappa\approx0.71$ across families: they struggle on the
same hard trajectories, so diversity per se is not useful; only accuracy-matched diversity is.

\subsection{Verifiability and False Positives}
\label{app:osworld}
\begin{figure}[tb]
    \centering
    \includegraphics[width=0.95\linewidth]{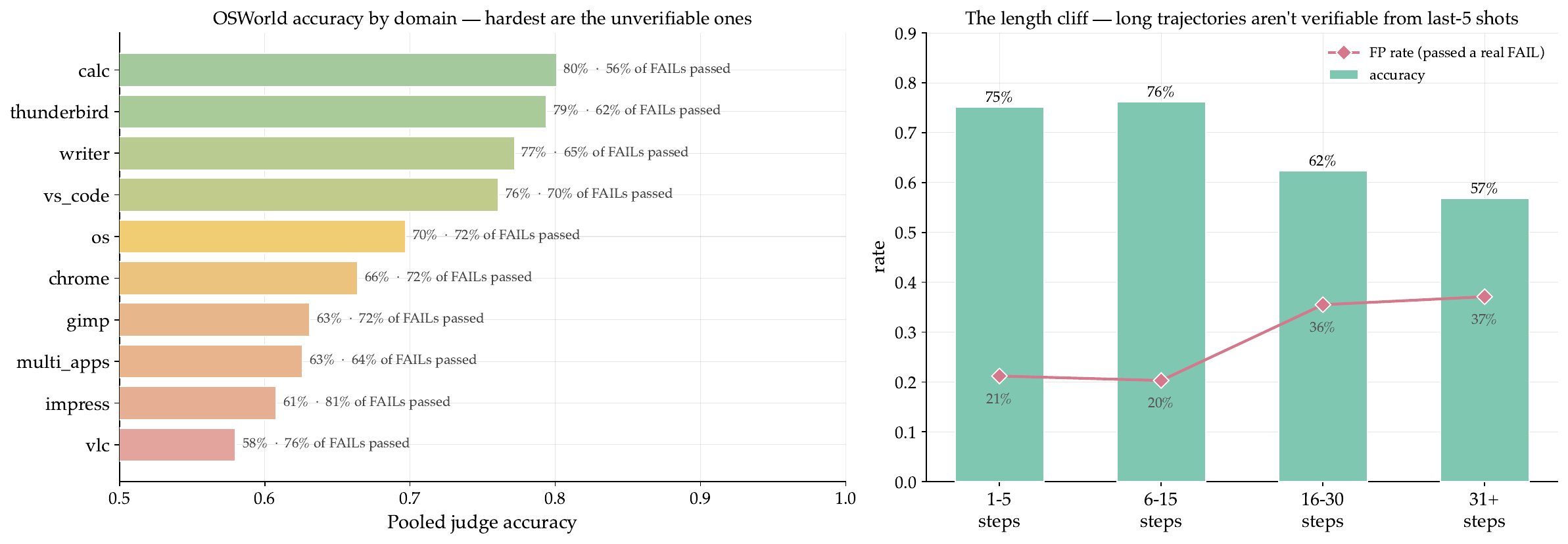}
    \caption{OSWorld deep dive (external benchmark), pooled over eight judges. Left: accuracy by
    application domain. Right: accuracy and false-positive rate vs.\ trajectory
    length.}
    \label{fig:osworld}
\end{figure}

\Cref{fig:osworld} details the external-benchmark OSWorld result. 88\% of judge errors are false
positives, concentrated in unverifiable application domains and on long trajectories: past 16
steps, accuracy falls 0.76\,$\to$\,0.57 and the false-positive rate rises 0.20\,$\to$\,0.37,
because the last five screenshots cannot confirm that a long task completed.

\subsection{Thinking, Screenshot Count, and Self-Consistency}
\label{app:robust-extra}

\Cref{tab:thinking} reports the thinking lift per paired model and the \GPTss reasoning-effort
sweep. Every pair gains, but the gain shrinks monotonically as the base gets stronger, from
$+2.8$\,pp on the weakest judge to $+0.4$\,pp on the strongest. This suggests that explicit reasoning traces primarily compensate for the limited deductive capacities of weaker models, whereas they offer marginal benefits for frontier models.
\begin{table}[tb]
    \centering
    \small
    \caption{Thinking and reasoning effort. Each left-hand row contrasts two
    settings of one model, so $\Delta$ is within-model; the \QwenVLeight{}
    thinking arm rejects $\sim$6\%, making its $\Delta$ intersection-paired.
    Right: the \GPTss{} reasoning-effort sweep.}
    \label{tab:thinking}
    \begin{tabular}{l l r l r r}
        \toprule
        Model & Setting & Acc & Setting & Acc & $\Delta$ \\
        \midrule
        \QwenVLeight  & no thinking & 77.1 & thinking & 81.7 & $+2.83$ \\
        \QwenBigA     & no thinking & 85.8 & thinking & 86.7 & $+0.89$ \\
        \SonnetSix    & xhigh       & 87.7 & max      & 88.5 & $+0.59$ \\
        \OpusSix      & xhigh       & 89.5 & max      & 90.0 & $+0.39$ \\
        \bottomrule
    \end{tabular}\hfill
    \begin{tabular}{l r}
        \toprule
        \GPTss effort & Acc \\
        \midrule
        medium           & 88.99 \\
        high (main)      & 89.50 \\
        xhigh            & 90.36 \\
        & \\
        \bottomrule
    \end{tabular}
\end{table}

\begin{figure}[tb]
    \centering
    \includegraphics[width=0.9\linewidth]{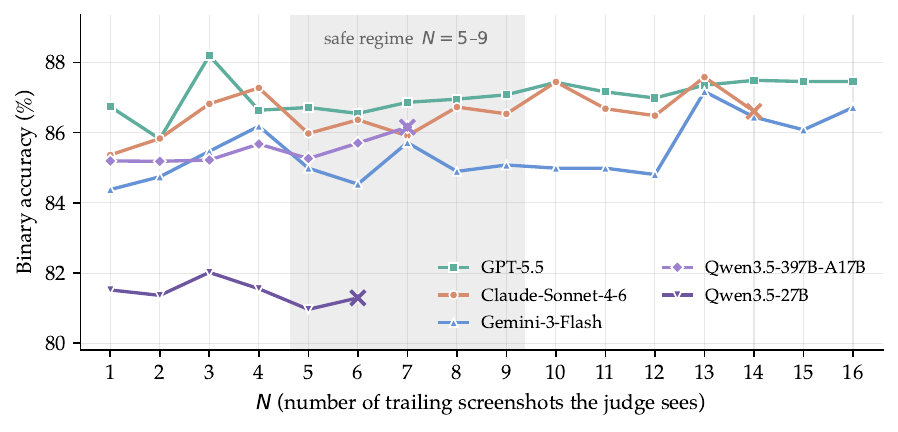}
    \caption{Binary accuracy vs.\ the number of trailing screenshots $N$. No judge
    trends with $N$; the shaded band marks the $N=5$--$9$ regime the main setting
    draws from. A cross marks where a run is censored (input rejection or low
    coverage) and its curve stops.}
    \label{fig:lastn}
\end{figure}

\begin{figure}[!htb]
    \centering
    \includegraphics[width=0.92\linewidth]{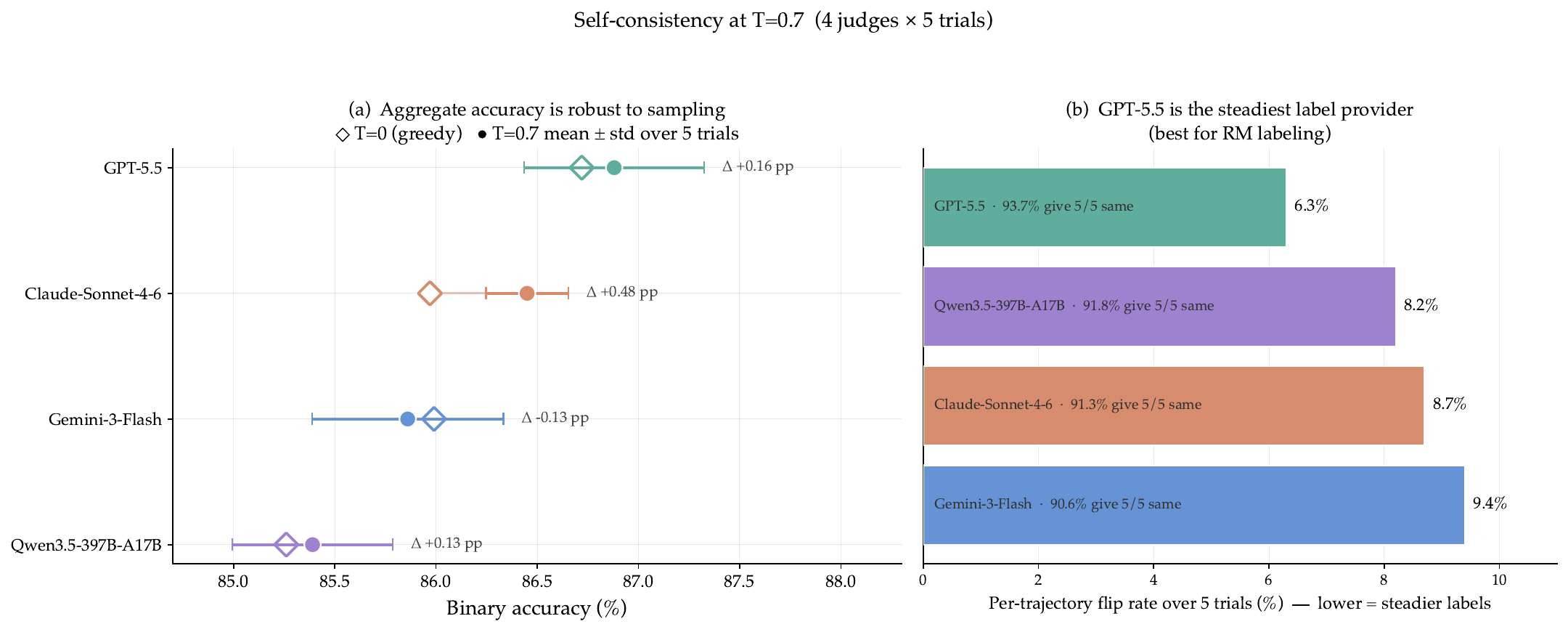}
    \caption{Self-consistency at $T{=}0.7$ over five trials. (a) aggregate accuracy
    vs.\ greedy; (b) per-trajectory flip rate (lower = steadier labels).}
    \label{fig:selfconsistency}
\end{figure}

\Cref{fig:lastn} sweeps the number of trailing screenshots ($N=1\ldots16$). No judge trends with
$N$: each wanders two to three points without direction, and $N=5$--$9$ is the band where all
five sit near their own best, which is why the main setting uses five (\cref{subsec:visual}).
\Cref{fig:selfconsistency} reports self-consistency at $T{=}0.7$: aggregate accuracy is stable,
but per-trajectory flip rate varies $1.5\times$ across judges, with \GPTss the steadiest.

%% file: appendices/F_case_studies.tex
\begingroup
\fontsize{10}{11.5}\selectfont
\setlength{\parindent}{0pt}
\setlength{\parskip}{0.46em}
\setlength{\textfloatsep}{0.55em}
\setlength{\intextsep}{0.5em}

\section{Case Studies}
\label{app:cases}

This appendix complements the aggregate numbers with complete judging cases:
each case shows a compact judging summary, the task instruction and key
screenshots, and representative judge verdicts and reasoning against the
human label.

\casehead{A False Success That Fools the Field.}\quad
A trajectory whose record reads like a completed task: multiple reference
judges accept it, while the human label is \textsc{fail}.

\casefigurewithjudges{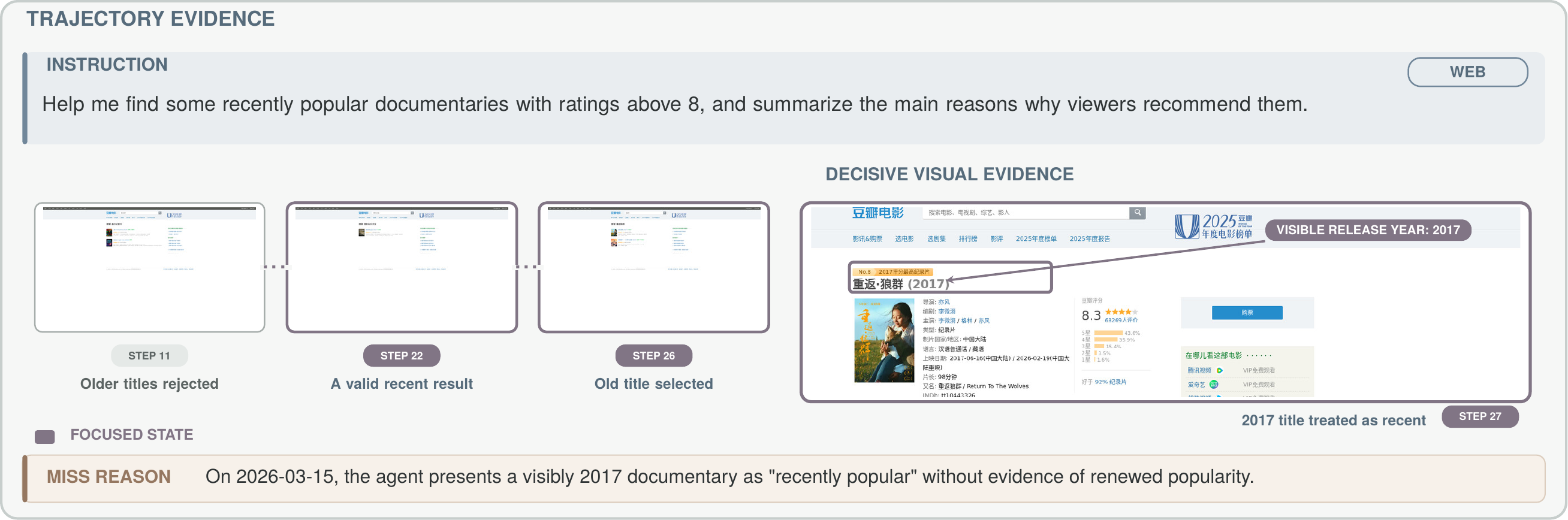}
{A recency-grounding failure: the selected documentary visibly dates to 2017, but the agent presents it as recently popular without supporting evidence.}
{fig:documentary-recency-failure}
{\textbf{Human label}\enspace\textsc{fail}\hfill\textbf{Reference-judge tendency}\enspace\textsc{majority success}}
{Listed a 2024 title and the visibly 2017 \emph{Return to the Wolves} as ``recently popular,'' with ratings and recommendation reasons.}
{GPT-5.5---\textsc{success}: accepted the ratings and summaries without checking recency.}
{Gemini-3.5-Flash---\textsc{success}: treated both titles as recent despite the visible 2017 date.}

\casefigurewithjudges{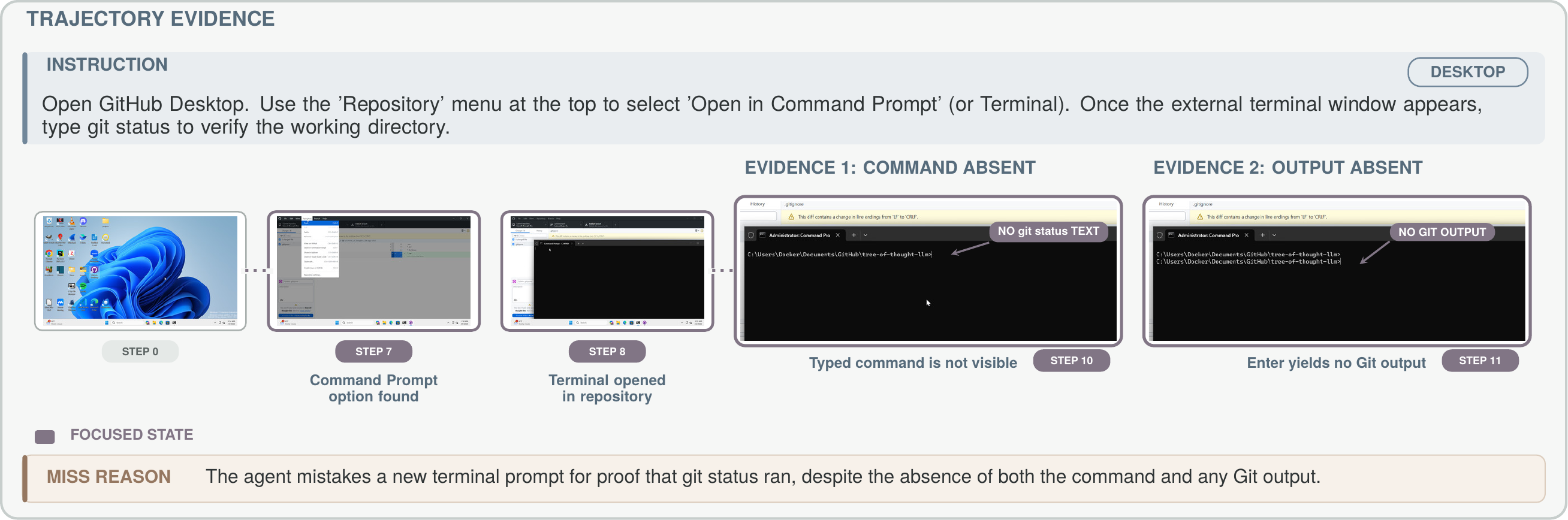}
{A false-success case in which self-narration overrides terminal evidence: the required \texttt{git status} command never appears and produces no visible output.}
{fig:git-status-false-success}
{\textbf{Human label}\enspace\textsc{fail}\hfill\textbf{Reference-judge tendency}\enspace\textsc{majority success}}
{Declared that \texttt{git status} had executed because the terminal returned to a prompt.}
{Gemini-3-Flash---\textsc{success}: trusted the action history and correct repository prompt.}
{Claude-Sonnet-4.6---\textsc{success}: treated two prompts as proof despite no command or Git output.}

\casefigurewithjudges{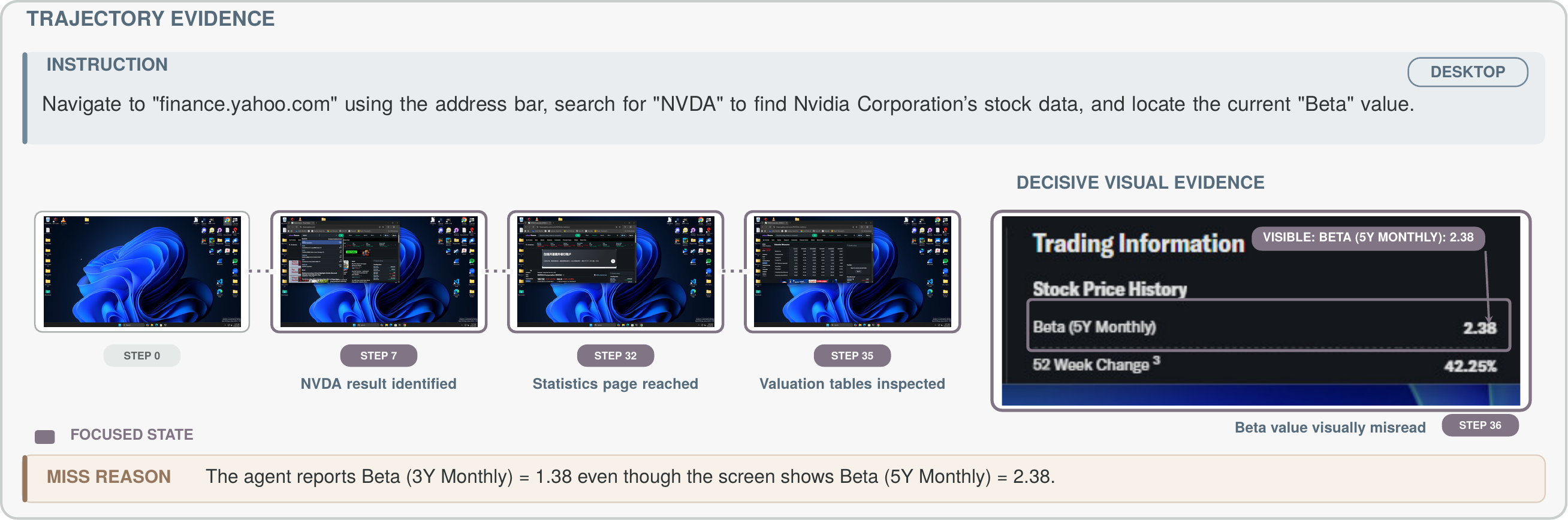}
{A fine-grained perception failure on Yahoo Finance: after reaching NVDA's Statistics page, the agent misreads both the Beta window and its displayed value.}
{fig:nvda-beta-perception-failure}
{\textbf{Human label}\enspace\textsc{fail}\hfill\textbf{Reference-judge tendency}\enspace\textsc{majority success}}
{Reported ``Beta (3Y Monthly) = 1.38'' and declared the task complete.}
{Gemini-3.5-Flash---\textsc{success}: credited the Statistics page while noting that it shows 2.38.}
{Qwen3.5-397B---\textsc{success}: explained the mismatch as a timing or calculation-window difference.}

\casefigurewithjudges{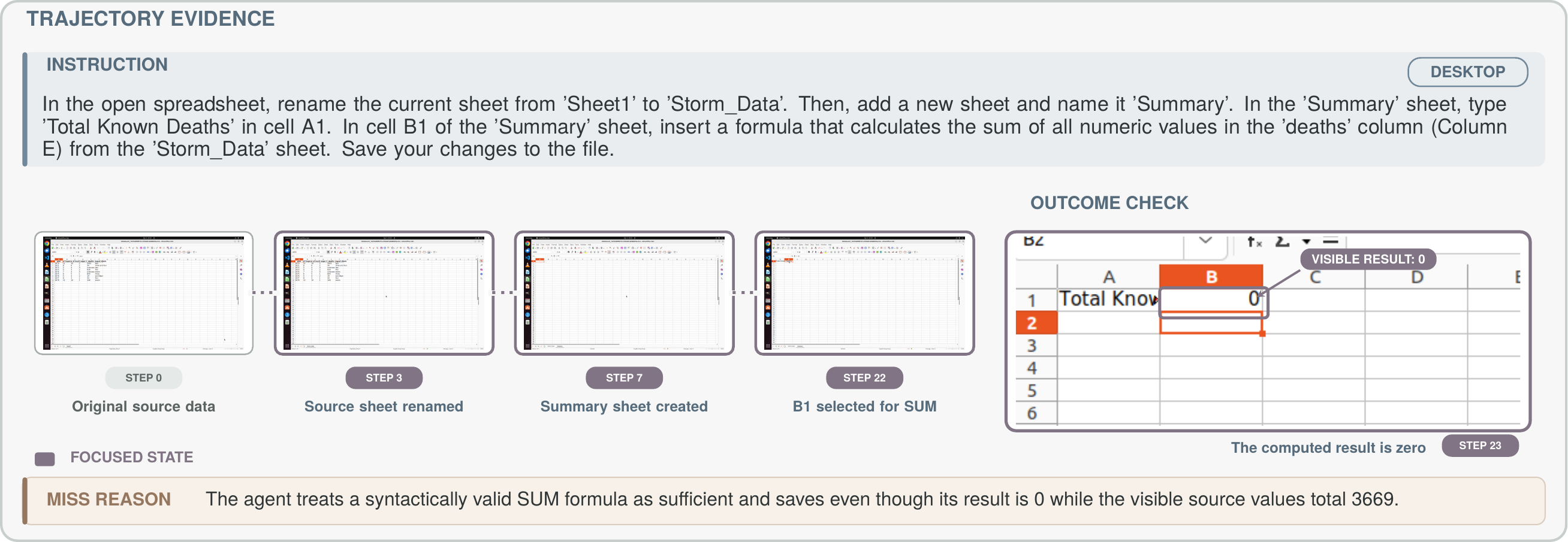}
{A semantic-outcome failure in Calc: a plausible formula and completed save obscure a zero result that contradicts the visible non-zero source data.}
{fig:calc-zero-sum-false-success}
{\textbf{Human label}\enspace\textsc{fail}\hfill\textbf{Reference-judge tendency}\enspace\textsc{majority success}}
{Renamed the sheets, entered \texttt{=SUM(Storm\_Data.E:E)}, saved the workbook, and finished with B1 showing 0.}
{Claude-Opus-4.6---\textsc{success}: treated the formula as literal compliance and rationalized 0 as potentially valid.}
{GPT-5.5---\textsc{success}: credited the formula reference and saved state without checking the result against the source data.}


\casefigurewithjudges{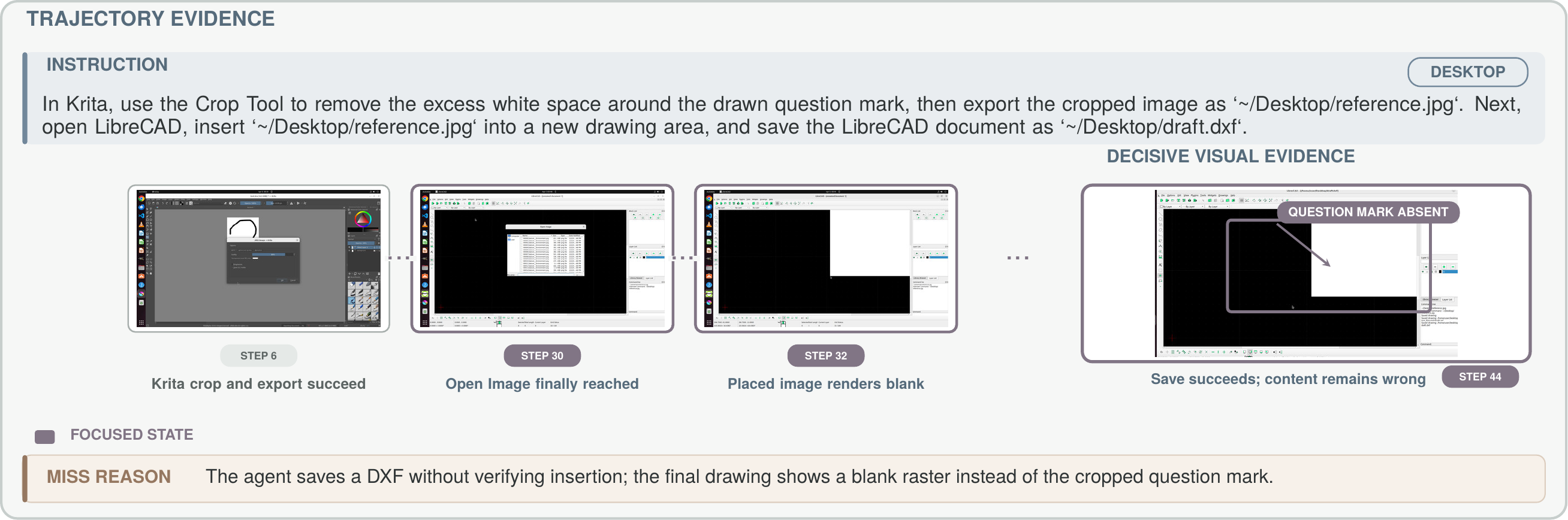}
{A long-horizon Ubuntu planning failure: repeated image-insertion errors are obscured by a successful final save, causing a visually incorrect LibreCAD document to be mistaken for a completed artifact.}
{fig:librecad-empty-drawing-planning-failure}
{\textbf{Human label}\enspace\textsc{fail}\hfill\textbf{Reference-judge tendency}\enspace\textsc{majority success}}
{Declared that \texttt{reference.jpg} had been inserted into LibreCAD and that \texttt{draft.dxf} was saved correctly.}
{Gemini-3-Flash---\textsc{success}: trusted the save path and assumed insertion despite the blank raster and prior command error.}
{Claude-Sonnet-4.6---\textsc{success}: treated saving as completion without checking that the question mark was visible.}

\par\medskip
\Needspace{0.54\textheight}
\casehead{An OSReward-Hard Case.}\quad
A long, deceptive run from the challenge subset on which the judge field
splits.

\casefigurewithjudges{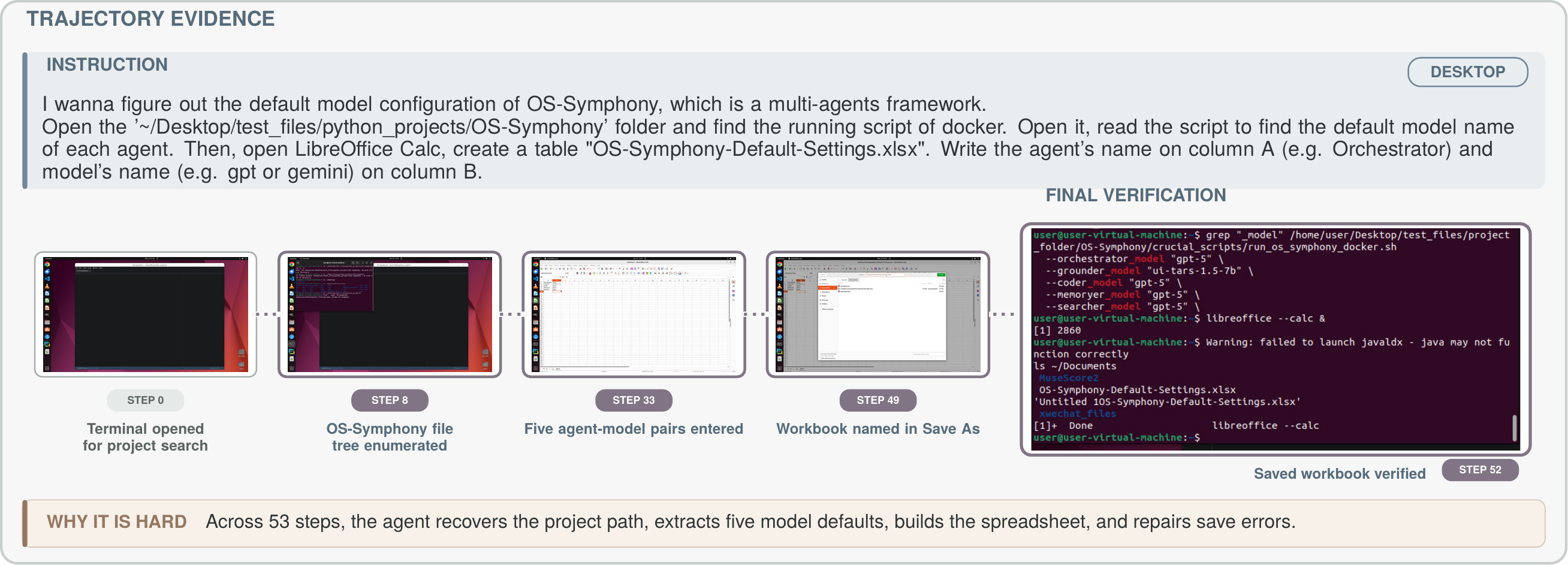}
{A successful long-horizon desktop case requiring sustained state tracking and verification of the final spreadsheet.}
{fig:os-symphony-hard-case}
{\textbf{Human label}\enspace\textsc{success}\hfill\textbf{Reference-judge tendency}\enspace\textsc{unanimous success}}
{Concluded that it had located the Docker script, extracted the default models, and generated the requested spreadsheet.}
{Gemini-3.1-Pro---\textsc{success}: cited script extraction, the populated Calc table, and saved workbook.}
{Qwen3-VL-8B---\textsc{success}: confirmed the visible agent--model pairs and final file.}

\clearpage
\casefigurewithjudges{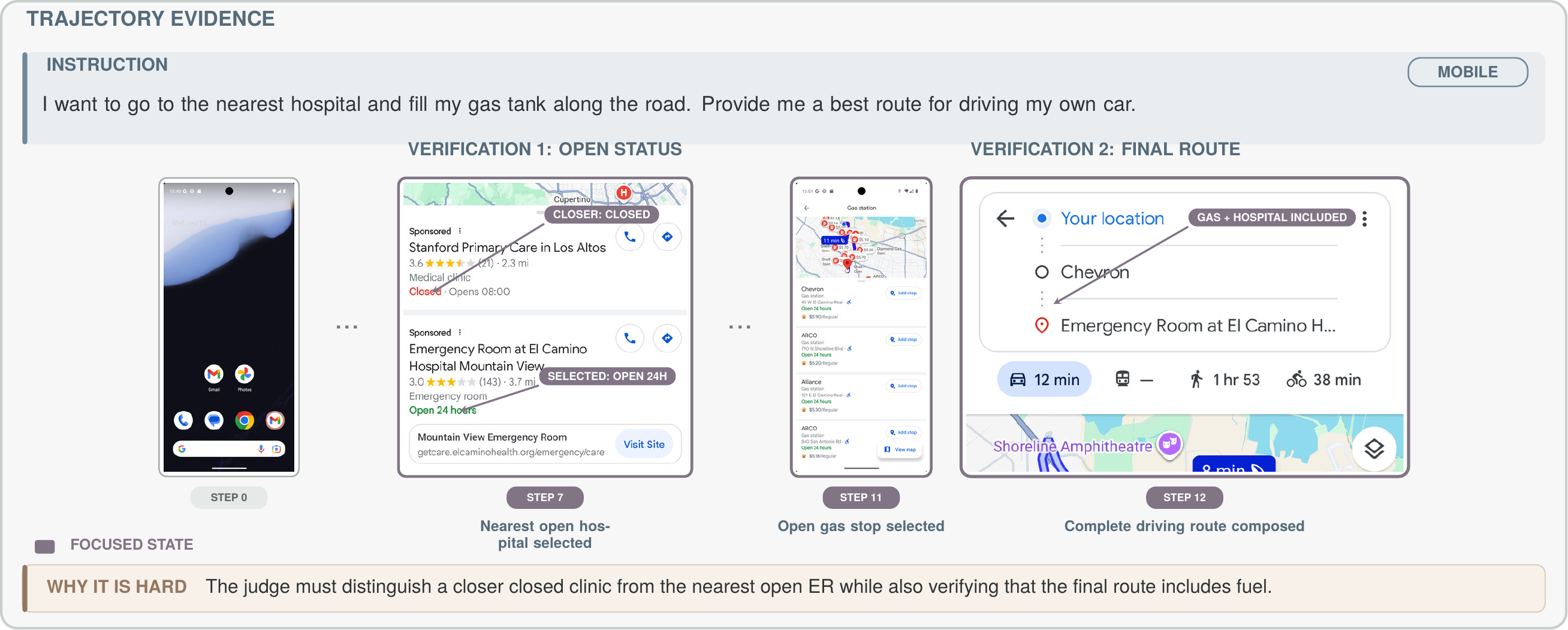}
{A mobile hard case requiring the judge to verify the nearest open hospital and a fuel stop in the final route.}
{fig:google-maps-open-hospital-hard-case}
{\textbf{Human label}\enspace\textsc{success}\hfill\textbf{Reference-judge tendency}\enspace\textsc{unanimous success}}
{Reported a route to the selected open hospital with Chevron added en route, then started navigation.}
{Gemini-3.1-Pro---\textsc{success}: credited the open hospital, gas stop, driving mode, and navigation.}
{Qwen3-VL-8B---\textsc{success}: used the active route to confirm that both stops were included.}


\endgroup

%% file: main.bbl
\begin{thebibliography}{74}
\providecommand{\natexlab}[1]{#1}
\providecommand{\url}[1]{\texttt{#1}}
\expandafter\ifx\csname urlstyle\endcsname\relax
  \providecommand{\doi}[1]{doi: #1}\else
  \providecommand{\doi}{doi: \begingroup \urlstyle{rm}\Url}\fi

\bibitem[Anthropic(2025)]{anthropic2025haiku45}
Anthropic.
\newblock Introducing {Claude Haiku 4.5}.
\newblock \url{https://www.anthropic.com/news/claude-haiku-4-5}, October 2025.

\bibitem[Anthropic(2026{\natexlab{a}})]{anthropic2026opus46}
Anthropic.
\newblock Introducing {Claude Opus 4.6}.
\newblock \url{https://www.anthropic.com/news/claude-opus-4-6}, February 2026{\natexlab{a}}.

\bibitem[Anthropic(2026{\natexlab{b}})]{anthropic2026opus48}
Anthropic.
\newblock Introducing {Claude Opus 4.8}.
\newblock \url{https://www.anthropic.com/news/claude-opus-4-8}, May 2026{\natexlab{b}}.

\bibitem[Anthropic(2026{\natexlab{c}})]{anthropic2026sonnet46}
Anthropic.
\newblock Introducing {Claude Sonnet 4.6}.
\newblock \url{https://www.anthropic.com/news/claude-sonnet-4-6}, February 2026{\natexlab{c}}.

\bibitem[Bai et~al.(2024)Bai, Zhou, Cemri, Pan, Suhr, Levine, and Kumar]{bai2024digirl}
Hao Bai, Yifei Zhou, Mert Cemri, Jiayi Pan, Alane Suhr, Sergey Levine, and Aviral Kumar.
\newblock Digirl: Training in-the-wild device-control agents with autonomous reinforcement learning, 2024.
\newblock URL \url{https://arxiv.org/abs/2406.11896}.

\bibitem[Bai et~al.(2025)Bai, Cai, Chen, Chen, Chen, Cheng, Deng, Ding, Gao, Ge, et~al.]{bai2025qwen3}
Shuai Bai, Yuxuan Cai, Ruizhe Chen, Keqin Chen, Xionghui Chen, Zesen Cheng, Lianghao Deng, Wei Ding, Chang Gao, Chunjiang Ge, et~al.
\newblock Qwen3-vl technical report.
\newblock \emph{arXiv preprint arXiv:2511.21631}, 2025.

\bibitem[Bonatti et~al.(2024)Bonatti, Zhao, Bonacci, Dupont, Abdali, Li, Lu, Wagle, Koishida, Bucker, Jang, and Hui]{bonatti2025windows}
Rogerio Bonatti, Dan Zhao, Francesco Bonacci, Dillon Dupont, Sara Abdali, Yinheng Li, Yadong Lu, Justin Wagle, Kazuhito Koishida, Arthur Bucker, Lawrence Jang, and Zack Hui.
\newblock Windows agent arena: Evaluating multi-modal os agents at scale, 2024.
\newblock URL \url{https://arxiv.org/abs/2409.08264}.

\bibitem[{ByteDance Seed Team}(2026)]{bytedance2026seed2}
{ByteDance Seed Team}.
\newblock Seed2.0 model card: Towards intelligence frontier for real-world complexity.
\newblock Technical report, ByteDance, February 2026.

\bibitem[Chae et~al.(2025)Chae, Kim, Cho, Kim, Moon, Hwangbo, Lim, Kim, Hwang, Gwak, Choi, Kang, Im, Cho, Kim, Han, Kwon, Kim, woo Kwak, Kang, and Yeo]{chae2026webshepherd}
Hyungjoo Chae, Sunghwan Kim, Junhee Cho, Seungone Kim, Seungjun Moon, Gyeom Hwangbo, Dongha Lim, Minjin Kim, Yeonjun Hwang, Minju Gwak, Dongwook Choi, Minseok Kang, Gwanhoon Im, ByeongUng Cho, Hyojun Kim, Jun~Hee Han, Taeyoon Kwon, Minju Kim, Beong woo Kwak, Dongjin Kang, and Jinyoung Yeo.
\newblock Web-shepherd: Advancing {PRM}s for reinforcing web agents.
\newblock In \emph{The Thirty-ninth Annual Conference on Neural Information Processing Systems}, 2025.
\newblock URL \url{https://openreview.net/forum?id=G2kMroO9UV}.

\bibitem[Chen et~al.(2025{\natexlab{a}})Chen, Ji, Zhong, Zhu, Li, Gan, Huang, Zou, Liu, Chen, Chen, and Shen]{chen2025guishepherdandroid}
Cong Chen, Kaixiang Ji, Hao Zhong, Muzhi Zhu, Anzhou Li, Guo Gan, Ziyuan Huang, Cheng Zou, Jiajia Liu, Jingdong Chen, Hao Chen, and Chunhua Shen.
\newblock Gui-shepherd: Reliable process reward and verification for long-sequence gui tasks, 2025{\natexlab{a}}.
\newblock URL \url{https://arxiv.org/abs/2509.23738}.

\bibitem[Chen et~al.(2024)Chen, Chen, Zhang, Liu, Wang, Zhou, Zhang, Wan, Zhou, and Sun]{chen2024mllm}
Dongping Chen, Ruoxi Chen, Shilin Zhang, Yinuo Liu, Yaochen Wang, Huichi Zhou, Qihui Zhang, Yao Wan, Pan Zhou, and Lichao Sun.
\newblock Mllm-as-a-judge: Assessing multimodal llm-as-a-judge with vision-language benchmark, 2024.
\newblock URL \url{https://arxiv.org/abs/2402.04788}.

\bibitem[Chen et~al.(2025{\natexlab{b}})Chen, Chen, Yuan, Peng, Chen, Li, Zhang, Huang, Huang, Liang, et~al.]{chen2025map}
Xuetian Chen, Yinghao Chen, Xinfeng Yuan, Zhuo Peng, Lu~Chen, Yuekeng Li, Zhoujia Zhang, Yingqian Huang, Leyan Huang, Jiaqing Liang, et~al.
\newblock Os-map: How far can computer-using agents go in breadth and depth?
\newblock \emph{arXiv preprint arXiv:2507.19132}, 2025{\natexlab{b}}.

\bibitem[Cheng et~al.(2024)Cheng, Sun, Chu, Xu, YanTao, Zhang, and Wu]{cheng2024seeclick}
Kanzhi Cheng, Qiushi Sun, Yougang Chu, Fangzhi Xu, Li~YanTao, Jianbing Zhang, and Zhiyong Wu.
\newblock {S}ee{C}lick: Harnessing {GUI} grounding for advanced visual {GUI} agents.
\newblock In \emph{Proceedings of the 62nd Annual Meeting of the Association for Computational Linguistics (Volume 1: Long Papers)}, pages 9313--9332, Bangkok, Thailand, August 2024. Association for Computational Linguistics.
\newblock URL \url{https://aclanthology.org/2024.acl-long.505}.

\bibitem[Cheng et~al.(2026)Cheng, Li, Ma, Chen, Cao, Sun, Ding, Xu, Yan, Chen, et~al.]{cheng2026openmobile}
Kanzhi Cheng, Zehao Li, Zheng Ma, Nuo Chen, Jialin Cao, Qiushi Sun, Zichen Ding, Fangzhi Xu, Hang Yan, Jiajun Chen, et~al.
\newblock Openmobile: Building open mobile agents with task and trajectory synthesis.
\newblock \emph{arXiv preprint arXiv:2604.15093}, 2026.

\bibitem[Cobbe et~al.(2021)Cobbe, Kosaraju, Bavarian, Chen, Jun, Kaiser, Plappert, Tworek, Hilton, Nakano, Hesse, and Schulman]{cobbe2021training}
Karl Cobbe, Vineet Kosaraju, Mohammad Bavarian, Mark Chen, Heewoo Jun, Lukasz Kaiser, Matthias Plappert, Jerry Tworek, Jacob Hilton, Reiichiro Nakano, Christopher Hesse, and John Schulman.
\newblock Training verifiers to solve math word problems, 2021.
\newblock URL \url{https://arxiv.org/abs/2110.14168}.

\bibitem[Cohen(1960)]{cohen1960coefficient}
Jacob Cohen.
\newblock A coefficient of agreement for nominal scales.
\newblock \emph{Educational and Psychological Measurement}, 20\penalty0 (1):\penalty0 37--46, 1960.

\bibitem[Comanici et~al.(2025)Comanici, Bieber, Schaekermann, Pasupat, Sachdeva, Dhillon, Blistein, Ram, Zhang, Rosen, et~al.]{comanici2025gemini}
Gheorghe Comanici, Eric Bieber, Mike Schaekermann, Ice Pasupat, Noveen Sachdeva, Inderjit Dhillon, Marcel Blistein, Ori Ram, Dan Zhang, Evan Rosen, et~al.
\newblock Gemini 2.5: Pushing the frontier with advanced reasoning, multimodality, long context, and next generation agentic capabilities.
\newblock \emph{arXiv preprint arXiv:2507.06261}, 2025.

\bibitem[{Gemini Team}(2025)]{gemini2025gemini3}
{Gemini Team}.
\newblock Gemini 3 pro model card, November 2025.
\newblock URL \url{https://deepmind.google/models/gemini/}.

\bibitem[Gou et~al.(2025)Gou, Wang, Zheng, Xie, Chang, Shu, Sun, and Su]{gou2025navigating}
Boyu Gou, Ruohan Wang, Boyuan Zheng, Yanan Xie, Cheng Chang, Yiheng Shu, Huan Sun, and Yu~Su.
\newblock Navigating the digital world as humans do: Universal visual grounding for {GUI} agents.
\newblock In \emph{The Thirteenth International Conference on Learning Representations}, 2025.
\newblock URL \url{https://openreview.net/forum?id=kxnoqaisCT}.

\bibitem[Hurst et~al.(2024)Hurst, Lerer, Goucher, Perelman, Ramesh, Clark, Ostrow, Welihinda, Hayes, Radford, et~al.]{hurst2024gpt}
Aaron Hurst, Adam Lerer, Adam~P Goucher, Adam Perelman, Aditya Ramesh, Aidan Clark, AJ~Ostrow, Akila Welihinda, Alan Hayes, Alec Radford, et~al.
\newblock {GPT-4o} system card.
\newblock \emph{arXiv preprint arXiv:2410.21276}, 2024.

\bibitem[Jia et~al.(2025)Jia, Luo, Dang, Sun, Xu, Hu, Xie, and Wu]{jia2025agentstore}
Chengyou Jia, Minnan Luo, Zhuohang Dang, Qiushi Sun, Fangzhi Xu, Junlin Hu, Tianbao Xie, and Zhiyong Wu.
\newblock {A}gent{S}tore: Scalable integration of heterogeneous agents as specialized generalist computer assistant.
\newblock In \emph{Findings of the Association for Computational Linguistics: ACL 2025}, pages 8908--8934, Vienna, Austria, July 2025. Association for Computational Linguistics.
\newblock ISBN 979-8-89176-256-5.
\newblock \doi{10.18653/v1/2025.findings-acl.466}.
\newblock URL \url{https://aclanthology.org/2025.findings-acl.466/}.

\bibitem[Jiang et~al.(2026)Jiang, Huang, Zhao, Chen, Zheng, Liu, Qiu, Shi, and Zeng]{jiang2026treecua}
Deyang Jiang, Jing Huang, Xuanle Zhao, Lei Chen, Liming Zheng, Fanfan Liu, Haibo Qiu, Peng Shi, and Zhixiong Zeng.
\newblock Tree{CUA}: Efficiently scaling {GUI} automation with tree-structured verifiable evolution.
\newblock In \emph{Forty-third International Conference on Machine Learning}, 2026.
\newblock URL \url{https://openreview.net/forum?id=KBCWS6OBnD}.

\bibitem[{Kimi Team} et~al.(2026){Kimi Team}, Bai, Bai, Bao, Cai, Cao, Charles, Che, Chen, Chen, et~al.]{team2026kimi}
{Kimi Team}, Tongtong Bai, Yifan Bai, Yiping Bao, SH~Cai, Yuan Cao, Y~Charles, HS~Che, Cheng Chen, Guanduo Chen, et~al.
\newblock Kimi k2. 5: Visual agentic intelligence.
\newblock \emph{arXiv preprint arXiv:2602.02276}, 2026.

\bibitem[Kong et~al.(2026)Kong, Zhang, Yang, Gao, Liu, Tong, Cai, Zhou, Zhang, Chen, et~al.]{kong2026mobileworld}
Quyu Kong, Xu~Zhang, Zhenyu Yang, Nolan Gao, Chen Liu, Panrong Tong, Chenglin Cai, Hanzhang Zhou, Jianan Zhang, Liangyu Chen, et~al.
\newblock Mobileworld: Benchmarking autonomous mobile agents in agent-user interactive and mcp-augmented environments.
\newblock In \emph{Proceedings of the 64th Annual Meeting of the Association for Computational Linguistics (Volume 1: Long Papers)}, pages 6142--6167, 2026.

\bibitem[Krippendorff(2011)]{krippendorff2011computing}
Klaus Krippendorff.
\newblock Computing {Krippendorff}'s alpha-reliability.
\newblock Technical report, University of Pennsylvania, Annenberg School for Communication, 2011.

\bibitem[Li et~al.(2025)Li, Wei, Xie, Yang, Song, Wang, An, Liu, Li, Lin, et~al.]{li2025vl}
Lei Li, Yuancheng Wei, Zhihui Xie, Xuqing Yang, Yifan Song, Peiyi Wang, Chenxin An, Tianyu Liu, Sujian Li, Bill~Yuchen Lin, et~al.
\newblock Vl-rewardbench: a challenging benchmark for vision-language generative reward models.
\newblock In \emph{Proceedings of the Computer Vision and Pattern Recognition Conference}, pages 24657--24668, 2025.

\bibitem[Li et~al.(2026)Li, Wu, Zhao, Yang, Xie, Liu, Liu, Jin, Liang, Li, Wu, Zhou, Wang, and Ding]{li2026osthemis}
Zehao Li, Zhenyu Wu, Yibo Zhao, Bowen Yang, Jingjing Xie, Zhaoyang Liu, Zhoumianze Liu, Kaiming Jin, Jianze Liang, Zonglin Li, Feng Wu, Bowen Zhou, Zun Wang, and Zichen Ding.
\newblock Os-themis: A scalable critic framework for generalist gui rewards, 2026.
\newblock URL \url{https://arxiv.org/abs/2603.19191}.

\bibitem[Lin et~al.(2025)Lin, Tan, Qin, Xu, Shi, Li, Li, Cai, Cai, Fu, Li, and Sun]{lin2025cuarewardbench}
Haojia Lin, Xiaoyu Tan, Yulei Qin, Zihan Xu, Yuchen Shi, Zongyi Li, Gang Li, Shaofei Cai, Siqi Cai, Chaoyou Fu, Ke~Li, and Xing Sun.
\newblock Cuarewardbench: A benchmark for evaluating reward models on computer-using agent, 2025.
\newblock URL \url{https://arxiv.org/abs/2510.18596}.

\bibitem[Liu et~al.(2026)Liu, Xie, Ding, Li, Yang, Wu, Wang, Sun, Liu, Wang, Ye, Li, Tian, Luo, Yue, Qi, Chen, Zhou, Qiao, Chen, and Wang]{liu2026scalecua}
Zhaoyang Liu, JingJing Xie, Zichen Ding, Zehao Li, Bowen Yang, Zhenyu Wu, Xuehui Wang, Qiushi Sun, Shi Liu, Weiyun Wang, Shenglong Ye, Qingyun Li, Zeyue Tian, Gen Luo, Xiangyu Yue, Biqing Qi, Kai Chen, Bowen Zhou, Yu~Qiao, Qifeng Chen, and Wenhai Wang.
\newblock Scale{CUA}: Scaling open-source computer use agents with cross-platform data.
\newblock In \emph{The Fourteenth International Conference on Learning Representations}, 2026.
\newblock URL \url{https://openreview.net/forum?id=yBFUqdJFZn}.

\bibitem[Lù et~al.(2025)Lù, Kazemnejad, Meade, Patel, Shin, Zambrano, Stańczak, Shaw, Pal, and Reddy]{lu2025agentrewardbenchweb}
Xing~Han Lù, Amirhossein Kazemnejad, Nicholas Meade, Arkil Patel, Dongchan Shin, Alejandra Zambrano, Karolina Stańczak, Peter Shaw, Christopher~J. Pal, and Siva Reddy.
\newblock Agentrewardbench: Evaluating automatic evaluations of web agent trajectories, 2025.
\newblock URL \url{https://arxiv.org/abs/2504.08942}.

\bibitem[OpenAI(2025{\natexlab{a}})]{openai2025cua}
OpenAI.
\newblock Computer-using agent: Introducing a universal interface for ai to interact with the digital world, 2025{\natexlab{a}}.
\newblock URL \url{https://openai.com/index/computer-using-agent}.

\bibitem[OpenAI(2025{\natexlab{b}})]{openai2025gpt5}
OpenAI.
\newblock {GPT-5 System Card}.
\newblock Technical report, OpenAI, August 2025{\natexlab{b}}.
\newblock URL \url{https://openai.com/index/gpt-5-system-card/}.

\bibitem[OpenAI(2026{\natexlab{a}})]{openai2026gpt54thinking}
OpenAI.
\newblock {GPT-5.4 Thinking System Card}.
\newblock Technical report, OpenAI, March 2026{\natexlab{a}}.
\newblock URL \url{https://openai.com/index/gpt-5-4-thinking-system-card/}.

\bibitem[OpenAI(2026{\natexlab{b}})]{openai2026gpt55}
OpenAI.
\newblock {GPT-5.5 System Card}.
\newblock Technical report, OpenAI, April 2026{\natexlab{b}}.
\newblock URL \url{https://openai.com/index/gpt-5-5-system-card/}.

\bibitem[Pan et~al.(2024)Pan, Zhang, Tomlin, Zhou, Levine, and Suhr]{pan2024autonomous}
Jiayi Pan, Yichi Zhang, Nicholas Tomlin, Yifei Zhou, Sergey Levine, and Alane Suhr.
\newblock Autonomous evaluation and refinement of digital agents.
\newblock In \emph{First Conference on Language Modeling}, 2024.
\newblock URL \url{https://openreview.net/forum?id=NPAQ6FKSmK}.

\bibitem[Qi et~al.(2024)Qi, Liu, Iong, Lai, Sun, Zhao, Yang, Yang, Sun, Yao, Zhang, Xu, Tang, and Dong]{qi2024webrl}
Zehan Qi, Xiao Liu, Iat~Long Iong, Hanyu Lai, Xueqiao Sun, Wenyi Zhao, Yu~Yang, Xinyue Yang, Jiadai Sun, Shuntian Yao, Tianjie Zhang, Wei Xu, Jie Tang, and Yuxiao Dong.
\newblock Webrl: Training llm web agents via self-evolving online curriculum reinforcement learning, 2024.
\newblock URL \url{https://arxiv.org/abs/2411.02337}.

\bibitem[Qin et~al.(2025)Qin, Ye, Fang, Wang, Liang, Tian, Zhang, Li, Li, Huang, et~al.]{qin2025uitars}
Yujia Qin, Yining Ye, Junjie Fang, Haoming Wang, Shihao Liang, Shizuo Tian, Junda Zhang, Jiahao Li, Yunxin Li, Shijue Huang, et~al.
\newblock Ui-tars: Pioneering automated gui interaction with native agents.
\newblock \emph{arXiv preprint arXiv:2501.12326}, 2025.

\bibitem[{Qwen Team}(2026)]{qwen3_5}
{Qwen Team}.
\newblock {Qwen3.5}: Towards native multimodal agents, February 2026.
\newblock URL \url{https://qwen.ai/blog?id=qwen3.5}.

\bibitem[Rawles et~al.(2025)Rawles, Clinckemaillie, Chang, Waltz, Lau, Fair, Li, Bishop, Li, Campbell-Ajala, Toyama, Berry, Tyamagundlu, Lillicrap, and Riva]{androidworld}
Christopher Rawles, Sarah Clinckemaillie, Yifan Chang, Jonathan Waltz, Gabrielle Lau, Marybeth Fair, Alice Li, William~E Bishop, Wei Li, Folawiyo Campbell-Ajala, Daniel~Kenji Toyama, Robert~James Berry, Divya Tyamagundlu, Timothy~P Lillicrap, and Oriana Riva.
\newblock Androidworld: A dynamic benchmarking environment for autonomous agents.
\newblock In \emph{The Thirteenth International Conference on Learning Representations}, 2025.
\newblock URL \url{https://openreview.net/forum?id=il5yUQsrjC}.

\bibitem[Shao et~al.(2024)Shao, Wang, Zhu, Xu, Song, Bi, Zhang, Zhang, Li, Wu, et~al.]{shao2024deepseekmath}
Zhihong Shao, Peiyi Wang, Qihao Zhu, Runxin Xu, Junxiao Song, Xiao Bi, Haowei Zhang, Mingchuan Zhang, YK~Li, Yang Wu, et~al.
\newblock Deepseekmath: Pushing the limits of mathematical reasoning in open language models.
\newblock \emph{arXiv preprint arXiv:2402.03300}, 2024.

\bibitem[Sheng et~al.(2024)Sheng, Zhang, Ye, Wu, Zhang, Zhang, Peng, Lin, and Wu]{sheng2024hybridflow}
Guangming Sheng, Chi Zhang, Zilingfeng Ye, Xibin Wu, Wang Zhang, Ru~Zhang, Yanghua Peng, Haibin Lin, and Chuan Wu.
\newblock Hybridflow: A flexible and efficient {RLHF} framework.
\newblock \emph{arXiv preprint arXiv:2409.19256}, 2024.

\bibitem[Sun et~al.(2024)Sun, Chen, Xu, Cheng, Ma, Yin, Wang, Han, Zhu, Yuan, et~al.]{sun2024survey}
Qiushi Sun, Zhirui Chen, Fangzhi Xu, Kanzhi Cheng, Chang Ma, Zhangyue Yin, Jianing Wang, Chengcheng Han, Renyu Zhu, Shuai Yuan, et~al.
\newblock A survey of neural code intelligence: Paradigms, advances and beyond.
\newblock \emph{arXiv preprint arXiv:2403.14734}, 2024.

\bibitem[Sun et~al.(2025{\natexlab{a}})Sun, Cheng, Ding, Jin, Wang, Xu, Wu, Jia, Chen, Liu, et~al.]{sun2025osgenesis}
Qiushi Sun, Kanzhi Cheng, Zichen Ding, Chuanyang Jin, Yian Wang, Fangzhi Xu, Zhenyu Wu, Chengyou Jia, Liheng Chen, Zhoumianze Liu, et~al.
\newblock Os-genesis: Automating gui agent trajectory construction via reverse task synthesis.
\newblock In \emph{Proceedings of the 63rd Annual Meeting of the Association for Computational Linguistics (Volume 1: Long Papers)}, pages 5555--5579, 2025{\natexlab{a}}.

\bibitem[Sun et~al.(2026{\natexlab{a}})Sun, Li, Liu, Xie, Xu, Yin, Cheng, Li, Ding, Liu, Wu, Zhang, Kao, and Kong]{sun2025sentinel}
Qiushi Sun, Mukai Li, Zhoumianze Liu, Zhihui Xie, Fangzhi Xu, Zhangyue Yin, Kanzhi Cheng, Zehao Li, Zichen Ding, Qi~Liu, Zhiyong Wu, Zhuosheng Zhang, Ben Kao, and Lingpeng Kong.
\newblock {OS}-sentinel: Towards safety-enhanced mobile {GUI} agents via hybrid validation in realistic workflows.
\newblock In \emph{Proceedings of the 64th Annual Meeting of the {A}ssociation for {C}omputational {L}inguistics (Volume 1: Long Papers)}, pages 9529--9553, San Diego, California, United States, July 2026{\natexlab{a}}. Association for Computational Linguistics.
\newblock ISBN 979-8-89176-390-6.
\newblock URL \url{https://aclanthology.org/2026.acl-long.431/}.

\bibitem[Sun et~al.(2026{\natexlab{b}})Sun, Liu, Ma, Ding, Xu, Yin, Zhao, Wu, Cheng, Liu, Wang, Li, Tang, Xie, Feng, Li, Kao, Wang, Qi, Kong, and Wu]{sun2026scienceboard}
Qiushi Sun, Zhoumianze Liu, Chang Ma, Zichen Ding, Fangzhi Xu, Zhangyue Yin, Haiteng Zhao, Zhenyu Wu, Kanzhi Cheng, Zhaoyang Liu, Jianing Wang, Qintong Li, Xiangru Tang, Tianbao Xie, Xiachong Feng, Xiang Li, Ben Kao, Wenhai Wang, Biqing Qi, Lingpeng Kong, and Zhiyong Wu.
\newblock Scienceboard: Evaluating multimodal autonomous agents in realistic scientific workflows.
\newblock In \emph{The Fourteenth International Conference on Learning Representations}, 2026{\natexlab{b}}.
\newblock URL \url{https://openreview.net/forum?id=bJvwJahJeF}.

\bibitem[Sun et~al.(2025{\natexlab{b}})Sun, Cao, Liang, Sun, Liu, Zhang, Zang, Dong, Chen, Lin, et~al.]{sun2025coda}
Zeyi Sun, Yuhang Cao, Jianze Liang, Qiushi Sun, Ziyu Liu, Zhixiong Zhang, Yuhang Zang, Xiaoyi Dong, Kai Chen, Dahua Lin, et~al.
\newblock Coda: Coordinating the cerebrum and cerebellum for a dual-brain computer use agent with decoupled reinforcement learning.
\newblock \emph{arXiv preprint arXiv:2508.20096}, 2025{\natexlab{b}}.

\bibitem[Trabucco et~al.(2025)Trabucco, Sigurdsson, Piramuthu, and Salakhutdinov]{trabucco2025insta}
Brandon Trabucco, Gunnar Sigurdsson, Robinson Piramuthu, and Ruslan Salakhutdinov.
\newblock {InSTA}: Towards internet-scale training for agents.
\newblock \emph{arXiv preprint arXiv:2502.06776}, 2025.

\bibitem[Wang et~al.(2024)Wang, Li, Shao, Xu, Dai, Li, Chen, Wu, and Sui]{wang2024mathshepherd}
Peiyi Wang, Lei Li, Zhihong Shao, Runxin Xu, Damai Dai, Yifei Li, Deli Chen, Yu~Wu, and Zhifang Sui.
\newblock Math-shepherd: Verify and reinforce {LLM}s step-by-step without human annotations.
\newblock In \emph{Proceedings of the 62nd Annual Meeting of the Association for Computational Linguistics (Volume 1: Long Papers)}, pages 9426--9439, Bangkok, Thailand, August 2024. Association for Computational Linguistics.
\newblock \doi{10.18653/v1/2024.acl-long.510}.
\newblock URL \url{https://aclanthology.org/2024.acl-long.510/}.

\bibitem[Wang et~al.(2025)Wang, Wang, Lu, Yang, Xie, Wang, Deng, Guo, Xu, Wu, Shen, Li, Li, Li, Chen, Boyuan, PEIHANG, Lei, Cao, Fu, Shin, Shin, Jiarui, Wang, Chen, Ye, Zhang, Wang, Wang, Yang, Zhong, Charles, Yang, and Yu]{wang2026opencua}
Xinyuan Wang, Bowen Wang, Dunjie Lu, Junlin Yang, Tianbao Xie, Junli Wang, Jiaqi Deng, Xiaole Guo, Yiheng Xu, Chen~Henry Wu, Zhennan Shen, Zhuokai Li, Ryan Li, Xiaochuan Li, Junda Chen, Zheng Boyuan, LI~PEIHANG, Fangyu Lei, Ruisheng Cao, Yeqiao Fu, Dongchan Shin, Martin Shin, Hu~Jiarui, Yuyan Wang, Jixuan Chen, Yuxiao Ye, Danyang Zhang, Yipu Wang, Heng Wang, Diyi Yang, Victor Zhong, Y.~Charles, Zhilin Yang, and Tao Yu.
\newblock Open{CUA}: Open foundations for computer-use agents.
\newblock In \emph{The Thirty-ninth Annual Conference on Neural Information Processing Systems}, 2025.
\newblock URL \url{https://openreview.net/forum?id=6iRZvJiC9Q}.

\bibitem[Wang et~al.(2026)Wang, Liang, Zhang, Wu, Han, Bastos, Wang, Bansal, Peng, Gao, Rajmohan, and Yao]{wang2026synthagent}
Zhaoyang Wang, Yiming Liang, Xuchao Zhang, Qianhui Wu, Siwei Han, Anson Bastos, Rujia Wang, Chetan Bansal, Baolin Peng, Jianfeng Gao, Saravan Rajmohan, and Huaxiu Yao.
\newblock {S}ynth{A}gent: Adapting web agents with synthetic supervision.
\newblock In \emph{Proceedings of the 64th Annual Meeting of the {A}ssociation for {C}omputational {L}inguistics (Volume 1: Long Papers)}, pages 15730--15752, San Diego, California, United States, July 2026. Association for Computational Linguistics.
\newblock ISBN 979-8-89176-390-6.
\newblock URL \url{https://aclanthology.org/2026.acl-long.716/}.

\bibitem[Wu et~al.(2025{\natexlab{a}})Wu, Cheng, Yang, Zhang, Yang, Jiang, Mu, Peng, Qiao, Tan, et~al.]{wu2025gui}
Qianhui Wu, Kanzhi Cheng, Rui Yang, Chaoyun Zhang, Jianwei Yang, Huiqiang Jiang, Jian Mu, Baolin Peng, Bo~Qiao, Reuben Tan, et~al.
\newblock Gui-actor: Coordinate-free visual grounding for gui agents.
\newblock \emph{arXiv preprint arXiv:2506.03143}, 2025{\natexlab{a}}.

\bibitem[Wu et~al.(2026)Wu, Xie, Li, Yang, Sun, Liu, Liu, Qiao, Yue, Wang, et~al.]{wu2026oracle}
Zhenyu Wu, Jingjing Xie, Zehao Li, Bowen Yang, Qiushi Sun, Zhaoyang Liu, Zhoumianze Liu, Yu~Qiao, Xiangyu Yue, Zun Wang, et~al.
\newblock Os-oracle: A comprehensive framework for cross-platform gui critic models.
\newblock In \emph{Proceedings of the IEEE/CVF Conference on Computer Vision and Pattern Recognition}, pages 27514--27524, 2026.

\bibitem[Wu et~al.(2024)Wu, Han, Ding, Weng, Liu, Yao, Yu, and Kong]{wu2024oscopilot}
Zhiyong Wu, Chengcheng Han, Zichen Ding, Zhenmin Weng, Zhoumianze Liu, Shunyu Yao, Tao Yu, and Lingpeng Kong.
\newblock Os-copilot: Towards generalist computer agents with self-improvement, 2024.
\newblock URL \url{https://arxiv.org/abs/2402.07456}.

\bibitem[Wu et~al.(2025{\natexlab{b}})Wu, Wu, Xu, Wang, Sun, Jia, Cheng, Ding, Chen, Liang, and Qiao]{wu2025osatlas}
Zhiyong Wu, Zhenyu Wu, Fangzhi Xu, Yian Wang, Qiushi Sun, Chengyou Jia, Kanzhi Cheng, Zichen Ding, Liheng Chen, Paul~Pu Liang, and Yu~Qiao.
\newblock {OS}-{ATLAS}: Foundation action model for generalist {GUI} agents.
\newblock In \emph{The Thirteenth International Conference on Learning Representations}, 2025{\natexlab{b}}.
\newblock URL \url{https://openreview.net/forum?id=n9PDaFNi8t}.

\bibitem[Xiao et~al.(2025)Xiao, Wang, Chai, Lu, Lin, He, Fan, Bian, Hu, Liu, Ren, Wen, Chen, Zhou, and Li]{xiao2025uigenie}
Han Xiao, Guozhi Wang, Yuxiang Chai, Zimu Lu, Weifeng Lin, Hao He, Lue Fan, Liuyang Bian, Rui Hu, Liang Liu, Shuai Ren, Yafei Wen, Xiaoxin Chen, Aojun Zhou, and Hongsheng Li.
\newblock {UI}-genie: A self-improving approach for iteratively boosting {MLLM}-based mobile {GUI} agents.
\newblock In \emph{The Thirty-ninth Annual Conference on Neural Information Processing Systems}, 2025.
\newblock URL \url{https://openreview.net/forum?id=3uUmJzSSOW}.

\bibitem[Xie et~al.(2024)Xie, Zhang, Chen, Li, Zhao, Cao, Hua, Cheng, Shin, Lei, Liu, Xu, Zhou, Savarese, Xiong, Zhong, and Yu]{osworld}
Tianbao Xie, Danyang Zhang, Jixuan Chen, Xiaochuan Li, Siheng Zhao, Ruisheng Cao, Toh~Jing Hua, Zhoujun Cheng, Dongchan Shin, Fangyu Lei, Yitao Liu, Yiheng Xu, Shuyan Zhou, Silvio Savarese, Caiming Xiong, Victor Zhong, and Tao Yu.
\newblock {OSW}orld: Benchmarking multimodal agents for open-ended tasks in real computer environments.
\newblock In \emph{The Thirty-eight Conference on Neural Information Processing Systems Datasets and Benchmarks Track}, 2024.
\newblock URL \url{https://openreview.net/forum?id=tN61DTr4Ed}.

\bibitem[Xie et~al.(2025)Xie, Yuan, Zhang, Xiong, Shen, Zhou, Wang, Chen, Deng, Chen, Wang, Wu, Chen, Wang, Lu, Hu, and Yu]{osworldverified}
Tianbao Xie, Mengqi Yuan, Danyang Zhang, Xinzhuang Xiong, Zhennan Shen, Zilong Zhou, Xinyuan Wang, Yanxu Chen, Jiaqi Deng, Junda Chen, Bowen Wang, Haoyuan Wu, Jixuan Chen, Junli Wang, Dunjie Lu, Hao Hu, and Tao Yu.
\newblock Introducing osworld-verified.
\newblock \emph{xlang.ai}, July 2025.
\newblock URL \url{https://xlang.ai/blog/osworld-verified}.

\bibitem[Xu et~al.(2026)Xu, Yan, Sun, Wu, Huang, Huang, Gong, Ding, Cheng, Wang, et~al.]{xu2026odysseyarena}
Fangzhi Xu, Hang Yan, Qiushi Sun, Jinyang Wu, Zixian Huang, Muye Huang, Jingyang Gong, Zichen Ding, Kanzhi Cheng, Yian Wang, et~al.
\newblock Odysseyarena: Benchmarking large language models for long-horizon, active and inductive interactions.
\newblock \emph{arXiv preprint arXiv:2602.05843}, 2026.

\bibitem[Xu et~al.(2025{\natexlab{a}})Xu, Lu, Shen, Wang, Wang, Mao, Xiong, and Yu]{xu2025agenttrek}
Yiheng Xu, Dunjie Lu, Zhennan Shen, Junli Wang, Zekun Wang, Yuchen Mao, Caiming Xiong, and Tao Yu.
\newblock Agenttrek: Agent trajectory synthesis via guiding replay with web tutorials.
\newblock In \emph{The Thirteenth International Conference on Learning Representations}, 2025{\natexlab{a}}.
\newblock URL \url{https://openreview.net/forum?id=EEgYUccwsV}.

\bibitem[Xu et~al.(2025{\natexlab{b}})Xu, Wang, Wang, Lu, Xie, Saha, Sahoo, Yu, and Xiong]{xu2025aguvis}
Yiheng Xu, Zekun Wang, Junli Wang, Dunjie Lu, Tianbao Xie, Amrita Saha, Doyen Sahoo, Tao Yu, and Caiming Xiong.
\newblock Aguvis: Unified pure vision agents for autonomous {GUI} interaction.
\newblock In \emph{Forty-second International Conference on Machine Learning}, 2025{\natexlab{b}}.
\newblock URL \url{https://openreview.net/forum?id=PlihOwfx4r}.

\bibitem[Xue et~al.(2026{\natexlab{a}})Xue, Peng, Huang, Guo, Han, Wang, Zhang, Yang, Zhao, Ding, Ma, Xie, Pei, Cai, and Qiu]{xue2026evocua}
Taofeng Xue, Chong Peng, Mianqiu Huang, Linsen Guo, Tiancheng Han, Haozhe Wang, Xiaocheng Zhang, Xin Yang, Dengchang Zhao, Jinrui Ding, Xiandi Ma, Yuchen Xie, Peng Pei, Xunliang Cai, and Xipeng Qiu.
\newblock Evo{CUA}: Evolving computer use agents via learning from scalable synthetic experience.
\newblock In \emph{Second Workshop on Agents in the Wild: Safety, Security, and Beyond}, 2026{\natexlab{a}}.
\newblock URL \url{https://openreview.net/forum?id=pzbtm6fRIo}.

\bibitem[Xue et~al.(2025)Xue, Qi, Shi, Song, Gou, Song, Sun, and Su]{xue2025an}
Tianci Xue, Weijian Qi, Tianneng Shi, Chan~Hee Song, Boyu Gou, Dawn Song, Huan Sun, and Yu~Su.
\newblock An illusion of progress? assessing the current state of web agents.
\newblock In \emph{Second Conference on Language Modeling}, 2025.
\newblock URL \url{https://openreview.net/forum?id=6jZi4HSs6o}.

\bibitem[Xue et~al.(2026{\natexlab{b}})Xue, Liao, Shi, Wang, Zhang, Song, Su, and Sun]{xue2026acurl}
Tianci Xue, Zeyi Liao, Tianneng Shi, Zilu Wang, Kai Zhang, Dawn Song, Yu~Su, and Huan Sun.
\newblock Autonomous continual learning for environment adaptation of computer-use agents, 2026{\natexlab{b}}.
\newblock URL \url{https://arxiv.org/abs/2602.10356}.

\bibitem[Yang et~al.(2026)Yang, Jin, Wu, Liu, Sun, Li, Xie, Liu, Xu, Cheng, Wang, Li, Qiao, Wang, and Ding]{yang2026symphony}
Bowen Yang, Kaiming Jin, Zhenyu Wu, Zhaoyang Liu, Qiushi Sun, Zehao Li, JingJing Xie, Zhoumianze Liu, Fangzhi Xu, Kanzhi Cheng, Yian Wang, Qingyun Li, Yu~Qiao, Zun Wang, and Zichen Ding.
\newblock {OS}-symphony: A holistic framework for robust and generalist computer-using agents.
\newblock In \emph{Proceedings of the 64th Annual Meeting of the {A}ssociation for {C}omputational {L}inguistics (Volume 1: Long Papers)}, pages 22300--22330, San Diego, California, United States, July 2026. Association for Computational Linguistics.
\newblock ISBN 979-8-89176-390-6.
\newblock URL \url{https://aclanthology.org/2026.acl-long.1021/}.

\bibitem[Yang et~al.(2023)Yang, Zhang, Li, Zou, Li, and Gao]{yang2023setofmark}
Jianwei Yang, Hao Zhang, Feng Li, Xueyan Zou, Chunyuan Li, and Jianfeng Gao.
\newblock Set-of-mark prompting unleashes extraordinary visual grounding in gpt-4v.
\newblock \emph{arXiv preprint arXiv:2310.11441}, 2023.

\bibitem[Zhang et~al.(2025)Zhang, Ding, Ma, Chen, Sun, Lan, and He]{zhang2025guimid}
Junlei Zhang, Zichen Ding, Chang Ma, Zijie Chen, Qiushi Sun, Zhenzhong Lan, and Junxian He.
\newblock Breaking the data barrier -- building {GUI} agents through task generalization.
\newblock In \emph{Second Conference on Language Modeling}, 2025.
\newblock URL \url{https://openreview.net/forum?id=QDtORaZt8K}.

\bibitem[Zhang et~al.(2026)Zhang, Chen, Liu, Xue, Liao, Liu, Wang, Ning, Chen, Fu, Xie, Sun, Gou, Qi, Meng, Yang, Zhang, Li, Shah, Huynh, Li, Yang, Cao, Jang, Zhou, Zhu, Sun, Weston, Su, and Wu]{zhang2026agent}
Kai Zhang, Xiangchao Chen, Bo~Liu, Tianci Xue, Zeyi Liao, Zhihan Liu, Xiyao Wang, Yuting Ning, Zhaorun Chen, Xiaohan Fu, Jian Xie, Yuxuan Sun, Boyu Gou, Qi~Qi, Zihang Meng, Jianwei Yang, Ning Zhang, Xian Li, Ashish Shah, Dat Huynh, Hengduo Li, Zi~Yang, Xuefei Cao, Lawrence~Keunho Jang, Shuyan Zhou, Jiacheng Zhu, Huan Sun, Jason~E Weston, Yu~Su, and Yifan Wu.
\newblock Agent learning via early experience.
\newblock In \emph{Forty-third International Conference on Machine Learning}, 2026.
\newblock URL \url{https://openreview.net/forum?id=N3dXUHY5dD}.

\bibitem[Zheng et~al.(2024)Zheng, Gou, Kil, Sun, and Su]{zheng2024seeact}
Boyuan Zheng, Boyu Gou, Jihyung Kil, Huan Sun, and Yu~Su.
\newblock Gpt-4v(ision) is a generalist web agent, if grounded.
\newblock In \emph{Forty-first International Conference on Machine Learning}, 2024.
\newblock URL \url{https://openreview.net/forum?id=piecKJ2DlB}.

\bibitem[Zheng et~al.(2023{\natexlab{a}})Zheng, Chiang, Sheng, Zhuang, Wu, Zhuang, Lin, Li, Li, Xing, Zhang, Gonzalez, and Stoica]{llmjudge}
Lianmin Zheng, Wei-Lin Chiang, Ying Sheng, Siyuan Zhuang, Zhanghao Wu, Yonghao Zhuang, Zi~Lin, Zhuohan Li, Dacheng Li, Eric~P. Xing, Hao Zhang, Joseph~E. Gonzalez, and Ion Stoica.
\newblock Judging llm-as-a-judge with mt-bench and chatbot arena.
\newblock In \emph{Proceedings of the 37th International Conference on Neural Information Processing Systems}, NIPS '23, Red Hook, NY, USA, 2023{\natexlab{a}}. Curran Associates Inc.

\bibitem[Zheng et~al.(2023{\natexlab{b}})Zheng, Yin, Xie, Sun, Huang, Yu, Cao, Kozyrakis, Stoica, Gonzalez, Barrett, and Sheng]{zheng2023sglang}
Lianmin Zheng, Liangsheng Yin, Zhiqiang Xie, Chuyue Sun, Jeff Huang, Cody~Hao Yu, Shiyi Cao, Christos Kozyrakis, Ion Stoica, Joseph~E. Gonzalez, Clark Barrett, and Ying Sheng.
\newblock {SGLang}: Efficient execution of structured language model programs.
\newblock \emph{arXiv preprint arXiv:2312.07104}, 2023{\natexlab{b}}.

\bibitem[Zhou et~al.(2025)Zhou, Zheng, Wang, Xi, Dou, Bao, Shen, Xiong, Fan, Mou, Zheng, Gui, Zhang, and Huang]{zhou2025rmb}
Enyu Zhou, Guodong Zheng, Binghai Wang, Zhiheng Xi, Shihan Dou, Rong Bao, Wei Shen, Limao Xiong, Jessica Fan, Yurong Mou, Rui Zheng, Tao Gui, Qi~Zhang, and Xuanjing Huang.
\newblock {RMB}: Comprehensively benchmarking reward models in {LLM} alignment.
\newblock In \emph{The Thirteenth International Conference on Learning Representations}, 2025.
\newblock URL \url{https://openreview.net/forum?id=kmgrlG9TR0}.

\bibitem[Zhou et~al.(2024)Zhou, Xu, Zhu, Zhou, Lo, Sridhar, Cheng, Ou, Bisk, Fried, Alon, and Neubig]{webarena}
Shuyan Zhou, Frank~F. Xu, Hao Zhu, Xuhui Zhou, Robert Lo, Abishek Sridhar, Xianyi Cheng, Tianyue Ou, Yonatan Bisk, Daniel Fried, Uri Alon, and Graham Neubig.
\newblock Webarena: A realistic web environment for building autonomous agents.
\newblock In \emph{The Twelfth International Conference on Learning Representations}, 2024.
\newblock URL \url{https://openreview.net/forum?id=oKn9c6ytLx}.

\bibitem[Zhuge et~al.(2025)Zhuge, Zhao, Ashley, Wang, Khizbullin, Xiong, Liu, Chang, Krishnamoorthi, Tian, Shi, Chandra, and Schmidhuber]{zhuge2025agentasajudge}
Mingchen Zhuge, Changsheng Zhao, Dylan~R. Ashley, Wenyi Wang, Dmitrii Khizbullin, Yunyang Xiong, Zechun Liu, Ernie Chang, Raghuraman Krishnamoorthi, Yuandong Tian, Yangyang Shi, Vikas Chandra, and J{\"u}rgen Schmidhuber.
\newblock Agent-as-a-judge: Evaluate agents with agents.
\newblock In \emph{Forty-second International Conference on Machine Learning}, 2025.
\newblock URL \url{https://openreview.net/forum?id=Nn9POI9Ekt}.

\bibitem[Zou et~al.(2026)Zou, Zhu, Zhu, Zhu, Zhou, Zhou, Zhou, Zhou, Zhou, Zhou, et~al.]{zou2026interns1pro}
Yicheng Zou, Dongsheng Zhu, Lin Zhu, Tong Zhu, Yunhua Zhou, Peiheng Zhou, Xinyu Zhou, Dongzhan Zhou, Zhiwang Zhou, Yuhao Zhou, et~al.
\newblock Intern-s1-pro: Scientific multimodal foundation model at trillion scale.
\newblock \emph{arXiv preprint arXiv:2603.25040}, 2026.

\end{thebibliography}
